\documentclass[11pt]{article}

\usepackage[preprint]{acl}

\usepackage{times}
\usepackage{latexsym}

\usepackage{subcaption}
\usepackage{booktabs}
\usepackage{multirow}
\usepackage{amssymb}
\usepackage{tikz}
\usepackage{euflag}
\usepackage{comment}
\usepackage{orcidlink}
\usepackage{amsmath}
\usepackage{mathtools}
\usepackage{amsthm}
\usepackage{amsfonts}
\usepackage{placeins}
\usepackage{algorithmic}
\usepackage{algorithm}
\usepackage{orcidlink}
\usepackage[capitalize,noabbrev]{cleveref}
\usepackage[textsize=tiny]{todonotes}

\input{macros}

\usepackage[T1]{fontenc}

\usepackage[utf8]{inputenc}

\usepackage{microtype}

\usepackage{inconsolata}

\usepackage{graphicx}

\title{Quantifying Retriever-Generator Alignment in RAG with Local Explanations}

\author{
  \textbf{Korbinian Randl\orcidlink{0000-0002-7938-2747}\textsuperscript{1,\dag},}
  \textbf{Guido Rocchietti\orcidlink{0009-0004-9704-0662}\textsuperscript{2,\ddag},}
  \textbf{Aron Henriksson\orcidlink{0000-0001-9731-1048}\textsuperscript{1},} \\%
  \textbf{Ziawasch Abedjan\orcidlink{0000-0002-2846-1373}\textsuperscript{2},}
  \textbf{Tony Lindgren\orcidlink{0000-0001-7713-1381}\textsuperscript{1}, \and}%
  \textbf{John Pavlopoulos\orcidlink{0000-0001-9188-7425}\textsuperscript{1,3,4}} \\[5pt]%
  \textsuperscript{1}Department of Computer and Systems Sciences, Stockholm University, Sweden \\%
  \textsuperscript{2}BIFOLD, Technische Universität Berlin, Germany \\%
  \textsuperscript{3}Department of Informatics, Athens University of Economics and Business, Greece \\%
  \textsuperscript{4}Archimedes, Athena Research Centre, Greece \\[5pt]%
  \textsuperscript{\dag}\texttt{korbinian.randl@dsv.su.se},
  \textsuperscript{\ddag}\texttt{guido.rocchietti@tu-berlin.de}
}

\begin{document}
\maketitle
\begin{abstract}
  Retrieval-Augmented Generation (RAG) systems combine dense retrievers and language models to ground outputs in external documents. However, the interaction between these components remains opaque, creating challenges for deployment in high-stakes domains. We present RAG-E, an end-to-end explainability framework that quantifies retriever-generator alignment through mathematically grounded attribution methods. Our approach adapts Integrated Gradients for retriever analysis, proposes a Monte Carlo-stabilized Shapley Value approximation for generator attribution, and introduces the \textit{Weighted Alignment between Retriever and Generator}~(WARG) metric to measure how closely the generator's document usage aligns with retriever rankings. Experiments on PopQA, QAMPARI, and TREC CAST datasets reveal substantial misalignment: depending on the model and setting, generators often ignore top-ranked documents and rely on documents ranked as less relevant. We show that WARG captures retriever-generator alignment better than Pearson and Spearman correlations and can serve as an indicator of RAG performance. RAG-E and WARG provide a practical framework for auditing this interaction, enabling more reliable and transparent RAG systems.
\end{abstract}

\section{Introduction}

\begin{figure*}[t]
\centering

\begin{subfigure}{0.58\textwidth}
\centering

\resizebox{\linewidth}{!}{
\begin{tikzpicture}
\setlength{\tabcolsep}{2pt}
\renewcommand{\arraystretch}{1.1}

\draw (0,0) node {\resizetable{10cm}{
\begin{tabular}{rl}
\toprule

\textbf{Query:}
 & {\small
    \tok{cyan!11}{Where}
    \tok{cyan!5}{ was}
    \tok{blue!100}{ Marie}
    \tok{blue!86}{ Cur}
    \tok{violet!51}{ie}
    \tok{violet!47}{ born}
    \tok{violet!34}{?}
   } \\[-2pt]
 & {\scriptsize\textit{Legend:}\quad \gradlegend} \\

\midrule

\textbf{Doc. 1:} &
{\small
\tok{cyan!67}{ Maria}
\tok{cyan!8}{ Sk}
\tok{cyan!5}{\l{}o}
\tok{cyan!4}{d}
\tok{cyan!8}{owska}
\tok{cyan!8}{,}
\tok{cyan!5}{ later}
\tok{cyan!9}{ known}
\tok{cyan!7}{ as}
\tok{cyan!86}{ Marie}
\tok{cyan!70}{ Cur}
\tok{cyan!47}{ie}
\tok{cyan!2}{,}
\tok{cyan!16}{ was}
\tok{cyan!25}{ born}
\tok{cyan!0}{ on}
\tok{cyan!7}{ November}
\tok{cyan!6}{ 7}
\tok{cyan!4}{,}
\tok{cyan!13}{ 1867}
\tok{cyan!2}{.}
}
\\

\midrule

\multicolumn{1}{c}{$\vdots$} & \multicolumn{1}{c}{$\vdots$} \\

\midrule

\textbf{Doc. 5:} &
{\small
\tok{cyan!63}{ Maria}
\tok{cyan!4}{ Sk}
\tok{cyan!1}{\l{}o}
\tok{cyan!3}{d}
\tok{cyan!3}{owska}
\tok{cyan!15}{ was}
\tok{cyan!28}{ born}
\tok{cyan!6}{ in}
\tok{cyan!12}{ Wars}
\tok{cyan!1}{aw}
\tok{cyan!0}{,}
\tok{cyan!2}{ in}
\tok{cyan!6}{ Congress}
\tok{cyan!8}{ Poland}
\tok{cyan!0}{ in}
\tok{cyan!3}{ the}
\tok{cyan!8}{ Russian}
\tok{cyan!11}{ Empire}
\tok{cyan!0}{,}} \\
& {\small
\tok{cyan!5}{ as}
\tok{cyan!3}{ the}
\tok{cyan!2}{ fi}
\tok{cyan!0}{ft}
\tok{cyan!0}{h}
\tok{cyan!4}{ and}
\tok{cyan!0}{ young}
\tok{cyan!1}{est}
\tok{cyan!0}{ child}
\tok{cyan!1}{ of}
\tok{cyan!0}{ well}
\tok{cyan!1}{-}
\tok{cyan!2}{know}
\tok{cyan!3}{n}
\tok{cyan!2}{ teachers}
\tok{cyan!0}{ Bron}
\tok{cyan!1}{i}
\tok{cyan!3}{s\l{}aw}
\tok{cyan!0}{a}
\tok{cyan!1}{,}
\tok{cyan!0}{ }
\tok{cyan!1}{n\'ee}
\tok{cyan!1}{ Bogu}
\tok{cyan!1}{ska}
\tok{cyan!2}{,}} \\
& {\small
\tok{cyan!4}{ and}
\tok{cyan!0}{ W}
\tok{cyan!2}{\l{}a}
\tok{cyan!0}{dy}
\tok{cyan!1}{s\l{}aw}
\tok{cyan!2}{ Sk}
\tok{cyan!0}{\l{}o}
\tok{cyan!0}{d}
\tok{cyan!3}{owski}
\tok{cyan!8}{.}
}
\\
\midrule

\textbf{Answer:} &
{\small
\tok{red!3}{Based}
\tok{violet!7}{ on}
\tok{violet!8}{ the}
\tok{blue!3}{ provided}
\tok{violet!4}{ documents}
\tok{red!4}{,}
\tok{blue!6}{ Marie}
\tok{blue!15}{ Cur}
\tok{blue!3}{ie}
\tok{blue!22}{ was}
\tok{violet!25}{ born}
\tok{violet!30}{ in}
\tok{violet!36}{ Warsaw}
\tok{violet!16}{,}} \\
& {\small
\tok{violet!53}{ in}
\tok{violet!100}{ Congress}
\tok{violet!41}{ Poland}
\tok{violet!74}{ in}
\tok{violet!71}{ the}
\tok{violet!75}{ Russian}
\tok{violet!86}{ Empire}
\tok{violet!17}{ (}
\tok{violet!46}{Document}
\tok{violet!60}{ }
\tok{orange!4}{5}
\tok{violet!11}{).}
}\\[-2pt]
& {\scriptsize\textit{Legend:}\quad
\docchip{blue!70}{Doc. 1}{-5\%}\hspace{3pt}
\docchip{orange!80}{Doc. 2}{-19\%}\hspace{3pt}
\docchip{green!70}{Doc. 3}{-15\%}\hspace{3pt}
\docchip{red!70}{Doc. 4}{-7\%}\hspace{3pt}
\docchip{violet!70}{Doc. 5}{54\%}
}
\\
\bottomrule
\end{tabular}
}};

\draw[Tab0, thick, dashed] (-3.5,2.8) rectangle (1,1.95);
\draw[Tab0, anchor=north west] (1,2.8) node {\begin{minipage}{5.5cm}
    \flushleft\scriptsize Importance of each token inside the query for both retrieval and generation.
\end{minipage}};

\draw[Tab1, thick, dashed] (-3.5,1.85) rectangle (5,-.9);
\draw[Tab1, anchor=north west] (5,1.85) node {\begin{minipage}{1.5cm}
    \flushleft\scriptsize Importance of each token inside the documents for retrieval.
\end{minipage}};

\draw[Tab2, thick, dashed] (-3.5,-1) rectangle (3.7,-2.8);
\draw[Tab2, anchor=south west] (3.7,-2.8) node {\begin{minipage}{2.8cm}
    \flushleft\scriptsize Attribution of generated tokens to the documents. Percentages in brackets are the overall influence of each document on the output.
\end{minipage}};

\end{tikzpicture}
}

\vspace{-5pt}

\caption{}
\label{fig:rag-attribution-example}
\end{subfigure}
\hfill
\begin{subfigure}{0.4\textwidth}
  \centering
  \hspace{-0.15\linewidth}%
  \includegraphics[width=1.15\linewidth]{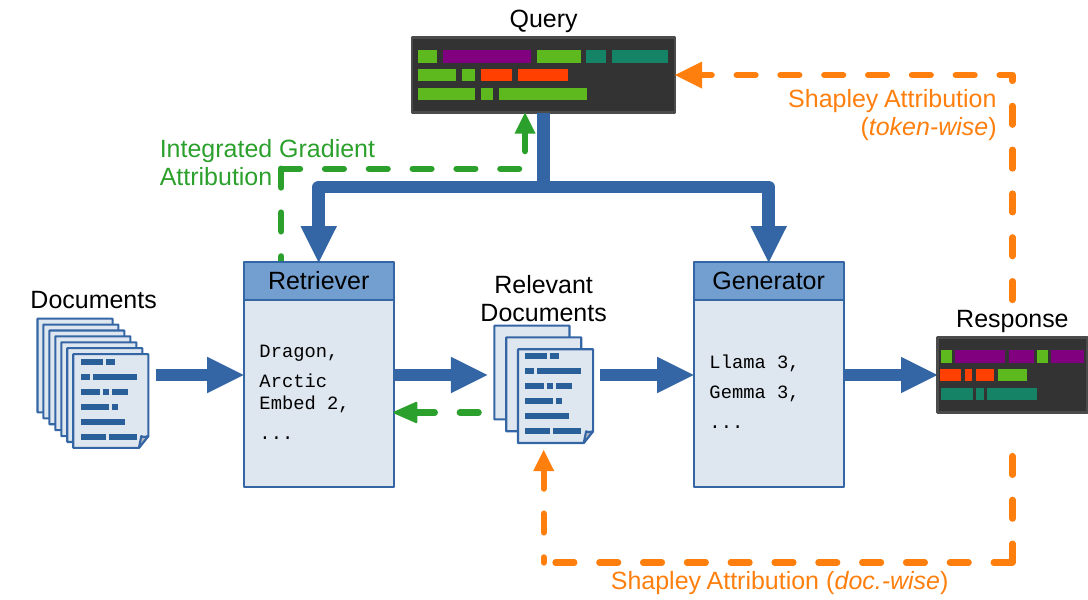}

  \vspace{5pt}

  \caption{}
  \label{fig:overview}
\end{subfigure}

\vspace{-5pt}

\caption[\expandafter\xrag{} overview.]{\expandafter\xrag{} overview. We detect important spans influencing both the retrieval and generation steps.
\textbf{(a):}~Visual example of \xrag{}'s explanations (generated using Arctic Embed 2 and Llama).
\textbf{(b):}~Explanations are based on intrinsic IG~(\textbf{\color{Tab2}- -}) for the \ret{} and extrinsic Shapley for the \gen{} (\textbf{\color{Tab1}- -}).}

\end{figure*}

Retrieval Augmented Generation~\citep[RAG]{Lewis2020_RAG} has become a standard in modern Question-Answering tasks, from the everyday use of ChatGPT to applications in critical domains like medicine and law~\cite{medical-rag-survey, rag-survey}. 
While RAG reduces the opacity of Large Language Models~(LLMs) by grounding its responses in specific sources, its internal reasoning is not inherently transparent; in fact, LLM-generated attribution is statistical in nature and not necessarily faithful~\citep{Randl2025_SelfExplanations}.
In this work, we address this uncertainty by introducing \xrag{}, a mathematically grounded explainability framework for RAG. An example of the explanations provided by our framework is shown in Fig.~\ref{fig:rag-attribution-example}, where we compute token saliency for the \textit{Retriever}~(\ret{}) input and the \textit{Generator}~(\gen{}) query input and document attribution.

Using \xrag{}, we seek to gain a better understanding of RAG.
An empirical analysis, conducted with two \emph{encoder-only} \ret{} models and three \emph{decoder-only} \gen{} models on three open-domain Question-Answering (QA) datasets, suggests that even neural network-based \ret{}s select documents based on the existence of keywords also found in the query.
Furthermore, we find that \gen{} attributions use context documents in a near-binary fashion: relevant documents receive consistently high attribution scores, while irrelevant documents cluster close to zero.

Furthermore, our results indicate no general alignment of \ret{} and \gen{}, but show frequent cases where \gen{} ignores \ret{}'s top-ranked document (Wasted Retrieval), or relies primarily on a document \ret{} attributed with lower relevancy (Noise Distraction). Across our setups, these misalignments can affect over 70\% of the queries for the top-$3$ documents. This reflects that RAG output depends not solely on individual component performance but on their interplay and motivates our research question:

\question{rq:2}{How can the agreement between \ret{} and \gen{} be quantified?}

As RAG systems proliferate in high-stakes domains, addressing misalignment between \ret{} and \gen{} becomes increasingly important.
Prior work on RAG systems typically focuses on explaining either \ret{} or \gen{} in isolation. For example, 
\citet{Zhuang2021_InterpretableRanking,Fernando2019_RetDS} study retriever explanations, while \citet{Qi2024_MIRAGE,Cohen-Wang2024_ContextCite} offer source attribution for the generator (more details in \S\ref{sec:background}); however, there is no metric to quantify how information flows across both components or how they diverge.
We argue that such a metric can help not only improve the transparency of RAG systems but also increase alignment between the two components and possibly reduce computational cost (e.g., by retrieving fewer documents if some will not be used by \gen{}). To address this gap, we propose \xrag{}, an \textbf{end-to-end RAG auditing framework} with two components:

\noindent\textbf{(i)~Attribution Methods Tailored to RAG} We introduce \shap{pmc}, a Monte Carlo stabilized variant of Kernel-SHAP~\citep[\shap{k}]{Lundberg2017_SHAP} that achieves significantly more accurate and reproducible approximations of Shapley Values~\citep[SV]{Shapley1953_Value} for autoregressive \gen{}s.
Furthermore, we establish a baseline embedding for Integrated Gradients~\citep[IG]{Sundarajan2017_IntegratedGradients} on dense retrievers through systematic empirical analysis, showing that replacing non-special tokens with the \texttt{[unk]} embedding significantly outperforms baselines.

\noindent\textbf{(ii)~Metric for \ret{}-\gen{} Alignment} We propose and validate \textit{Weighted Alignment between Retriever and Generator}~(WARG), a novel metric that quantifies how well \gen{}'s use of documents aligns with \ret{}'s ranking. Specifically, we show that WARG is better suited for assessing \ret{}-\gen{} alignment than standard correlation metrics, and that high \ret{}-\gen{} alignment can improve RAG's downstream performance.

The remainder of the paper is structured as follows. We present related work and provide important background (\S\ref{sec:background}), followed by details about the technical implementation of our framework (\S\ref{sec:method}). Then, we present a comprehensive empirical analysis across two \ret{} models, three \gen{} models, and three datasets (\S\ref{sec:empirical-analysis} and \S\ref{sec:results}), demonstrating how RAG-E and WARG can be used to audit RAG systems and quantify alignment. Last, we discuss our results (\S\ref{sec:discussion}) and conclude (\S\ref{sec:conclusions}).
\section{Related Work \& Background}\label{sec:background}
Since the original RAG proposal by \citet{Lewis2020_RAG}, several extensions have been proposed. Examples include Self-RAG \citep{asai2024_selfrag}, which introduces retrieval on demand, and ATLAS \citep{izacard2022_atlas}, which leverages RAG to improve few-shot learning. We focus on the original framework, which serves as a conceptual basis for most RAG approaches.

\paragraph{Related Work}
Early approaches in the literature on explaining RAG focus on the \ret{} component in isolation. \citet{Zhuang2021_InterpretableRanking} propose an \textit{interpretable-by-design} ranking model applicable to a RAG context. We argue, however, that limiting oneself to a single \ret{} architecture is not future-proof.
\citet{Fernando2019_RetDS} evaluate intrinsic and extrinsic explainability methods on a single neural retriever, analyzing the resulting top-attributed terms and observing substantial variability across methods, but without assessing their individual faithfulness.

Recent work on explaining RAG outputs has largely focused on the \gen{} component, proposing algorithmic approaches to source attribution rather than relying on model-generated citations.
ContextCite~\citep{Cohen-Wang2024_ContextCite} attributes generation to context using a linear surrogate model and can be seen as an extension of LIME~\citep{Ribeiro2016_LIME} to RAG.
MIRAGE~\citep{Qi2024_MIRAGE} identifies context-sensitive tokens by measuring changes in generation probabilities under document removal and applies contrastive attribution scores \citep{yin2022_contrastiveexplanations}. Notably, both ContextCite and MIRAGE calculate attribution towards the \emph{entire} output, while our proposed \shap{pmc} attributes to each output token separately.

While the literature recognizes a gap between RAG's \ret{} and \gen{} components, papers focus on methods that adapt \ret{} output to the \gen{}'s preference \citep{ke2024_bridging, ye2024_r2ag}. Notably, GainRAG~\citep{jiang2025gainrag} uses perplexity to quantifiy the \gen{}'s internal preference for a retrieved passage. Contrary to GainRAG, our WARG quantifies misalignment by measuring actual usage of retrieved information by the \gen{}.

\paragraph{Background}
Generally, methods explaining ML models, such as \ret{} and \gen{}, are separated into \emph{intrinsic} methods, deriving explanations based on the internal state of the model~\citep{Sundarajan2017_IntegratedGradients, Shrikumar2017_GradIn, Liu2021_AttentionBasedExplanation}, and \emph{extrinsic} or \emph{model-agnostic} methods, which statistically infer explanations from input-to-output-relationships without considering model internals~\citep{Shapley1953_Value, Lundberg2017_SHAP, Ribeiro2016_LIME}. We argue that intrinsic methods are preferable over extrinsic methods, as their output is directly tied to the explained models' function, while extrinsic methods are only statistically correct. Nevertheless, extrinsic methods are independent of the model's design and therefore more flexible.
See Appendix~\ref{app:background} for extensive background on such methods; we focus purely on \emph{saliency-based}, \emph{local}, and \emph{additive} methods in this paper.

\begin{definition}[Saliency-Based Local Explanation]
    \label{def:locexp}
    Given a Machine Learning~(ML) model $f:\mathbb{R}^{n}\rightarrow\mathbb{R}^{m}$, that maps input vectors ${\bf x} = [x_1, x_2, ..., x_{n}]$ to output vectors ${\bf y}=[y_1, y_2, ..., y_{m}]$, the matrix ${\bf B} \in \mathbb{R}^{n \times m}$ is a local explanation for the specific input-output pair ${\bf \bar{y}} = f({\bf \bar{x}})$, \textbf{iff} its elements $\beta_{i,j}$ describe the impact of feature $\bar{x}_i$ on the output $\bar{y}_j$ for all indices $i \in \{1, 2, ..., n\}$ and $j \in \{1, 2, ..., m\}$.
\end{definition}
\noindent Since the Language Models~(LMs) used in this paper operate on sequences of discrete tokens $\in \mathcal{T}$ instead of real numbers, they rely on an \emph{embedding function}~$\Phi:\mathcal{T}^n\rightarrow\mathbb{R}^n$ and/or a \emph{decoder function}~$\Omega:\mathbb{R}^n\rightarrow\mathcal{T}^n$ for conversion.

\begin{definition}[Additivity]
\label{def:additivity}
A linear feature attribution method adhering to Definition~\ref{def:locexp} is called \textbf{additive}, if, given a baseline input~${\bf x}^0 = [x^0_1, x^0_2, ..., x^0_n]$ and its corresponding model output  ${\bf y}^0 = [y^0_1, y^0_2, ..., y^0_m]$, the sum of the attribution scores $\beta_{i,j}$ adds up to the difference of the model prediction $\bar{y}_j$ and the baseline prediction $y^0_j$:
\begin{equation}
    \left.\bar{y}_j - y^0_j = \sum_{i = 1}^{n} \beta_{i, j}~\right|~\forall j \in \{1, 2, ... , m\}~.
\end{equation}
\end{definition}

\noindent Methods matching Definition~\ref{def:additivity} therefore directly attribute a specific part of the output to each input feature. Some methods require explicitly specifying $({\bf x}^0, {\bf y}^0)$~\citep{Sundarajan2017_IntegratedGradients, Shrikumar2017_GradIn}, while others assume them implicitly~\citep{Shapley1953_Value, Lundberg2017_SHAP}. While in theory any saliency map can be normalized to fulfil the additivity attribute, requiring intrinsic additivity in all our methods is favourable in two respects:
\textbf{(i)} it ensures comparability of all produced saliency maps both in scale and interpretation, and
\textbf{(ii)} the error ratio $\frac{\sum_{i = 1}^{n} \beta_{i, j}}{\bar{y}_j - \bar{y}^0_j}$ can be used as a quality metric for the explanation $\bf B$. 

In summary, existing approaches attempting to explain RAG systems address retrieval and generation in isolation, leaving the interaction between their components unexamined. \xrag{} targets the full pipeline while WARG provides, to the best of our knowledge, the first principled metric for quantifying \ret{}-\gen{} alignment.
\section{Method}\label{sec:method}
Generally, RAG consists of two components:
\textbf{(i)}~A mapping $\text{\ret{}}: ({\textbf{q}}, \mathcal{B}) \rightarrow \mathcal{D}$ that retrieves a \textit{set of relevant documents}~$\mathcal{D} = \{{\bf d}_1,\dots,{\bf d}_k\}$ from a \textit{knowledge base}~$\mathcal{B}$, given a \textit{query}~$\bf q$.
\textbf{(ii)}~A function ${\bf y} = \text{\gen{}}({\bf q}, \mathcal{D})$ that generates an answer to the query based on $\mathcal{D}$.
$\bf q$, ${\bf d}_i$, and $\bf y$ are vectors of tokens $t_j$.

\subsection{The \xrag{} Framework}
We propose an explainability framework for RAG.\footnote{
    \anonymous[%
        The Python module will be available in PyPi after the anonymity phase.
        The source code is available on GitHub: \href{https://anonymous.4open.science/r/RAG-E}{\texttt{https://anonymous.4open.science/r/RAG-E}.}%
    ]{%
        The Python module is available on PyPI and can be installed via \texttt{pip install rag-exp}.
        The source code is available on GitHub under \href{https://github.com/k-randl/Interpretable_RAG}{\texttt{k-randl/Interpretable\_RAG}.}%
    }
}
As can be seen in Fig.~\ref{fig:overview}, the framework relies on separate methods to track saliency through \ret{} and \gen{}.
Specifically, we compute IG attributions on the \ret{}, and SV based attributions on the \gen{} output.
This twofold choice optimizes the trade-off between explanation faithfulness and runtime:
as we focus on purely neural-network-based \ret{}s which are fully differentiable, the choice of an intrinsic method is possible. Furthermore, IG's time-complexity does not scale with the number of input tokens (which can be high for the documents) as is the case with comparable extrinsic methods.
In the face of the increasing architectural variability of LLMs, we opt for an extrinsic method for \gen{}. Given the typically low number of query tokens and context documents, the previously mentioned time-complexity issue is less problematic here.
We provide a brief overview of \xrag{} in the following section. A more in-depth explanation of IG and Shapley methods is available in Appendix~\ref{app:methods}.

\paragraph{\ret{} Explanations: Integrated Gradients}
We compute token-level attributions $\beta_{i,j}^{\text{ret}}$ by applying IG over the token embeddings (Eq.~\ref{eq:ig}), calculating saliency for the query and each retrieved document separately to avoid cross-effects. 
We compare different choices for the baseline embedding~$\Phi^{0}({\bf x})$ in \S\ref{sec:ablations} and use the best candidate: replacing non-special tokens with the model's \texttt{[unk]} token in the rest of the paper. Fig.~\ref{fig:baseline} illustrates this process.

\begin{figure}[t]
    \centering
    \resizetable{\linewidth}{\begin{tabular}{r|cccccccccc}
        \textbf{$\bf x$} &
        \texttt{[sot]} & {\color{red}$t_2$} & {\color{red}$t_3$} & {\color{red}$\dotsc$} & {\color{red}$t_{m'-1}$} & \texttt{[eot]} & \texttt{[pad]} & $\dotsc$ & \texttt{[pad]} \\
    
        & {\color{gray}$\downarrow$} & {\color{gray}$\downarrow$} & {\color{gray}$\downarrow$}  & & {\color{gray}$\downarrow$} & {\color{gray}$\downarrow$} & {\color{gray}$\downarrow$} & & {\color{gray}$\downarrow$} \\
        
        \textbf{$\Phi({\bf x})$} &
        $\phi_1^\text{\texttt{[sot]}}$ & {\color{red}$\phi_2$} & {\color{red}$\phi_3$} & {\color{red}$\dotsc$} & {\color{red}$\phi_{m'- 1}$} & $\phi_{m'}^\text{\texttt{[eot]}}$ & $\phi_{m' + 1}^\text{\texttt{[pad]}}$ & $\dotsc$ &  $\phi_{m}^\text{\texttt{[pad]}}$ \\
    
        & {\color{gray}$\downarrow$} & {\color{gray}$\downarrow$} & {\color{gray}$\downarrow$}  & & {\color{gray}$\downarrow$} & {\color{gray}$\downarrow$} & {\color{gray}$\downarrow$} & & {\color{gray}$\downarrow$} \\
        
        \textbf{$\Phi^0({\bf x})$} &
        $\phi_1^\text{\texttt{[sot]}}$ &  {\color{red}$\phi_2^\text{\texttt{[unk]}}$} & {\color{red}$\phi_3^\text{\texttt{[unk]}}$} & {\color{red}$\dotsc$} & {\color{red}$\phi_{m'-1}^\text{\texttt{[unk]}}$} & $\phi_{m'}^\text{\texttt{[eot]}}$ & $\phi_{m' + 1}^\text{\texttt{[pad]}}$ & $\dotsc$ &  $\phi_{m}^\text{\texttt{[pad]}}$ \\
    \end{tabular}}
    \caption{Baseline creation for IG. We replace the embeddings corresponding to non-special tokens with embeddings corresponding to the model's \texttt{[unk]} token evaluated at the same input position. \texttt{[XXX]} denotes a special token and $\phi_i^\text{\texttt{[XXX]}}$ the corresponding embedding at input position $i$.}
    \label{fig:baseline}
\end{figure}

\paragraph{\gen{} Explanations: Shapley Style Attributions}
We attribute each generated token $t_j$ to each retrieved document ${\bf d}_i$ via SVs, measuring each document's marginal contribution across all subsets $\mathcal{D'} \subseteq \mathcal{D}$. Exact SVs are feasible only for small $|\mathcal{D}|$ due to $\mathcal{O}(2^{|\mathcal{D}|})$ complexity. For larger sets, we use \shap{k}, setting $N = \min(2^{|\mathcal{D}|}, 10 \cdot |\mathcal{D}|)$ samples to balance approximation quality and cost (Fig.~\ref{fig:gen_mc_stab}). Using this choice, \shap{k} also exhibits comparable runtime to state-of-the-art methods (see Appendix~\ref{app:abl_time}). Crucially, we do \textit{not} apply \shap{k}'s marginal contribution step, as retrieved documents frequently share overlapping content, violating the independence assumption underlying that approach.

\paragraph{Monte-Carlo~(MC) Stabilization of \shap{k} Attributions%
}
To reduce the error of the \shap{k} approximation towards precise SV, we evaluate different sampling strategies of perturbed input-output pairs. Specifically, we propose repetitive sampling of \shap{k} in an MC fashion~(referred to as \shap{mc}) and compare it to native \shap{k}. We also try complementary sampling (i.e. the sampling of opposed input pairs $(\mathcal{D'}^1, \mathcal{D'}^2)$, where ${D'}^1 \cap \mathcal{D'}^2 = \emptyset$ and ${D'}^1 \cup \mathcal{D'}^2 = \mathcal{D}$), proposed by~\citet{covert2021_PairedSamling}, for \shap{k} and refer to this as \shap{p}.
For \shap{mc}, we try both paired (i.e. complementary input-output pairs in each MC sample) and random Monte-Carlo sampling. We refer to the paired method as \shap{pmc}.
Algorithm~\ref{alg:mc_shap} shows the precise procedure. Limiting the number of LLM calls to a fixed number~$N$ guarantees that the runtime of this procedure stays comparable to native \shap{k} (see Appendix~\ref{app:abl_time}). As shown in \S\ref{sec:results_design_choices}, \shap{pmc} leads to a significant improvement of the approximation's accuracy at an acceptable improvement of reproducibility.

\begin{algorithm}[t]
\caption{\shap{(p)mc}}\label{alg:mc_shap}
\begin{algorithmic}
    \REQUIRE Query $\bf q$
    \REQUIRE Set of context documents $\mathcal{D}$
    \REQUIRE Number of perturbations $N \le 2^{|\mathcal{D}|}$
    \REQUIRE Number of MC samples $M$
    \REQUIRE Size of MC samples $N' < N$
   
    \STATE $\mathcal{P} \Leftarrow \{\}$\hfill\COMMENT{create perturbations}
    \WHILE{$|\mathcal{P}| < N$}
    \STATE $\mathcal{D'} \Leftarrow \text{take sample} \subseteq \mathcal{D}$\hfill\COMMENT{paired (\shap{pmc}) \\\hfill or random (\shap{mc})}
    \STATE ${\bf x} \Leftarrow \mathrm{create\_prompt}({\bf q},~\mathcal{D'})$
    \STATE ${\bf y} \Leftarrow \mathrm{LLM}({\bf x})$
    \STATE $\mathcal{P} \Leftarrow \{({\bf x},{\bf y})\} \cup \mathcal{P}$
    \ENDWHILE
    \STATE $\mathcal{A} \Leftarrow \{\}$\hfill\COMMENT{sample attributions}
    \WHILE{$|\mathcal{A}| < M$}
    \STATE $\mathcal{A} \Leftarrow \left\{\mathrm{\shap{k}}\left({\mathcal{P'} \subseteq \mathcal{P}}~|~{N' = |\mathcal{P'}|}\right)\right\} \cup \mathcal{A}$
    \ENDWHILE
   
    \STATE \textbf{return} ${1 \over M} \sum \mathcal{A}$\hfill\COMMENT{return average attributions}
\end{algorithmic}
\end{algorithm}

\paragraph{Constrained Token Generation}
Both SV and \shap{k} were developed for classification scenarios where a single call to the ML model produces a single output~${\bf y}$ based on a single input~{\bf x} (i.e. ${\bf y} = f({\bf x})$). However, \gen{} models in RAG are often autoregressive (L)LMs that iteratively complete a sequence of tokens starting from an initial prompt ${\bf x} = [t_{0},\, t_{1},\dots,\, t_{n}]$. In order to keep the \gen{} output comparable for different $\mathcal{D'}$, we first generate the output for the unperturbed set of documents $\mathcal{D}$:
\begin{equation}
    t^{\mathcal{D}}_{i+1} \;=\; \mathrm{LLM}\!\left(\,[
    \,t^{\mathcal{D}}_{1}, \dots, \,t^{\mathcal{D}}_{n},
    ~t^{\mathcal{D}}_{n+1}, \dots, t^{\mathcal{D}}_{i}
    \,]\,\right).
\end{equation}
Then we constrain the generation output for perturbed sets of documents $\mathcal{D'}$ on the previous \textit{original} output and the prompt based on $\mathcal{D'}$:
\begin{equation}
    t^{\mathcal{D'}}_{i+1} \;=\; \mathrm{LLM}\!\left(\,[
    \,t^{\mathcal{D'}}_{1}, \dots, \,t^{\mathcal{D'}}_{n'},
    ~t^{\mathcal{D}}_{n'+1}, \dots, t^{\mathcal{D}}_{i}
    \,]\,\right).
\end{equation}
This constrained generation is proven in literature~\citep{Cohen-Wang2024_ContextCite, Qi2024_MIRAGE}.

\subsection{Quantifying \ret{}-\gen{} Agreement}\label{sec:agreement}
To answer our \questionref{rq:2}, we propose and test the WARG metric. 
Recall that $\beta^\text{ret}_{i,k}$ is a scalar denoting the influence of input token ${\bf t}_i$ on document ${\bf d}_k$, and $\beta^\text{gen}_{k,j}$ a scalar denoting the influence of document ${\bf d}_k$ on output token ${\bf t}_j$. By summing over tokens, $\bar{\beta}^\text{ret}_{k} = \sum_{i} \beta^\text{ret}_{i,k}$ recovers the \ret{} similarity scores under Definition~\ref{def:additivity}, and $\bar{\beta}^\text{gen}_{k} = \sum_{j} \beta^\text{gen}_{k,j}$ gives the influence of document~${\bf d}_k$ on generation.

WARG automatically identifies generation-relevant documents exceeding uniform at random attribution levels via threshold~$\tau$. Letting $\mathcal{K}^+ = \{k \mid \bar{\beta}^\text{gen}_{k} > 0\}$, we normalize attributions and assign binary weights:
\begin{align}
    \tilde{\beta}^\text{gen}_{k} &= \frac{\bar{\beta}^\text{gen}_{k}}
        {\sum_{k' \in \mathcal{K}^+} \bar{\beta}^\text{gen}_{k'}}, \\
    w_k &= \mathbf{I}\!\left(\tilde{\beta}^\text{gen}_{k} > \frac{\tau}{|\mathcal{D}|}\right)
\end{align}
where $\mathbf{I}(\cdot)$ is the indicator function. Here, we use $\tau = 1.5$. Appendix~\ref{app:warg+} verifies that this choice is sensible.

\begin{definition}[Weighted Alignment between \ret{} and \gen{}]
WARG is the normalized difference in mean \ret{} importance between relevant ($w_k=1$) and irrelevant ($w_k=0$) documents:
\begin{equation}
    \mathrm{WARG} = \frac{
        \sum_k \bar{\beta}^\text{ret}_k \cdot \frac{w_k}{h} - \sum_k \bar{\beta}^\text{ret}_k \cdot \frac{1 - w_k}{|\mathcal{D}| - h}
    }{
        \frac{1}{h}\sum_{k = 1}^{h} \bar{\beta}^\text{ret}_k - \frac{1}{|\mathcal{D}| - h}\sum_{k = h+1}^{|\mathcal{D}|} \bar{\beta}^\text{ret}_k
    }
\end{equation}
where $h = \sum_k w_k$, and we assume the scores $\bar{\beta}^\text{ret}_k$ to be ordered from highest to lowest. 
The score is clipped to $[-1,1]$.
\end{definition}
\subsection{Experimental Setup}\label{sec:empirical-analysis}
To evaluate the impact of our framework on different RAG architectures, we consider two encoder-only dense \ret{} models and three open-weight \gen{} models representing the current state-of-the-art in our experiments.
For the \textbf{retrieval phase}, we use 
\textit{DRAGON}~\citep{lin2023_dragon}, a bi-encoder model built upon the BERT-base architecture (110M parameters), and
\textit{Snowflake Arctic Embed 2}~\citep{yu2024arcticembed20multilingualretrieval}, a single encoder model fine-tuned from the multilingual XLM-R Large (568M parameters).
For the \textbf{generation phase}, we employ
\textit{Llama 3.1 8B}~\citep{Grattafiori2024_llama3}, \textit{Qwen 2.5 7B}~\citep{Yang2025_qwen25}, and \textit{Gemma 3 12B}~\citep{Kamath2025_gemma3}, which offer a balance between computational efficiency and reasoning depth.
To save resources, we compress the \gen{} models to \texttt{bfloat16}.

\paragraph{Datasets}
We perform our analysis on the following open-source datasets available for non-commercial use:

\textbf{MS-Marco~v2.1:} a large, classic benchmark for Q\&A tasks \cite{Nguyen2016_MSMarco}. It consists of roughly 1M queries (808k train / 101k dev / 101k test) and 8.8M passages from 3.5M web documents (10 passages per query).
We evaluate on 200 randomly selected samples from the test set, using five random passages for each query.
MS-Marco is available under the \href{https://github.com/microsoft/msmarco/blob/master/LICENSE}{CC BY 4.0} License.

\textbf{PopQA:} a large open-domain question answering benchmark comprising 14k QA pairs derived from Wikidata entities \citep{mallen2023_popqa}.
We evaluate on a random sample of 200 QA pairs, using automatically crawled Wikipedia passages from corresponding entity pages as context documents.
PopQA's queries are available under an \href{https://github.com/AlexTMallen/adaptive-retrieval/blob/main/LICENSE}{MIT License}, while the automatically retrieved Wikipedia passages fall under \href{https://en.wikipedia.org/wiki/Wikipedia:Text_of_the_Creative_Commons_Attribution-ShareAlike_4.0_International_License}{CC BY 4.0}.

\textbf{QAMPARI:} a multi-answer open-domain QA benchmark comprising questions that require retrieving and aggregating multiple answers from different passages~\citep{amouyal-etal-2023-qampari}. It consists of roughly 63k generated examples (61k train / 1k dev / 1k test), including passages based on Wikipedia.
We evaluate on a random sample of 100 questions, each with at least five human-validated answers, using Wikipedia passages as context.
QAMPARI is released under \href{https://github.com/samsam3232/qampari/blob/master/LICENSE}{CC 0}.
\textbf{TREC CAsT 2019~(TC):} a conversational IR benchmark (38.6M texts) composed of MS-MARCO~\citep{Nguyen2016_MSMarco}, TREC CAR~\citep{dietz_trec_2017}, and the Washington Post corpus\footnote{\href{https://trec.nist.gov/data/wapost/}{https://trec.nist.gov/data/wapost/}}, with evaluation topics and human relevance judgments.
We evaluate a random sample of 100 questions, each with at least ten human-validated answers.
The TC queries are available under an \href{https://github.com/daltonj/treccastweb/blob/master/LICENSE}{MIT License}.

\begin{table}[t]
\centering
\setlength{\tabcolsep}{1pt}
\renewcommand{\arraystretch}{1.5}
\resizetable{\linewidth}{
\begin{tabular}{lcrlcrlcrl}
\toprule
\multirow{2}{*}{\textbf{Dataset}}
& \hspace{5pt} & \multicolumn{2}{c}{\multirow{2}{*}{\textbf{Queries}}} 
& \hspace{5pt} & \multicolumn{5}{c}{\textbf{Context Documents}} \\[-5pt]

&&&&& \multicolumn{2}{c}{\scriptsize DRAGON} 
& \hspace{5pt} & \multicolumn{2}{c}{\scriptsize Arctic Embed 2} \\[0pt]
\midrule

MSMarco
&&    6.4 & \scriptsize [  4.8,    8.0]
&&   55.8 & \scriptsize [ 42.0,   64.0]
&&   55.8 & \scriptsize [ 42.0,   64.0] \\
PopQA
&&    6.7 & \scriptsize [  6.0,    7.0]
&&   49.6 & \scriptsize [ 40.0,   59.0]
&&   48.7 & \scriptsize [ 38.0,   58.0] \\
Qampari
&&    7.5 & \scriptsize [  6.0,    8.0]
&&   90.1 & \scriptsize [ 82.0,  105.0]
&&   88.2 & \scriptsize [ 79.8,  104.0] \\
TC
&&    7.2 & \scriptsize [  4.0,   10.0]
&&   65.8 & \scriptsize [ 33.0,  105.0]
&&   64.9 & \scriptsize [ 32.1,  104.9] \\
\bottomrule
\end{tabular}
}
\caption{Mean word count of data after retrieval. Squared brackets contain the 25\% and 75\% percentiles of word counts.}
\label{tab:data_stats}
\end{table}

Table~\ref{tab:data_stats} shows word count statistics for the used datasets. We present mean word count, as well as 25\% and 75\% percentiles after retrieval to highlight the impact of \ret{} choice. Note that for MSMarco we have $\mathcal{B} = \mathcal{D}$, effectively using the \ret{} for reranking only.

\paragraph{Analysis of Design Choices}\label{sec:ablations}
To verify the faithfulness of our approach and the validity of our design choices, we conduct small experiments on 200 randomly selected samples from the \textbf{MS-Marco~v2.1} Q\&A dataset~\citep{Nguyen2016_MSMarco}. We use $\mathcal{B} = \mathcal{D} = 5$ random context documents for each query.
Experiments are run on 8 NVIDIA RTX A5500 GPUs, each with 24GB VRAM.

We measure faithfulness as the Area Inside the Perturbation Curves~(AIPC) via input perturbation. Since the query texts are naturally short, and perturbing a single token can easily distort the meaning of the whole text independent of the token's impact on the decision, we test faithfulness only on the context documents. However, as the applied methods are analogous for both contexts and queries, we argue that the results are transferable. Appendix~\ref{app:abl_faith} details the process.

\paragraph{Main Experiments}
For each dataset, we build flat FAISS~\citep{douze2024faiss} indexes over document embeddings and present documents to the \gen{} under four prompt configurations:
\textbf{(\textsc{c1})}~the top-10 retrieved documents in rank order,
\textbf{(\textsc{c2})}~the top-5 retrieved documents followed by 5 randomly sampled non-retrieved documents,
\textbf{(\textsc{c3})}~5 randomly sampled non-retrieved documents followed by the top-5 retrieved documents, and
\textbf{(\textsc{c4})}~the top-10 retrieved documents in randomized order,
allowing us to analyze \gen{} sensitivity to document position and ordering.
For configurations \textsc{c2} to \textsc{c4}, we assume that \ret{} importance decreases linearly with document rank (i.e., rank 1 has the highest \ret{} importance and rank 10 the lowest). This assumption mimics the retrieval scores obtained for \textsc{c1}.
The experiments conducted on these datasets are performed on an Intel(R) Xeon(R) Platinum 8480CL with 8 NVIDIA H100, each with 80GB of memory.
\section{Results}\label{sec:results}
We report the results of our experiments in the following three sections: \textbf{\S\ref{sec:results_design_choices}} motivates the design choices of our proposed \xrag{} framework; \textbf{\S\ref{sec:results_rq1}} presents an exploratory analysis of RAG explanations using RAG-E; and \textbf{\S\ref{sec:results_rq2}} empirically evaluates the utility of our WARG metric addressing our \questionref{rq:2}.

\subsection{Empirical Analysis of Design Choices}\label{sec:results_design_choices}

\begin{table}[b]
\centering
\resizetable{\linewidth}{
\begin{tabular}{l|cccc}
\toprule
\multirow{2}{*}{\textbf{\ret{}}} &
\multicolumn{4}{c}{\textbf{Baselines ($\Phi^0$)}} \\
&
\footnotesize{$\mathbf{0}$} &
\footnotesize{\texttt{[mask]}} &
\footnotesize{\texttt{[unk]}} &
\footnotesize{\texttt{[pad]}} \\
\midrule
\multirow{2}{*}{DRAGON} &
0.45 &
0.46 &
\textbf{0.50} &
0.41 \\
&
\footnotesize [0.44, 0.46] &
\footnotesize [0.45, 0.47] &
\textbf{\footnotesize [0.49, 0.51]} &
\footnotesize [0.39, 0.42] \\
\multirow{2}{*}{Arctic Embed 2} &
0.68 &
0.61 &
\textbf{0.73} &
0.67 \\
&
\footnotesize [0.66, 0.71] &
\footnotesize [0.59, 0.64] &
\textbf{\footnotesize [0.70, 0.76]} &
\footnotesize [0.65, 0.69] \\
\bottomrule
\end{tabular}
}
\caption{Faithfulness in terms of AIPC (\textit{higher is better}) for different baselines $\Phi^0$ of IG (with $L{=}100$). Bold scores mark the most faithful baseline per \ret{}. 95\% confidence intervals (over 1000 bootstrap samples) are reported as [lower, upper].}
\label{tab:faith_ret_ig_base}
\end{table}

\paragraph{Retriever Design Choices}
To select a useful baseline for the IG-based saliency values for the \ret{}, we compare replacing the embeddings of non-special tokens of the transformer input with the following values (see Fig.~\ref{fig:baseline}):
\textbf{(i)}~zeros (discarding the positional embeddings), 
\textbf{(ii)}~the \texttt{[mask]} token,
\textbf{(iii)}~the \texttt{[pad]} token (suggested by the IG paper), and
\textbf{(iv)}~the \texttt{[unk]} token, all embedded at the input position.
As shown in Tab.~\ref{tab:faith_ret_ig_base}, a baseline replacing non-special tokens with the model's \texttt{[unk]} token clearly outperforms the other choices. While the original IG paper proposes using the \texttt{[pad]} token for masking \cite{Sundarajan2017_IntegratedGradients}, our results suggest that the choice of baseline token matters. Intuitively, the \texttt{[pad]} token encodes empty space outside the text, whereas the \texttt{[unk]} token encourages the model to infer meaning during inference.

Further tests of IG on the \ret{} component are reported in Appendix~\ref{app:abl_faith}. Specifically, we find that using $L=100$ integration steps sufficiently approximates the integral and that IG is more faithful than other explainability methods. Based on these findings, we apply IG with a \texttt{[unk]} baseline and $L=100$ on \ret{}.

\begin{figure}[t]
\centering
\includegraphics[width=\linewidth]{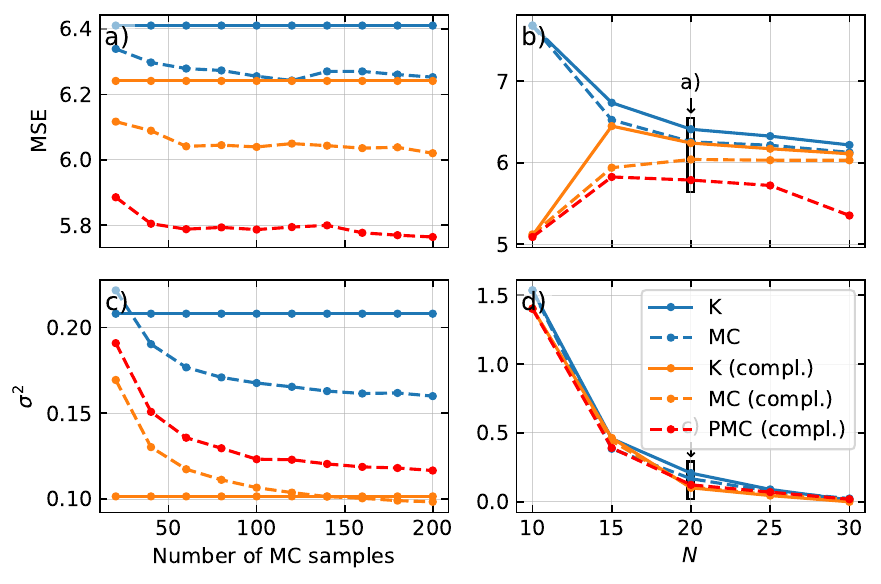}
\caption{MSE compared to precise SV [\textit{plots a})\textit{ \& b})] and variance~$\sigma^2$ over 10 repetitions [\textit{plots c})\textit{ \& d})] of \shap{k}, \shap{mc}, and \shap{pmc} for $|\mathcal{D}| = 5$ and $N' = 10$ (\textit{Left~column:} results for $N=20$; \textit{Right~column:} results for 200 MC samples).}
\label{fig:gen_mc_stab}
\end{figure}

\paragraph{Generator Design Choices}
Fig.~\ref{fig:gen_mc_stab} presents the approximation error (i.e., MSE), and reproducibility (i.e., variance~$\sigma^2$) of native \shap{k}, \shap{mc}, and our proposed \shap{pmc}.
While the highest stability is achieved by \shap{k} with complementary sampling, we find \shap{pmc} the better choice, as it achieves the best approximation of SV while displaying sufficiently low variance, especially for high numbers of MC samples and a medium $N$.

Further experiments reported in Appendix~\ref{app:abl_shap_mc} show that \shap{(p)mc} approximations are statistically significantly closer to the true SV than native \shap{k} under identical sampling conditions. Our results in the Appendix further show that LIME and all SHAP-based methods perform comparably in terms of AIPC (Appendix~\ref{app:abl_faith}). Nevertheless, they drastically improve on ContextCite and MIRAGE, as the latter disregard output-token-specific attributions.
In conclusion, we apply \shap{pmc} with 200 MC samples to explain \gen{}.

\subsection{Exploratory Analysis of RAG Attributions}\label{sec:results_rq1}
\paragraph{Qualitative Analysis} We conduct a qualitative analysis of \ret{} and \gen{} attributions in Appendix~\ref{app:showcase}. Overall, we find two emerging patterns: 
\textbf{(i)}~even dense \ret{} rely on keyword-matching to retrieve documents and
\textbf{(ii)}~\gen{} document attributions follow a binary pattern, strictly distinguishing relevant documents with high attribution scores from irrelevant documents at attribution scores around $\frac{1}{|\mathcal{D}|}$ and lower.

\paragraph{Retrieval Analysis}
To further investigate our qualitative finding that \ret{}s rely on keyword matching, we perform quantitative analysis on token-level attributions of the query and context documents and report the results in the following\footnote{We report retrieval performance in Appendix~\ref{app:ret_performance}.}.

We conduct a Part-of-Speech~(POS)-based grammatical analysis of queries and documents to examine how different POS influence the retrieval phase.
Our analysis reveals that the retrieval mechanism heavily relies on nouns~(NOUN) and proper nouns~(PROPN), which together make up more than a $70\%$ of the top-30\% attributed words for Arctic Embed 2 and around $50\%$ for DRAGON. In comparison, other POS tags usually make up around 10\% or less each (Appendix \ref{app:pos_analysis}). This indicates that \ret{} matches entities and key concepts rather than structural or function words.

We also investigate the role of exact term matching. Our analysis reveals a strong overlap between the query words and the top-attributed document words on PopQA. Between \textbf{29.68\%}~(Arctic Embed 2) and \textbf{99.11\%}~(DRAGON) of the query words appear among the top~50\% attributed tokens in the documents. These results reinforce our claim that despite the use of dense embeddings, the models still strongly rely on keyword overlap between queries and documents.

\paragraph{Generation Analysis}
\begin{figure}
    \centering
    \begin{tabular}{cp{0.97\linewidth}}
        \rotatebox{90}{~~~~~~~\small\textsf{PopQA}} & \includegraphics[trim={0 33pt 0 25pt},clip,width=\linewidth]{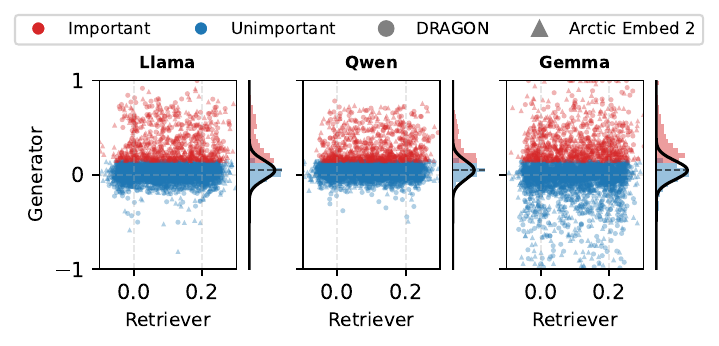} \\
        \rotatebox{90}{~~~~\small\textsf{QAMPARI}} & \includegraphics[trim={0 33pt 0 36.5pt},clip,width=\linewidth]{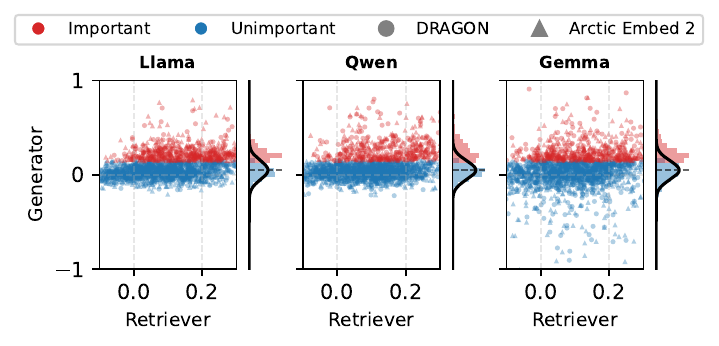} \\
        \rotatebox{90}{~~~~~~~~~~~~~~~~\small\textsf{TC}} & \includegraphics[trim={0 5pt 0 36.5pt},clip,width=\linewidth]{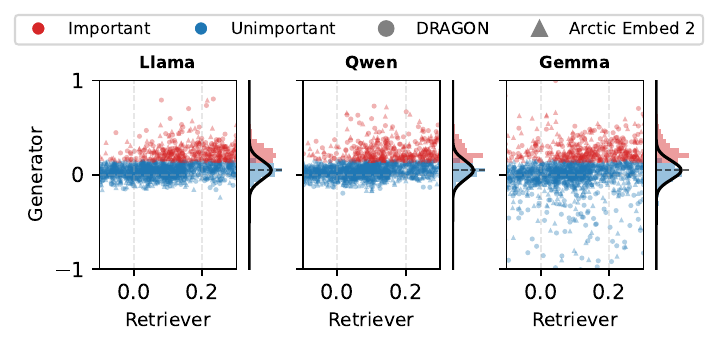} \\
    \end{tabular}
    \caption{Scatterplot of \ret{} importance (\textit{x-axis}) versus \gen{} importance (\textit{y-axis}, normalized such that positive values sum to 1). Documents exceeding the WARG threshold of $\tau=1.5$ are shown in red; those below are shown in blue (DRAGON:~$\circ$, Arctic Embed~2:~$\triangle$). The histogram on the right of each plot shows the distribution of \gen{} importance scores, overlaid with a normal distribution ($\mu=0.05, \sigma=0.1$).}
    \label{fig:gen_scatter}
\end{figure}
To verify our qualitative findings regarding the binary nature of \gen{} attributions, we analyse the distribution of document importances. Figure~\ref{fig:gen_scatter} shows that the main body of documents is approximately normally distributed at a mean of $\mu=\frac{1}{2|\mathcal{D}|}$ in terms of \gen{} document importance. While high \gen{} scores are slightly more probable at \ret{} document importance $> 0$ for QAMPARI and TC, \ret{} and \gen{} attributions are generally not meaningfully correlated. A selection of $\tau=\mu+\sigma=1.5$ meaningfully distinguishes important documents, appearing as positive outliers on the \gen{} axis of the plot from the normal distribution of irrelevant documents.

\paragraph{Failure Mode Analysis}
To analyse consequences of the different inherent behaviours of \ret{} and \gen{}, indicated by the above results, we quantify two specific failure modes:
\textbf{(i)~Wasted Retrieval} occurs when the top-ranked retrieved document (\ret{}~Rank 0) receives \gen{}~Rank $> k$, indicating that the \ret{}’s most relevant document was largely ignored during generation.
\textbf{(ii)~Noise Distraction} occurs when \gen{} assigns its highest importance (\gen{}~Rank 0) to a document with \ret{}~Rank $> k$, suggesting that the model was distracted by content deemed less relevant by \ret{}.
The results are shown in Table~\ref{tab:failure_modes}.

Considering the top-3 ranked documents ($k=3$), Wasted Retrieval ranges from 16.0\% (DRAGON + Qwen on QAMPARI) to 77.5\% (DRAGON + Llama on PopQA), and Noise Distraction from 15.0\% (DRAGON + Qwen on QAMPARI) to 83.0\% (Arctic Embed 2 + Llama on PopQA). The lower end of these ranges shows that certain \ret{}-\gen{} combinations perform reasonably on specific datasets, but the high upper bounds and the circumstance that low values seem to be highly dependent on the data indicate a mismatch between the document importance attributed by \ret{} and that assigned by \gen{}. This motivates further analysis into the quantification of the disagreement using our proposed WARG.

\begin{table}[t]
\centering
\setlength{\tabcolsep}{3pt}
\renewcommand{\arraystretch}{1.05}
\small
\resizebox{\linewidth}{!}{%
\begin{tabular}{lll|ccc|ccc}
\toprule
\multirow{2}{*}{\textbf{Dataset}} & \multirow{2}{*}{\textbf{\ret{}}} & \multirow{2}{*}{\textbf{\gen{}}} & \multicolumn{3}{c|}{\textbf{Wasted Retrieval} [\%]} & \multicolumn{3}{c}{\textbf{Noise Distraction} [\%]} \\
\footnotesize & \footnotesize & \footnotesize &
\footnotesize ($k=1$) & \footnotesize ($k=3$) & \footnotesize ($k=5$) & 
\footnotesize ($k=1$) & \footnotesize ($k=3$) & \footnotesize ($k=5$) \\
\midrule
\multirow{6}{*}{PopQA} & \multirow{3}{*}{DRAGON}
  & Llama & 87.0 & 77.5 & 63.5 & 87.0 & 76.0 & 65.0 \\
& & Qwen  & 90.0 & 72.0 & 60.0 & 88.0 & 79.0 & 67.5 \\
& & Gemma & 52.0 & 40.5 & 32.5 & 54.5 & 38.5 & 31.5 \\
\cline{2-9}
& \multirow{2}{*}{Arctic}
  & Llama & 90.0 & 76.5 & 63.5 & 92.0 & 83.0 & 73.5 \\
& \multirow{2}{*}{Embed 2}
  & Qwen  & 86.5 & 74.5 & 60.5 & 88.0 & 79.5 & 72.0 \\
& & Gemma & 53.5 & 41.5 & 34.0 & 50.5 & 31.5 & 24.0 \\
\midrule
\multirow{6}{*}{TC} & \multirow{3}{*}{DRAGON}
  & Llama & 73.0 & 40.0 & 19.0 & 73.0 & 36.0 & 12.0 \\
& & Qwen  & 66.0 & 32.0 & 21.0 & 66.0 & 36.0 & 13.0 \\
& & Gemma & 78.0 & 47.0 & 28.0 & 78.0 & 48.0 & 25.0 \\
\cline{2-9}
& \multirow{2}{*}{Arctic}
  & Llama & 72.0 & 34.0 & 24.0 & 72.0 & 30.0 & 12.0 \\
& \multirow{2}{*}{Embed 2}
  & Qwen  & 67.0 & 35.0 & 14.0 & 67.0 & 33.0 &  9.0 \\
& & Gemma & 70.0 & 45.0 & 26.0 & 70.0 & 40.0 & 22.0 \\
\midrule
\multirow{6}{*}{QAMPARI} & \multirow{3}{*}{DRAGON}
  & Llama & 50.0 & 23.0 &  9.0 & 50.0 & 22.0 &  2.0 \\
& & Qwen  & 47.0 & 16.0 &  9.0 & 47.0 & 15.0 &  2.0 \\
& & Gemma & 65.0 & 41.0 & 32.0 & 65.0 & 38.0 & 21.0 \\
\cline{2-9}
& \multirow{2}{*}{Arctic}
  & Llama & 61.0 & 28.0 &  6.0 & 61.0 & 29.0 &  2.0 \\
& \multirow{2}{*}{Embed 2}
  & Qwen  & 56.0 & 21.0 &  7.0 & 56.0 & 23.0 & 13.0 \\
& & Gemma & 63.0 & 38.0 & 24.0 & 63.0 & 44.0 & 21.0 \\
\bottomrule
\end{tabular}%
}
\caption{Wasted Retrieval and Noise Distraction across datasets and models under \textsc{c1}.}
\label{tab:failure_modes}
\end{table}

\subsection{Verification of the WARG Metric}\label{sec:results_rq2}
To evaluate WARG's ability to quantify the alignment between \ret{} and \gen{} relevance signals (our \textbf{\questionref{rq:2}}), we compare it against standard Spearman and Pearson correlations.
As shown in Table~\ref{tab:agreement}, WARG more reliably distinguishes \ret{}-\gen{} agreement from disagreement than the baselines. Under prompt condition \textsc{c2}, where top-ranked documents appear first in the prompt, WARG achieves distinctly higher values than Spearman and Pearson. Conversely, under condition \textsc{c3}, where top-ranked documents appear late in the prompt, WARG achieves consistently lower values. Under condition \textsc{c4}, all metrics yield mean values comparably close to 0, indicating no systematic pattern.
We attribute WARG's superiority to three factors:
\textbf{(i)}~its better suitability for capturing the \gen{}'s tendency to focus on a subset of salient documents, and
\textbf{(ii)}~Spearman's susceptibility to noise in tailed distributions due to its rank-transformation step.
\textbf{(iii)}~Pearson's assumption of linear correlations, which is not compatible with the near-binary structure of \gen{} importance (relevant vs. irrelevant documents).

\begin{table}[t]
\centering
\setlength{\tabcolsep}{2.5pt}
\resizetable{\linewidth}{\begin{tabular}{ll|ccc|ccc|ccc}
\toprule
\multirow{3}{*}{\textbf{\ret{}}} &
\multirow{3}{*}{\textbf{\gen{}}} &
\multicolumn{3}{c|}{\textbf{\textsc{c2}}} &
\multicolumn{3}{c|}{\textbf{\textsc{c3}}} &
\multicolumn{3}{c}{\textbf{\textsc{c4}}} \\
\footnotesize & \footnotesize &
\multicolumn{3}{c|}{(top-k first)} &
\multicolumn{3}{c|}{(random first)} &
\multicolumn{3}{c}{(shuffled)} \\
\footnotesize & \footnotesize &
\footnotesize Sprmn. & \footnotesize Prsn. & \footnotesize WARG &
\footnotesize Sprmn. & \footnotesize Prsn. & \footnotesize WARG &
\footnotesize Sprmn. & \footnotesize Prsn. & \footnotesize WARG \\

\midrule
\midrule
\multicolumn{8}{l}{\footnotesize\textbf{PopQA}} \\
\midrule
\multirow{3}{*}{DRAGON}
& Llama & 0.50 & 0.55 & \textbf{0.74} & -0.26 & -0.26 & \textbf{-0.34} & \textbf{0.05} & \textbf{0.05} & \textbf{0.05} \\
& Qwen & 0.48 & 0.53 & \textbf{0.73} & -0.09 & -0.15 & \textbf{-0.30} & 0.15 & 0.12 & \textbf{0.09} \\
& Gemma & 0.34 & 0.39 & \textbf{0.59} & -0.06 & -0.09 & \textbf{-0.18} & 0.10 & \textbf{0.06} & 0.08 \\
\hline
\multirow{2}{*}{Arctic}
& Llama & 0.51 & 0.54 & \textbf{0.73} & -0.26 & -0.27 & \textbf{-0.36} & 0.05 & 0.02 & \textbf{0.00} \\
\multirow{2}{*}{Embed 2}
& Qwen & 0.49 & 0.55 & \textbf{0.76} & -0.05 & -0.13 & \textbf{-0.30} & 0.13 & 0.07 & \textbf{0.01} \\
& Gemma & 0.34 & 0.39 & \textbf{0.55} & -0.08 & -0.11 & \textbf{-0.20} & 0.07 & 0.04 & \textbf{0.01} \\

\midrule
\midrule
\multicolumn{8}{l}{\footnotesize\textbf{TC}} \\
\midrule
\multirow{3}{*}{DRAGON} & Llama & 0.42 & 0.44 & \textbf{0.53} & -0.10 & -0.12 & \textbf{-0.15} & 0.11 & 0.12 & \textbf{0.11} \\
 & Qwen & 0.44 & 0.46 & \textbf{0.51} & -0.12 & -0.15 & \textbf{-0.25} & 0.24 & \textbf{0.23} & 0.27 \\
 & Gemma & 0.30 & 0.30 & \textbf{0.40} & -0.10 & -0.12 & \textbf{-0.16} & \textbf{0.08} & 0.09 & 0.14 \\
\hline
\multirow{2}{*}{Arctic} & Llama & 0.43 & 0.45 & \textbf{0.57} & -0.12 & -0.14 & \textbf{-0.17} & \textbf{0.09} & 0.11 & 0.12 \\
\multirow{2}{*}{Embed 2} & Qwen & 0.52 & 0.53 & \textbf{0.65} & -0.13 & -0.14 & \textbf{-0.20} & 0.24 & \textbf{0.23} & 0.25 \\
 & Gemma & 0.39 & 0.39 & \textbf{0.45} & -0.14 & -0.15 & \textbf{-0.19} & \textbf{0.10} & 0.11 & 0.13 \\

\midrule
\midrule
\multicolumn{8}{l}{\footnotesize\textbf{Qampari}} \\
\midrule
\multirow{3}{*}{DRAGON}
& Llama & 0.65 & 0.66 & \textbf{0.73} & -0.46 & -0.45 & \textbf{-0.49} & \textbf{0.09} & 0.10 & 0.10 \\
      & Qwen & 0.54 & 0.58 & \textbf{0.74} & -0.23 & -0.26 & \textbf{-0.32} & 0.12 & 0.13 & \textbf{0.09} \\
      & Gemma & 0.33 & 0.35 & \textbf{0.44} & -0.16 & -0.16 & \textbf{-0.20} & -0.02 & \textbf{-0.00} & 0.05 \\
     \hline
     \multirow{2}{*}{Arctic} & Llama & 0.56 & 0.59 & \textbf{0.68} & -0.35 & -0.35 & \textbf{-0.36} & 0.05 & 0.06 & \textbf{0.01} \\
     \multirow{2}{*}{Embed 2} & Qwen & 0.43 & 0.50 & \textbf{0.69} & -0.23 & -0.26 & \textbf{-0.35} & 0.07 & \textbf{0.06} & 0.08 \\
      & Gemma & 0.34 & 0.37 & \textbf{0.48} & -0.19 & -0.18 & \textbf{-0.24} & 0.08 & \textbf{0.07} & 0.09 \\

\bottomrule
\end{tabular}%
}
\caption{Mean Spearman~$r$, Pearson~$r$, and WARG per \ret{}-\gen{} combination under conditions \textsc{c2} (\textit{higher is better}), \textsc{c3} (\textit{lower is better}), and \textsc{c4} (\textit{closer to 0 is better}). Best results per condition are in bold.}
\label{tab:agreement}
\end{table}

\begin{figure}
\centering
\includegraphics[width=\linewidth,trim={0 5pt 0 5pt},clip]{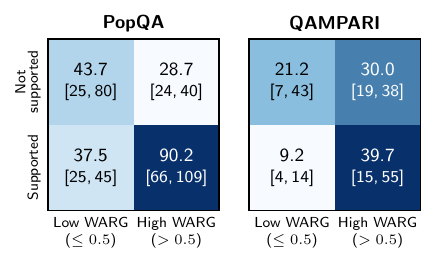}
\caption{Confusion matrix of WARG score against lexical support of the ground-truth labels by the generated answer, on PopQA (\textit{left}) and QAMPARI (\textit{right}). Cell values report the mean number of queries per group, with minimum and maximum across all \ret{}/\gen{} combinations shown in brackets.}
\label{fig:warg_confusion_combined}
\end{figure}

In addition to the above, we show that there is a connection between WARG and downstream performance. Figure~\ref{fig:warg_confusion_combined} presents results on PopQA and QAMPARI \footnote{We do not include TC here because it provides passage-level relevance judgments rather than extractive gold answers, so lexical support of a ground-truth label by the generation is undefined; TC is instead used for our ordering-based WARG validation (Table~\ref{tab:agreement}).}, showing that when a generated answer contains a ground-truth label, it is more than twice as likely (PopQA) and more than four times as likely (QAMPARI) to achieve a high WARG score than not. A $\chi^2$ test of independence confirms the association on both datasets (PopQA: $\chi^2=16.5$, $p<10^{-4}$; QAMPARI: $\chi^2=35.2$, $p<10^{-8}$). While no mathematical guarantee exists that a high WARG score implies a correct answer, this finding is intuitive: when the \ret{} and \gen{} align well, the relevant documents appear early in the context, and the \gen{} has sufficient information to answer correctly. We provide a more detailed version of these results in Appendix~\ref{app:warg+}.

\section{Discussion}\label{sec:discussion}
\paragraph{IG and the \texttt{[unk]}-token} Although IG and \shap{k} are established in the literature, we demonstrate that our adaptations meaningfully increase their utility for RAG pipelines. Notably, we find that the \texttt{[unk]} token is best suited for masking when constructing IG baselines, which contradicts the original IG paper's recommendation for \texttt{[pad]} \cite{Sundarajan2017_IntegratedGradients}. This is intuitive, because \texttt{[pad]} encodes empty space beyond the text boundary, whereas \texttt{[unk]} prompts the model to actively infer meaning, making it a more appropriate baseline for capturing token-level contributions.

\paragraph{The Relation between \ret{} and \gen{}} A grammatical analysis (Appendix~\ref{app:pos_analysis}) revealed that dense \ret{} models primarily rely on content-bearing nouns, with limited sensitivity to adpositions and syntactic connectors. This behaviour closely resembles keyword matching between queries and documents.
What we also observe, however, is that the tested generators successfully avoid primacy bias in short contexts, distributing attributions based on document content rather than prompt position when the input order is randomized (Appendix~\ref{app:primacy_bias}). While this contradicts literature~\citep{Liu2024_lostinthemiddle}, we assume that, due to our relatively small context sizes, the prompts are short enough for the LLMs to capture them entirely.
Beyond these component-level insights, our analyses confirm a systematic misalignment between \ret{} and \gen{} models, also seen in the literature~\citep{ye2024_r2ag,ke2024_bridging,jiang2025gainrag}, and motivates our WARG metric.
We find that \gen{} models use context documents in a near-binary fashion: relevant documents receive consistently high attribution scores well above the chance level $\frac{1}{|\mathcal{D}|}$, while irrelevant documents cluster close to $\frac{1}{2|\mathcal{D}|}$. WARG leverages this finding and therefore improves on standard correlation metrics.

\section{Conclusions}\label{sec:conclusions}
We presented \xrag, a novel framework for end-to-end explainability of RAG pipelines.
To that end, we adapted gradient-based IG to explain \ret{}s by establishing suitable baseline embeddings, and introduced \shap{pmc}, a stabilized variant of \shap{k} that enables reliable attribution for autoregressive \gen{} models. 
\xrag{} allows us to audit RAG systems, here comparing two \ret{} models, \textit{DRAGON} and \textit{Arctic Embed~2}, and three \gen{} models, \textit{Llama~3~8B}, \textit{Qwen~2.5~7B} and \textit{Gemma~3~12B}, to understand which parts of the inputs and outputs are most influential for \ret{} and \gen{}.
Answering our \questionref{rq:2}, we also proposed WARG, a novel metric for quantifying misalignment of the \ret{} and \gen{} components, high values of which can serve as an indication of correct answers on the downstream task.
Future directions include scalability, extending \xrag{} and WARG to include rerankers, using WARG as an optimization signal in query rewriting or reinforcement learning-based RAG systems.

\section*{Limitations}
While RAG-E provides a comprehensive framework for auditing RAG systems, several limitations remain.

\paragraph{Time complexity of \gen{} attributions} First, the computational cost of Monte Carlo-stabilized Shapley values (\shap{pmc}), although more stable than previous approximations, still scales with the number of context documents~$|\mathcal{D}|$. Even at $N=\min(2^{|\mathcal{D}|}, 10 \cdot |\mathcal{D}|)$, as proposed in the paper, time complexity may be problematic, potentially limiting its use for real-time monitoring of very large pipelines. 
While this is already an improvement over precise SV (scaling linearly instead of exponentially with $|\mathcal{D}|$), and other SOTA methods show comparable runtime (see Appendix~\ref{app:abl_time}), we see future work on reducing time complexity, for example, by adaptively reducing the number of features, as necessary. 

\paragraph{Limitation to standard RAG pipelines} Second, our current analysis focuses on standard RAG, based on dense \ret{}s and autoregressive \gen{}s. While retrievers and rerankers are very similar in design, theoretically rendering WARG and \xrag{} directly applicable, the framework's applicability to hybrid retrieval systems, systems including rerankers, or non-autoregressive \gen{} models requires further validation. Future evaluation could also address adaptation to other RAG derivatives such as agentic-RAG systems.

\paragraph{Context length} Third, we limit our evaluation to relatively short documents (50 to 90 words on average). This helps keep prompt size concise, avoiding primacy bias (see Appendix~\ref{app:primacy_bias}), and inference time low. Nevertheless, this choice could have implications for the LLMs' ability to trace information. Future work must therefore examine how RAG-E and WARG transfer to long-document-settings.
\section*{Ethics Statement}
RAG systems are increasingly deployed in applications that support decision-making in high-stakes domains such as healthcare and law \citep{medical-rag-survey,pipitone2024_RAGlegal}. This work contributes tools for auditing RAG pipelines by making the interaction between retrieval and generation explicit and measurable. By quantifying \ret{}–\gen{} misalignment and identifying systematic failure modes, our framework can support more reliable system design, post-hoc analysis, and informed human oversight.

At the same time, we are aware that the methods proposed in this paper rely on a certain level of technical understanding to interpret. Untrained users may misinterpret results, and, for example, confuse high WARG with direct evidence of factual correctness or truthfulness. Misuse of such metrics could lead to overconfidence in aligned but incorrect answers.

Overall, we view this work as enabling more transparent and accountable use of RAG systems, while recognizing that explainability tools must be applied judiciously and in conjunction with domain expertise and external validation.

\paragraph{Disclaimer} We used generative AI for spellchecking and improvement of short text passages during the writing of this manuscript.
\section*{Acknowledgements}
This work has been partially supported by project MIS 5154714 of the National Recovery and Resilience Plan Greece 2.0 funded by the European Union under the NextGenerationEU Program. Funding for this research has also been provided by the European Union’s Horizon Europe research and innovation programme EFRA (Grant Agreement Number 101093026).
Views and opinions expressed are, however, those of the authors only and do not necessarily reflect those of the European Union or European Commission-EU. Neither the European Union nor the granting authority can be held responsible for them. {\normalsize\euflag}
\bibliography{Sections/biblio_clean}

\appendix
\section{Extended Background}\label{app:background}
Generally, methods explaining ML models are commonly separated into \emph{intrinsic} methods, deriving explanations based on the internal state of the model, and \emph{extrinsic} or \emph{model agnostic} methods, which statistically infer explanations purely from input-to-output-relationships.

\begin{figure*}[t]
    \centering

    \begin{tikzpicture}
    \draw[anchor=east] (0,0) node {\begin{subfigure}{.35\linewidth}
        \centering
        \includegraphics[trim={0 0 0 30}, clip, width=\linewidth]{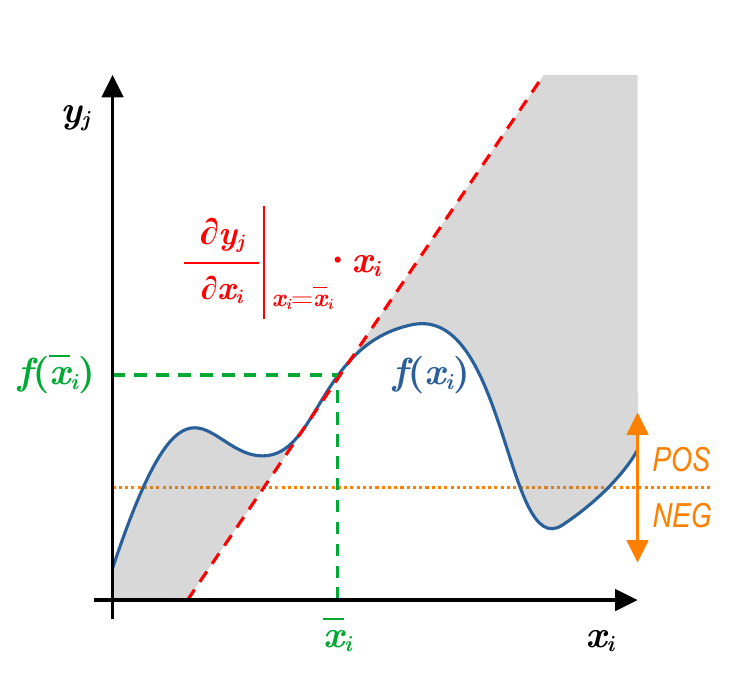}
        \caption{Gradient as a linear approximation.}
    \end{subfigure}};
    \draw[anchor=west] (0,0) node {\begin{subfigure}{.35\linewidth}
        \centering
        \includegraphics[trim={0 0 0 30}, clip, width=\linewidth]{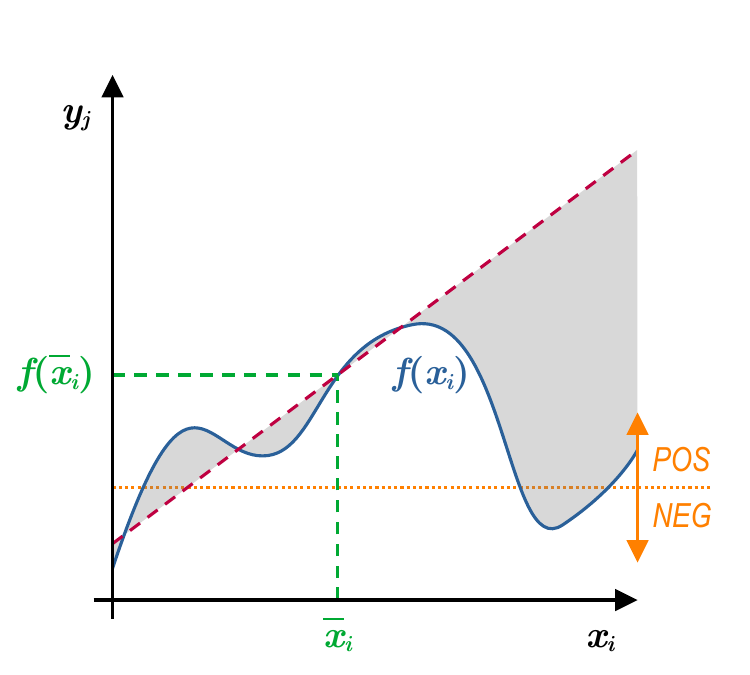}
        \caption{Non-gradient linear approximation.}
    \end{subfigure}};
    \draw (5.5,2) node {\includegraphics[trim={0 260 0 5}, clip, width=.4\linewidth]{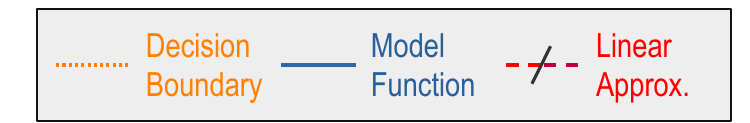}};
    \end{tikzpicture}

  \caption[Gradient-based saliency and its caveats.]{Gradient-based saliency and its caveats. While the gradient is an optimal local linear approximation in the point of the input $\bar{x_i}$, this is not necessarily the case globally or with regard to the intersect of model function and decision boundary.}
  \label{fig:intuition_grad}
\end{figure*}

Early \textbf{intrinsic methods} tailored to transformer models were either \emph{attention-based} or \emph{gradient-based} methods. Finally, there are also hybrid approaches that consider the gradient of the output with regard to the attention weight~\citep{Liu2021_AttentionBasedExplanation}.
\textbf{Attention-based methods} use the fact that the self-attention weights of transformers can be interpreted as weights of how a specific input token impacts the output. They range from simply using the raw attention weights of the last layer as an explanation, to more holistic approaches tracking attention through the whole transformer \citep{Abnar2020_AttentionFlow}. Nevertheless, \citet{Jain2019_AttentionIsNotExplanation} criticise at least the use of raw attention weights as explanations to overestimate their correlation to the output.
\textbf{Gradient-based methods} evaluate the gradient $\left.\frac{\partial y_j}{\partial x_i}\right|_{x_i=\bar{x}_i}$ to get a linear approximation of the impact of the $i^\text{th}$ element of the input on the $j^\text{th}$ element of the output. Simple methods include directly using the gradient as an explanation or multiplying it with the input~\citep{Shrikumar2017_GradIn}. Nevertheless, while optimal in the immediate neighbourhood of $\hat{x}_i$, the raw gradient is not necessarily a good approximation of the global function learned by the ML model~\citep[see Figure~\ref{fig:intuition_grad}]{Shrikumar2017_GradIn, Sundarajan2017_IntegratedGradients}. IG~\citep{Sundarajan2017_IntegratedGradients} solves this by integrating gradients over the input dimension.

Among the most widely used \textbf{extrinsic methods} for LMs are Local Interpretable Model-agnostic Explanations~(LIME) by \citet{Ribeiro2016_LIME} and SHapley Additive exPlanations~(SHAP) by \citet{Lundberg2017_SHAP}.
LIME trains a surrogate Logistic Regression classifier on tuples $(\bar{\bf x}',  \bar{\bf y}')$, with $\bar{\bf x}'$ randomly sampled in the close vicinity of $\bar{\bf x}$, and $\bar{\bf y}' = f(\bar{\bf x}')$ by perturbing elements in $\bar{\bf x}$. The coefficients of this classifier are used as linear attribution scores. Recent extensions of this method focus mostly on improving the generation $\bar{\bf x}'$ for different types of data~\citep{Kuratomi2023_ORANGEOS, Tan2023_GLIME, Angiulli2025_LLime}.
\shap{k}~\citep{Lundberg2017_SHAP} adapts the LIME approach to approximate SV from game theory~\citep{Shapley1953_Value} by deriving a loss function for training the surrogate model which satisfies the \emph{additivity}, \emph{missingness}, and \emph{consistency} constraints of SVs. Extensions of the SHAP approach usually focus on creating faster, intrinsic methods that also satisfy these constraints for more specific classifiers~\citep{mitchell2022_gputreeshap}.
Nevertheless, \citet{covert2021_PairedSamling} test different methods for sampling the input-output pairs necessary to train the surrogate model. They find that sampling complementary pairs of feature sets at the input and their respective outputs improves the stability of \shap{k}.

\section{\expandafter{\xrag{}} Details}\label{app:methods}
\subsection{\expandafter{\ret{}}: Integrated Gradients}
To explain \ret{}, we adapt IG to RAG.
\ret{} encodes a query~${\bf q}$ and document~${\bf d}$ using encoders $e^\text{qry}(\cdot)$ and $e^\text{ctx}(\cdot)$, ranking documents by their dot-product:
\begin{equation}
    s^\text{ret}({\bf q},{\bf d})=e^\text{qry}({\bf q})\cdot e^\text{ctx}({\bf d})~,
\end{equation}
and retrieving the $k$ documents for which the summed similarity is maximal.
Both encoders are pre-trained transformers~$f(\cdot)$ applied to token embeddings $\Phi^\text{qry}(\cdot)$ and $\Phi^\text{ctx}(\cdot)$:
\begin{align}
e^\text{qry}({\bf q}) &= f^\text{qry}(\Phi^\text{qry}({\bf q})),\\
e^\text{ctx}({\bf d}) &= f^\text{ctx}(\Phi^\text{ctx}({\bf d}))
\end{align}
As the computation for IG is analogous for query and contexts, we omit the specifiers ``$^\text{qry}$" and ``$^\text{ctx}$" in the following, and refer to both $\bf q$ and $\bf d$ as $\bf x$.
IG approximates a model ${\bf y}=f({\bf x})$ by integrating its gradients with regard to each input feature~$x_i$ over $x_i$, starting from a chosen baseline of $x_i^0$. 
Following IG, we compute the attributions $\beta_i^{\text{ret,}{\bf x}}$ based on the embeddings using Riemann integration with $L$ steps.
Since the retrieval pipeline receives multiple inputs, we calculate the saliency for the query and each of the retrieved documents separately, holding all other inputs fixed to avoid cross-effects. 
Given $[\phi_1,\dots,\phi_n] = \Phi({\bf x})$ and a baseline embedding~$[\phi^{0}_1,\dots,\phi^{0}_n] = \Phi^{0}({\bf x})$, we compute saliency as:
\begin{equation}\label{eq:ig}
    \beta_i^{\text{ret}} = (\phi_i - \phi^{0}_i) \cdot \sum_{l=1}^L \frac{\partial {\bf s}}{\partial \phi_i}\frac{1}{L}~.
\end{equation}
Here ${\bf s} = s^\text{ret}\left(\delta({\bf q}, l),~{\bf d} \right)$ for the query and $s^\text{ret}\left({\bf q}, \delta({\bf d}, l) \right)$ for each document, with 
\begin{equation}
    \delta({\bf x}, l) = f\left(\Phi^{0}({\bf x}) + \frac{l}{L} \cdot \left(\Phi({\bf x}) - \Phi^{0}({\bf x})\right)\right)~.
\end{equation}

\subsection{\expandafter{\gen{}}: Shapley Style Attributions}
In general, ${\bf y} = \text{\gen{}}({\bf q}, \mathcal{D})$ can be split into two parts:
\textbf{(i)}~the function $[t^\mathcal{D}_{0},t^\mathcal{D}_{1}, \dots, \,t^\mathcal{D}_{n}] = \mathrm{create\_prompt}({\bf q},~\mathcal{D})$ combines the \textit{query~$\bf q$} and \textit{set of retrieved documents~$\mathcal{D} = \{{\bf d}_1, {\bf d}_2, ..., {\bf d}_k\}$} (ordered by descending relevance) to a single sequence of tokens. The function used in this paper is illustrated in Appendix~\ref{app:prompts}.
\textbf{(ii)}~An autoregressive (L)LM iteratively completes this prompt sequence:
\begin{equation}
    t^\mathcal{D}_{i+1} \;=\; \mathrm{LLM}(\!\,[
    \, \underbrace{t^\mathcal{D}_{0}, \dots, \,t^\mathcal{D}_{n},}_{\text{prompt}~\bf x}
    \, \underbrace{t^\mathcal{D}_{n+1}, \dots, t^\mathcal{D}_{i}}_\text{previous generation}
    \!\!])
\end{equation}
The output of this (L)LM is the sequence of generated tokens excluding the prompt: ${\bf y} = [t^\mathcal{D}_{n+1}, \dots, t^\mathcal{D}_m]$.

Given $\bf q$, $\mathcal{D}$, and a function $[t^{\mathcal{D'}}_0, \dots, t^{\mathcal{D'}}_n] = \text{\gen{}}({\bf q}, \mathcal{D'})$ that returns the sequence of tokens based on the subset of documents $\mathcal{D'} \subseteq \mathcal{D}$, the SV for input document ${\bf d}_i$ and output token~$t^{\mathcal{D'}}_j$ averages the marginal contributions over all permutations~\citep{Shapley1953_Value}:
\begin{align}
    {\footnotesize \beta^\text{gen}_{i,j}} &=
    {\footnotesize \!\!\!\!\!\!
        \sum_{\mathcal{D'} \subseteq \mathcal{D} \setminus \{{\bf d}_i\}}
        \!\!\!\!\!\!\mathrm{Pr}(\mathcal{D'}) \cdot
        \underbrace{
            \left( \mathrm{Pr}(t^{\mathcal{D'} \cup \{{\bf d}_i\}}_j) - \mathrm{Pr}(t^{\mathcal{D'}}_j) \right)
        }_{\tiny\text{marginal contribution}}
    } \\
    {\footnotesize \mathrm{Pr}(\mathcal{D'})} &= 
    {\footnotesize \frac{|\mathcal{D'}|! (|\mathcal{D}| - |\mathcal{D'}| - 1)!}{|\mathcal{D}|!}
    }
 \end{align}
$\mathrm{Pr}(t_j)$ symbolizes the softmax output of the (L)LM when generating $t_j$. Computing this precise SV has exponential time complexity $\mathcal{O}(2^{|\mathcal{D}|})$ as the \gen{} model needs to be called for each possible subset $\mathcal{D'} \subseteq \mathcal{D}$. This makes computation feasible only for small numbers of documents (e.g. $|\mathcal{D}|=6$).%

A well-established approximation of SVs for higher $|\mathcal{D}|$ is \shap{k}, a linear surrogate model~$\mathrm{g}'(\cdot)$, trained on simplified features ${\bf z} = [z_1,\dots,z_{|\mathcal{D}|}]$:
\begin{equation}
    z_k = \left\{\begin{array}{cl}
         1 & \text{iff}~{\bf d}_k \in \mathcal{D}' \\
         0 & \text{else}.
    \end{array}\right.
\end{equation}
\citet{Lundberg2017_SHAP} show that the coefficients of $\mathrm{g'}(\cdot)$ approximate SVs and are consistent with their mathematical properties if training minimizes the loss function
\begin{equation}
    \resizebox{\linewidth-2pt}{!}{$
    \sum\limits_{\mathcal{D'} \subseteq \mathcal{D}}
    \left(\begin{matrix}|\mathcal{D}|\\|\mathcal{D'}|\end{matrix}\right)^{-1}
    \!\!\!\!\!\!\cdot
    \frac{|\mathcal{D}| - 1}{|\mathcal{D'}| \cdot (|\mathcal{D}| - |\mathcal{D'}|)}
    \cdot
    \left( \mathrm{Pr}(\mathrm{y}'|\mathcal{D'}) - \mathrm{g}'({\bf z}) \right)^2$},
\end{equation}
given the element-wise model output probabilities
\begin{equation}
 \mathrm{Pr}(\mathrm{y}'|\mathcal{D'}) = \left[\cdots \mathrm{Pr}(t_j) \cdots\right]_{t_j\,\in\, \text{\gen{}}({\bf q}, \mathcal{D'})}.
\end{equation}
\shap{k} trades time complexity for faithfulness, as a lower number $N \le 2^{|\mathcal{D}|}$ of training samples $\mathcal{D'}$ can be used. Nevertheless, SV approximation quality degrades substantially at low $N$ (see Figure~\ref{fig:gen_mc_stab}). As a rule of thumb, we set $N = \min(2^{|\mathcal{D}|}, 10\cdot|\mathcal{D}|)$ to achieve reasonable approximation accuracy at a cost of $\mathcal{O}(N)$ LLM calls. Note that we do not compute \shap{k} on marginal contributions as proposed by \citet{Lundberg2017_SHAP}. This step assumes independent documents which is unrealistic, as documents collected to answer a single query are prone to have overlapping content.

\section{Generation Prompts}\label{app:prompts}

\begin{figure*}[t]
    \centering
    \begin{subfigure}{.45\linewidth}
        \centering
        \resizetable{\linewidth}{
        \begin{tabular}{rl}
            \toprule
            
            \textsc{system:}
                & \texttt{Use the following retrieved documents, ranked from highest} \\
                & \texttt{to lowest relevance, to answer the user's query.} \\
                & \texttt{Be thorough and accurate, and cite documents when useful.} \\
                & \texttt{Keep the answer under 200 words.} \\

            \textsc{user:}
                & \texttt{Document 1: {\color{gray}[...]}} \\
                & \\
                & \texttt{Document 2: {\color{gray}[...]}} \\
                & \\
                & \texttt{Document 3: {\color{gray}[...]}} \\
                & \\
                & \texttt{Document 4: {\color{gray}[...]}} \\
                & \\
                & \texttt{Document 5: {\color{gray}[...]}} \\
                & \\
                & \texttt{Query: {\color{gray}[...]}} \\

            \textsc{model:}
                & \texttt{...} \\

            \bottomrule
        \end{tabular}
        }
        \caption{Llama \& Qwen}
    \end{subfigure}
    \hfill
    \begin{subfigure}{.45\linewidth}
        \centering
        \resizetable{\linewidth}{
        \begin{tabular}{rl}
            \toprule
            
            \textsc{user:}
                & \texttt{Use the following retrieved documents, ranked from highest} \\
                & \texttt{to lowest relevance, to answer the user's query.} \\
                & \texttt{Be thorough and accurate, and cite documents when useful.} \\
                & \texttt{Keep the answer under 200 words.} \\
                & \\
                & \texttt{Document 1: {\color{gray}[...]}} \\
                & \\
                & \texttt{Document 2: {\color{gray}[...]}} \\
                & \\
                & \texttt{Document 3: {\color{gray}[...]}} \\
                & \\
                & \texttt{Document 4: {\color{gray}[...]}} \\
                & \\
                & \texttt{Document 5: {\color{gray}[...]}} \\
                & \\
                & \texttt{Query: {\color{gray}[...]}} \\

            \textsc{model:}
                & \texttt{...} \\

            \bottomrule
        \end{tabular}
        }
        \caption{Gemma}
    \end{subfigure}
    \caption{Prompt formats for the \gen{} models. Grayed sections ``\texttt{\color{gray}[...]}" are replaced by the respective content.}
    \label{fig:prompts}
\end{figure*}

\noindent Fig.~\ref{fig:prompts} shows the formats for the prompts used to generate answers in our experiments and ablations. These correspond to the ``$\mathrm{create\_prompt}({\bf q},\mathcal{D})$" functions referenced earlier. In Llama we include the instructions in the system prompt and provide query and context documents in the user prompt. Since Gemma does not support system prompts, we include everything in the user prompt.

\section{Extended Analysis on Design Choices}
\subsection{Monte-Carlo stabilization for \expandafter{\shap{k}}}\label{app:abl_shap_mc}
\begin{figure*}[p]
    \centering
    \includegraphics[width=.7\linewidth]{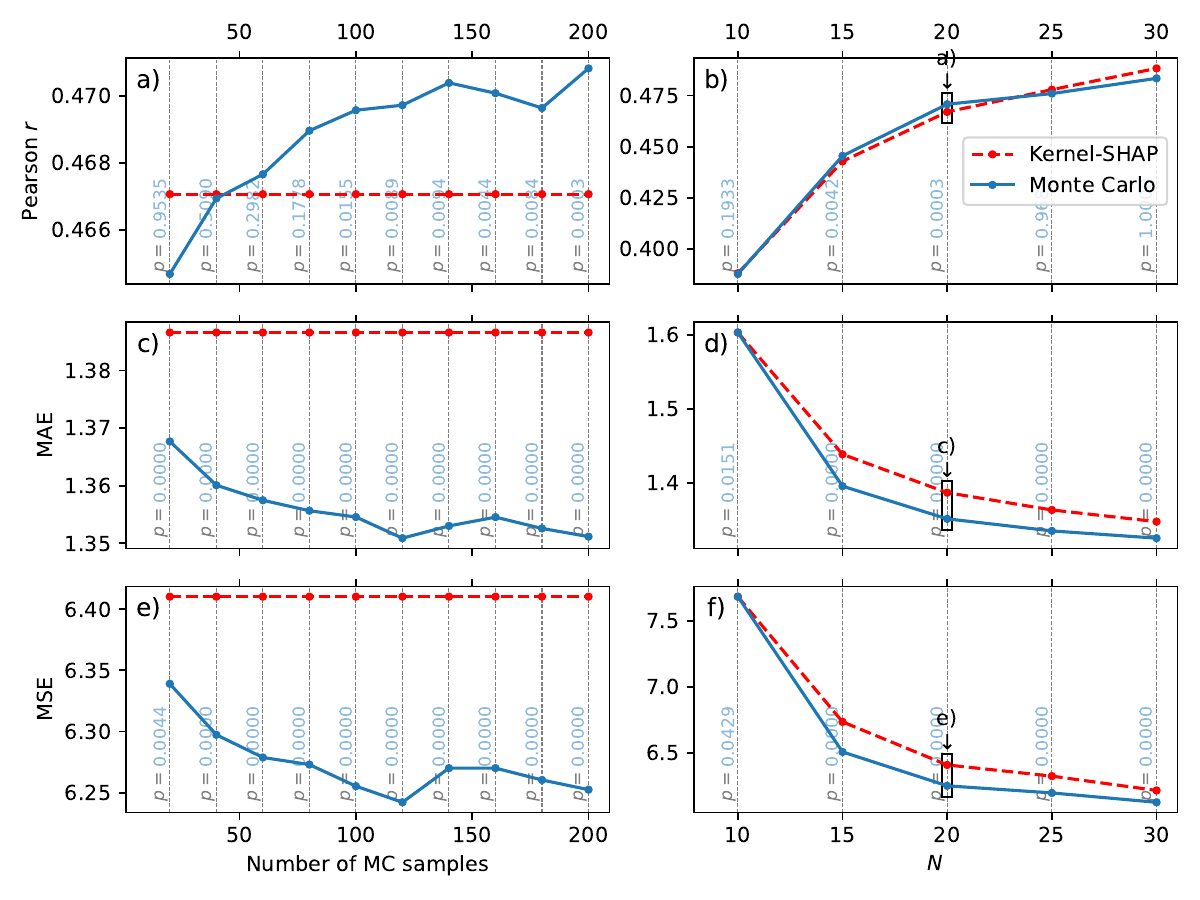}
    \caption{Mean Pearson~$r$, MAE, and MSE comparing \shap{k} and \shap{mc} to precise SV over 200 random samples of MS-Marco. Computed for $|\mathcal{D}| = 5$ and native \textbf{uniform sampling}. The blue vertical numbers are the $p$-values for a paired Wilcoxon Signed Rank test with alternate hypothesis that the \shap{mc} metric is greater than (for Pearson-$r$) or less than (MAE \& MSE) the \shap{k} metric (\textit{Left~column:} results for $N=20$; \textit{Right~column:} results for 200 MC samples).}\label{fig:shap_error_uniform}
\end{figure*}
\begin{figure*}[p]
    \centering
    \includegraphics[width=.7\linewidth]{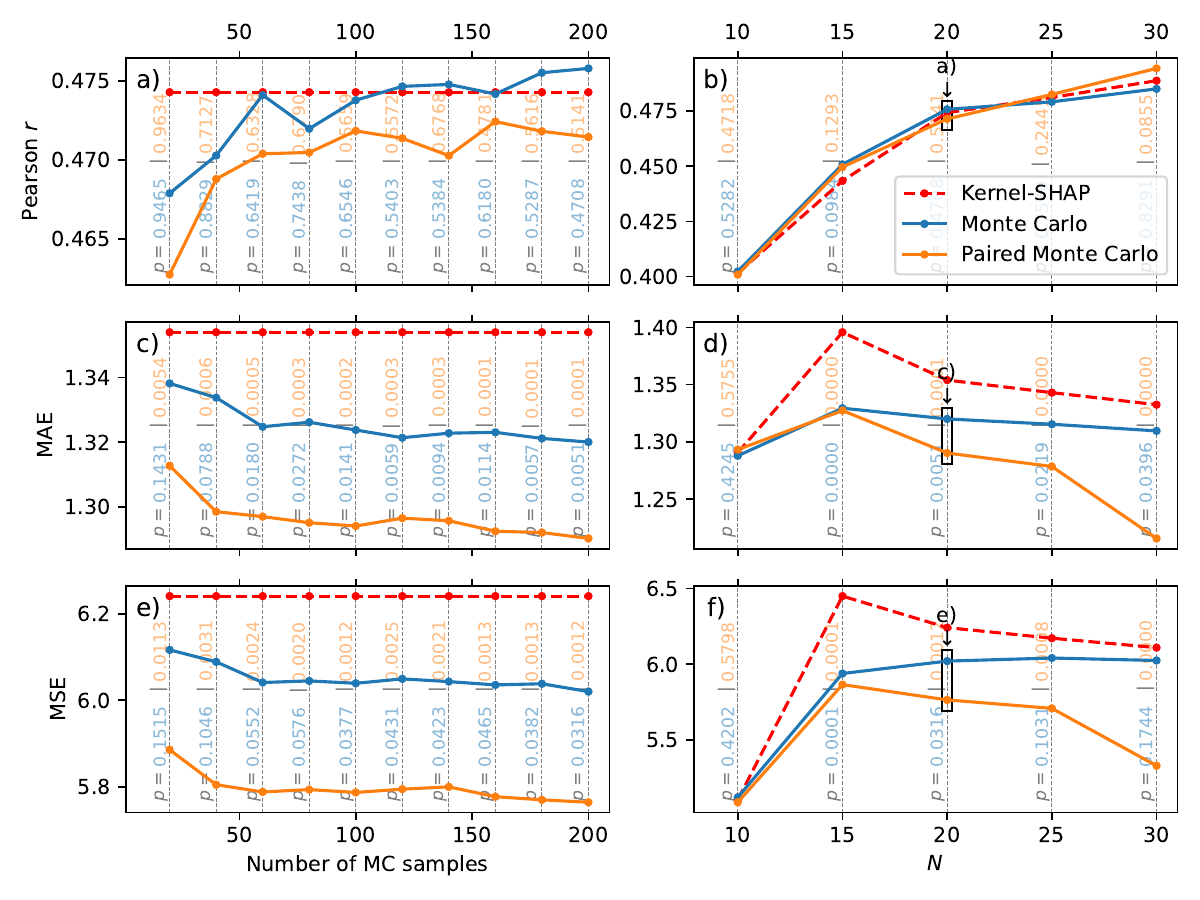}
    \caption{Pearson~$r$, MAE, and MSE comparing \shap{k}, \shap{mc}, and \shap{pmc} over 200 random samples of MS-Marco. Computed for precise SV for $|\mathcal{D}| = 5$ and \textbf{complementary sampling}. The blue/orange vertical numbers are the $p$-values for a paired Wilcoxon Signed Rank test with alternate hypothesis that the \shap{(p)mc} metric is greater than (for Pearson-$r$) or less than (MAE \& MSE) the \shap{k} metric (\textit{Left~column:} results for $N=20$; \textit{Right~column:} results for 200 MC samples).}\label{fig:shap_error_complementary}
\end{figure*}

Figures~\ref{fig:shap_error_uniform} and \ref{fig:shap_error_complementary} show comparisons of \shap{k} and \shap{(p)mc} methods for the \gen{}. We present comparisons for both native uniform sampling (Fig.~\ref{fig:shap_error_uniform}) and complementary sampling (Fig.~\ref{fig:shap_error_complementary}). A non-parametric Wilcoxon Signed Rank test shows a significant improvement of uniform \shap{mc} and complementary \shap{pmc} over \shap{k} under identical sampling for $\ge40$ MC samples at $N=20$ or $15 \le N \le 25$ at 200 MC samples. For uniform sampling, we also see significantly better correlation of SV and \shap{mc} for these values.

\subsection{Faithfulness Analysis}\label{app:abl_faith}
We measure faithfulness as the Area Inside the Perturbation Curves~(AIPC) via input perturbation. Concretely, we repeatedly call the model while gradually masking input tokens according to their attributed importance until the input is completely obscured. We perturb the input in both directions, removing tokens Most Relevant First (MoRF) and Least Relevant First (LeRF). A \textit{faithful} explanation should trigger an early change in the MoRF setting, since highly influential tokens are masked first, and only a late change in the LeRF setting, because initially unimportant tokens are removed and should not affect the prediction much.

This metric is widely applied in literature but not standardized~\cite{Liu2021_AttentionBasedExplanation, edin2025-aopc, Randl2025_SelfExplanations}. We compute AIPC per input sequence ${\bf x}$, masking it from $0$ to all $n$ tokens, one token at a time. Let $\mathrm{mask}_\text{dir}({\bf x},i)$ denote the input ${\bf x}$ with $i$ tokens masked according to direction ``dir". We measure the area in between the perturbation curves
\begin{equation}
{\footnotesize\begin{array}{rl}
    \!\!\mathrm{AIPC}_\text{qry} =

    & \!\!\!\!\left| \int_{i=0}^{n} g\left(\mathrm{mask}_\text{MoRF}({\bf q},i), {\bf d}\right) - g\left({\bf q}, {\bf d}\right)~di \right| -\\\\

    & \!\!\!\!\left| \int_{i=0}^{n} g\left(\mathrm{mask}_\text{LeRF}({\bf q},i), {\bf d}\right) - g\left({\bf q}, {\bf d}\right)~di \right|\\
\end{array}}
\end{equation}
for the query and
\begin{equation}
{\footnotesize\begin{array}{rl}
    \!\mathrm{AIPC}_\text{ctx} =
    
    & \!\!\!\!\left| \int_{i=0}^{n} g\left({\bf q}, {\bf d}\right) - g\left({\bf q}, \mathrm{mask}_\text{MoRF}({\bf d},i)\right)~di \right| - \\\\
    
    & \!\!\!\!\left| \int_{i=0}^{n} g\left({\bf q}, {\bf d}\right) - g\left({\bf q}, \mathrm{mask}_\text{LeRF}({\bf d},i)\right)~di \right| \\
\end{array}}
\end{equation}
for context documents. For the \ret{} we choose $g({\bf q}, {\bf d}) = s_{ret}({\bf q}, {\bf d})$ and for the \gen{} $g({\bf q}, {\bf d})$ is the model output. We then min-max-normalize per sample and report the mean AIPC over all inputs. Consequently, AIPC values theoretically range between $0$ (not faithful) and $1$ (maximally faithful). Note that the realistic maximum is model dependent and always less than $1$.

\paragraph{\expandafter\ret{} Explanations:} We compare the faithfulness of our IG \ret{} explanations to the raw gradient~($\text{Grad}: \beta_i = \frac{\partial f({\bf x})}{\partial x_i}$), gradient$\times$input~\cite[$\text{GradIn}: \beta_i = \frac{\partial f({\bf x})}{\partial x_i} \cdot x_i$]{Shrikumar2017_GradIn}, and Attention Gradient~\cite[AGrad]{Liu2021_AttentionBasedExplanation} as intrinsic baselines. As both $e^\text{qry}(\cdot)$ and $e^\text{ctx}(\cdot)$ are encoder-only transformers ${\bf{y}} = f(\Phi(x))$ atop an embedding $\Phi({\bf x})={\bf Wx}+{\bf p}$ (where ${\bf p}$ is a positional embedding), this means we have to also calculate the gradient over $\Phi({\bf x})$ which is not supported by PyTorch's \texttt{autograd}.
We therefore calculate the gradient to the input manually:
\begin{equation}
\frac{\partial e({\bf x})}{\partial {\bf x}}
=
\underbrace{\left.\frac{\partial f(\Phi({\bf x}))}{\partial \Phi({\bf x})}\right|_{{\bf x}=\bar{\bf x}}}_\text{computed by \texttt{autograd}}\!\!\! \cdot {\bf W}~.
\end{equation}

Furthermore, we compare to \shap{k}~\cite{Lundberg2017_SHAP} and LIME~\cite{Ribeiro2016_LIME} as extrinsic baselines. For both we use a sample size of $10\cdot\text{\it number of input features}$. To keep runtime of these methods within acceptable boundaries these input features are token multi-grams of size $\lceil \frac{\text{\it number of input features}}{63} \rceil$, so that the maximum number of features never exceeds 63. This also allows memory efficient storage of the simplified inputs~${\bf z} \in \{0,1\}^n$ in signed \texttt{int64} bitmasks. We do not use the sign bit to keep native compatibility between raw Python, NumPy, and PyTorch types.

\begin{table*}[p]
    \centering
    \resizetable{\linewidth}{\begin{tabular}{l|ccccccc}
        \toprule
        
        \multirow{2}{*}{\textbf{\ret{}}} &
        \multicolumn{7}{c}{\textbf{Method}} \\
        &
        \footnotesize{RND} &
        \footnotesize{Grad} &
        \footnotesize{GradIn} &
        \footnotesize{AGrad} &
        \footnotesize{LIME} &
        \footnotesize{\shap{k}} &
        \footnotesize{IG} \\

        \midrule
    
        \multirow{2}{*}{DRAGON} &
        0.00 &
        0.21 &
        0.10 &
        \textbf{0.50} &
        0.49 &
        0.48 &
        \textbf{0.50} \\
        &
        \footnotesize [-0.01, 0.01] &
        \footnotesize [0.19, 0.22] &
        \footnotesize [0.09, 0.12] &
        \textbf{\footnotesize [0.49, 0.51]} &
        \footnotesize [0.48, 0.50] &
        \footnotesize [0.47, 0.49] &
        \textbf{\footnotesize [0.49, 0.51]} \\
    
        \multirow{2}{*}{Arctic Embed 2} &
         0.00 &
        -0.18 &
         0.23 &
         0.64 &
         0.69 &
         0.69 &
        \textbf{0.73} \\
        &
        \footnotesize [-0.01, 0.02] &
        \footnotesize [-0.21, -0.15] &
        \footnotesize [0.21, 0.25] &
        \footnotesize [0.62, 0.66] &
        \footnotesize [0.67, 0.70] &
        \footnotesize [0.67, 0.70] &
        \textbf{\footnotesize [0.70, 0.76]} \\

        \bottomrule
    \end{tabular}}
    \caption{\ret{} faithfulness in terms of AIPC (\textit{higher is better}) for different explanation methods. Bold scores mark the most faithful explainability method per \ret{} model. 95\% confidence intervals (computed over 1000 bootstrap samples) are reported as [lower, upper].}
    \label{tab:faith_ret_meth}
\end{table*}

Since the query texts are naturally short in nature, and perturbing a single token can easily distort the meaning of the whole text, independent of the tokens' impact on the decision, we test faithfulness only on the context documents. However, as the applied IG method is identical for contexts and query, we argue that the results are transferable.
The results of this test are shown in Tab.~\ref{tab:faith_ret_meth}. For both tested \ret{}s, IG performs best. Although, surprisingly, AGrad is equally faithful as IG for the DRAGON \ret{}, IG remains the overall best.

\begin{table*}[p]
    \centering
    \setlength{\tabcolsep}{.4cm}
    \begin{tabular}{lrr|ccc}
        \toprule
        \multicolumn{3}{c|}{\multirow{2}{*}{\textbf{\ret{} / Metric}}} &
        \multicolumn{3}{c}{\textbf{Integration Steps ($L$)}} \\
        & & &
        \footnotesize{$L{=}10$} &
        \footnotesize{$L{=}50$} &
        \footnotesize{$L{=}100$} \\
        \midrule

        \multirow{3}{*}{DRAGON}
        &   \multicolumn{2}{r|}{\footnotesize AIPC} &
            0.40 \footnotesize [0.39, 0.41] &
            \textbf{0.41 \footnotesize [0.39, 0.42]} &
            \textbf{0.41 \footnotesize [0.39, 0.42]} \\
        \cmidrule{2-3}
        &   \multirow{2}{*}{$\frac{\sum_{i = 1}^{n} \beta_{i, j}}{\bar{y}_j - \bar{y}^0_j}$}
            &   \footnotesize query &
                0.84 \footnotesize [0.02, 1.99] &
                1.00 \footnotesize [0.93, 1.08] &
                0.94 \footnotesize [0.88, 0.99] \\
        &   &   \footnotesize context &
                0.89 \footnotesize [0.88, 0.91] &
                0.98 \footnotesize [0.97, 0.98] &
                0.99 \footnotesize [0.99, 0.99] \\

        \midrule

        \multirow{2}{*}{Arctic}
        &   \multicolumn{2}{r|}{\footnotesize AIPC} &
            0.65 \footnotesize [0.62, 0.67] &
            \textbf{0.67 \footnotesize [0.65, 0.69]} &
            \textbf{0.67 \footnotesize [0.65, 0.69]} \\
        \cmidrule{2-3}
        \multirow{2}{*}{Embed 2}
        &   \multirow{2}{*}{$\frac{\sum_{i = 1}^{n} \beta_{i, j}}{\bar{y}_j - \bar{y}^0_j}$}
            &   \footnotesize query &
                0.91 \footnotesize [0.87, 0.96] &
                0.98 \footnotesize [0.98, 0.99] &
                0.99 \footnotesize [0.99, 0.99] \\
        &   &   \footnotesize context &
                0.77 \footnotesize [0.65, 0.86] &
                0.99 \footnotesize [0.90, 1.12] &
                0.98 \footnotesize [0.95, 1.01] \\

        \bottomrule
    \end{tabular}
    \caption{Faithfulness in terms of AIPC (\textit{higher is better}) for different integration steps $L$ of IG ($\Phi^0$ fixed to \texttt{[pad]}), with additivity ratios reported for query and context. Bold scores mark the most faithful IG setting per \ret{}. 95\% confidence intervals (computed over 1000 bootstrap samples) are reported as [lower, upper].}
    \label{tab:faith_ret_ig_steps}
\end{table*}

To test the design choices we use for the IG explanations, we ablated the number of Riemann integration steps~$L$ in terms of faithfulness and additivity ratio. Our results, presented in Tab.~\ref{tab:faith_ret_ig_steps}, suggest that using $L=100$ steps sufficiently approximates the integral, as the attributions sum up to more than 90\% of the output, and faithfulness on the context attributions plateaus.

\paragraph{\expandafter\gen{} Explanations:}
In order to verify the faithfulness of our \gen{} explanations, we compute AIPC for LIME, native \shap{k}, \shap{mc}, and our proposed \shap{pmc} for native uniform sampling as well as complementary sampling~\citep{covert2021_PairedSamling} over different \textit{sample sizes}~$N$. Additionally, we compare to ContextCite and MIRAGE using their standard parameters. Note that ContextCite and MIRAGE attribute the \emph{entire} output to documents, while the remaining methods compute document influence \emph{on each output token}. We therefore assume flat attribution to all output tokens for the former. Note that, for ContextCite, we use the official code from PyPi, while we reimplemented MIRAGE to make it compatible with our hardware. The results, reported in Tab.~\ref{tab:faith_ret_est}, show comparable faithfulness for LIME and all SV-based methods. Context Cite and MIRAGE score approximately at half their faithfulness or lower, which can be attributed to their missing attribution accuracy per output token.

\begin{table*}[p]
    \centering
    \resizetable{\linewidth}{\begin{tabular}{ll|c|ccccc|c}
    \toprule
     
    \multicolumn{2}{c|}{\multirow{2}{*}{\textbf{Method / Sampling}}} &
    \multicolumn{7}{c}{\textbf{AIPC}} \\
        &
        &
        \footnotesize{Random} &
        \footnotesize{$N{=}10$} &
        \footnotesize{$N{=}15$} &
        \footnotesize{$N{=}20$} &
        \footnotesize{$N{=}25$} &
        \footnotesize{$N{=}30$} &
        \footnotesize{Precise} \\
    
    \midrule
    
    LIME & unif.
        & \multirow{8}{*}{\color{gray} 0.00 \footnotesize [-0.01, 0.01]}
        & 0.66 \footnotesize [0.64, 0.68]
        & 0.70 \footnotesize [0.68, 0.72]
        & 0.71 \footnotesize [0.69, 0.73]
        & 0.72 \footnotesize [0.70, 0.74]
        & 0.72 \footnotesize [0.70, 0.74]
        & \multirow{8}{*}{\color{gray} 0.72 \footnotesize [0.70, 0.74]} 
        \\
    
    \shap{k} & unif.
        &
        & 0.65 \footnotesize [0.63, 0.68]
        & 0.69 \footnotesize [0.68, 0.71]
        & 0.71 \footnotesize [0.69, 0.73]
        & 0.72 \footnotesize [0.70, 0.74]
        & 0.72 \footnotesize [0.70, 0.74]
        &
        \\
        
    \shap{mc} & unif.
        &
        & 0.66 \footnotesize [0.63, 0.68]
        & 0.70 \footnotesize [0.68, 0.71]
        & 0.71 \footnotesize [0.69, 0.73]
        & 0.71 \footnotesize [0.69, 0.73]
        & 0.72 \footnotesize [0.70, 0.74]
        &
        \\
    
    \shap{k} & compl.
        &
        & 0.65 \footnotesize [0.64, 0.68]
        & 0.69 \footnotesize [0.67, 0.71]
        & 0.71 \footnotesize [0.69, 0.73]
        & 0.72 \footnotesize [0.70, 0.74]
        & 0.72 \footnotesize [0.70, 0.74]
        &
        \\
    
    \shap{mc} & compl.
        &
        & 0.65 \footnotesize [0.63, 0.67]
        & 0.69 \footnotesize [0.67, 0.71]
        & 0.71 \footnotesize [0.69, 0.73]
        & 0.71 \footnotesize [0.69, 0.73]
        & 0.72 \footnotesize [0.70, 0.74]
        &
        \\
    
    \shap{pmc} & compl.
        &
        & 0.65 \footnotesize [0.64, 0.68]
        & 0.69 \footnotesize [0.68, 0.71]
        & 0.71 \footnotesize [0.69, 0.73]
        & 0.71 \footnotesize [0.69, 0.73]
        & 0.72 \footnotesize [0.70, 0.74]
        &
        \\[5pt]

    \multicolumn{2}{l|}{ContextCite}
        &
        & \multicolumn{5}{c|}{---------------------------------------- 0.38 {\footnotesize [0.35, 0.40]} ----------------------------------------}
        &
        \\

    \multicolumn{2}{l|}{MIRAGE}
        &
        & \multicolumn{5}{c|}{----------------------------------------  0.25 {\footnotesize [0.22, 0.27]} ---------------------------------------- }
        &
        \\
    
    \bottomrule
    \end{tabular}
    }
    \caption{Faithfulness in terms of AIPC (\textit{higher is better}) for LIME and different \shap{k} extensions and $|\mathcal{D}| = 5$. 95\% confidence intervals (computed over 1000 bootstrap samples) are reported as [lower, upper]. The columns ``Random'' and ``Precise'' show the faithfulness of randomized and precise SV attributions, and signify theoretical bounds for faithfulness. For the ContextCite and MIRAGE competitors, we only report faithfulness for their default parameters, and we assume equal attribution for all output tokens.}
    \label{tab:faith_ret_est}
\end{table*}

\subsection{Runtime Analysis}\label{app:abl_time}

\begin{table}[h!]
    \centering
    \setlength{\tabcolsep}{4mm}
    \begin{tabular}{lrrr}
        \toprule
        & \textbf{mean} & \textbf{min} & \textbf{max} \\
        \midrule
        LIME          & 28.7\,s & 18.7\,s & 44.8\,s \\
        \shap{k}      & 27.1\,s & 18.4\,s & 38.8\,s \\
        ContextCite   & 31.5\,s & 20.8\,s & 40.9\,s \\
        MIRAGE        & 28.9\,s & 19.6\,s & 41.5\,s \\
        \shap{pmc}    & 27.7\,s & 19.3\,s & 40.4\,s \\
        \bottomrule
    \end{tabular}
    \caption{Average, minimum, and maximum runtime of \gen{} document attribution methods over 10 queries from MS-Marco, supported by 10 documents each.}
    \label{tab:time_gen}
\end{table}

To provide a concrete estimate of \shap{pmc}'s runtime compared to competitors, we computed the average combined runtime of answer generation and document attribution on a random sample of ten queries from MS-Marco. Each query is supported by ten documents. For both ContextCite and MIRAGE, we use document-level perturbation along with their official repositories' default parameters.
As shown in Table~\ref{tab:time_gen}, runtime is largely similar across methods, indicating that \shap{pmc}'s runtime is compatible with the state of the art.

\section{Extended Analysis}
\subsection{Retrieval Performance}
\label{app:ret_performance}

To validate our retrieval component, we evaluated the performance of the \textit{DRAGON} and \textit{Arctic Embed 2} embedders on the PopQA, QAMPARI, and TC datasets. In Tab.~\ref{tab:ret_metrics} we report the NDCG@3, and Precision@1.
In this phase, we report only fine-grained metrics given the fact that our goal is not to evaluate the retrieval phase but to assess the alignment between the \ret{} and the \gen{}.

\begin{table}[h]
\centering
\setlength{\tabcolsep}{1pt}
\resizetable{\linewidth}{\begin{tabular}{ll|cc}
\midrule
\textbf{Dataset} & \textbf{\ret{}} & \textbf{P@1} & \textbf{NDCG@3}  \\ 

\midrule
\multirow{2}{*}{PopQA}
& Arctic Embed 2 &	0.845 & 0.845 \\
& DRAGON         &	0.820 &	0.803 \\

\midrule
\multirow{2}{*}{QAMPARI}
& Arctic Embed 2 & 0.650 &	0.673 \\
& DRAGON         & 0.620 &	0.662 \\

\midrule
\multirow{2}{*}{TC}
& Arctic Embed 2 & 0.723 & 0.475 \\
& DRAGON         & 0.699 & 0.441 \\

\bottomrule
\end{tabular}}
\caption{Retrieval Performances. We report here Precision@1 (P@1) and Normalized Discounted Cumulative Gain for the first three documents (NDCG@3)}
\label{tab:ret_metrics}
\end{table}

\subsection{POS Tag Analysis}\label{app:pos_analysis}
\begin{figure*}[t]
    \centering
    \begin{subfigure}{0.35\linewidth}
        \centering
        \includegraphics[trim={0 19 0 20},clip,height=5.5cm]{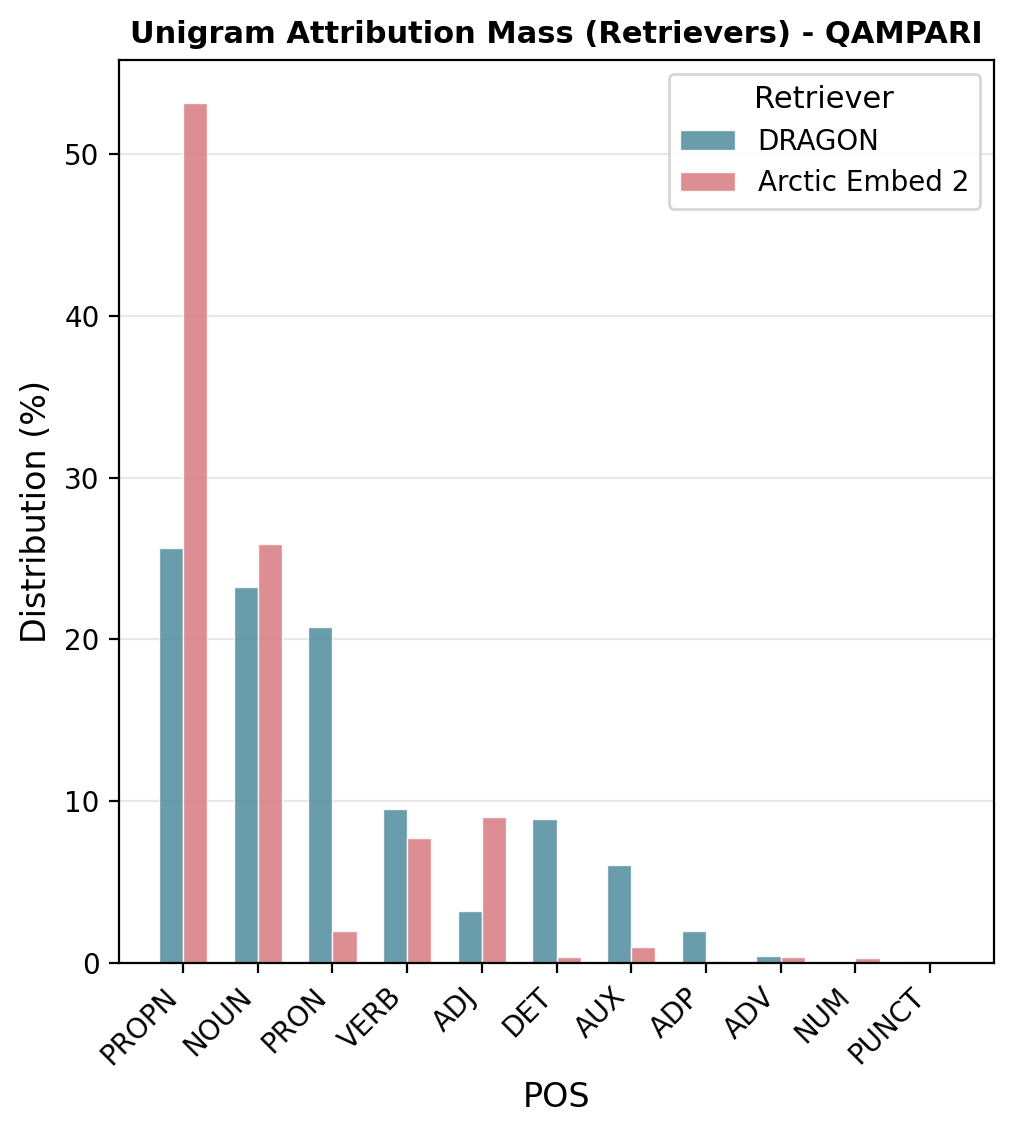}
        \caption{Unigram POS Attribution\\for \textit{Arctic Embed 2} and \textit{DRAGON}}
        \label{fig:pos_unigram_retrievers}
    \end{subfigure}
    \hfill
    \begin{subfigure}{0.60\linewidth}
        \centering
        \includegraphics[trim={0 24 0 22},clip,height=5.5cm]{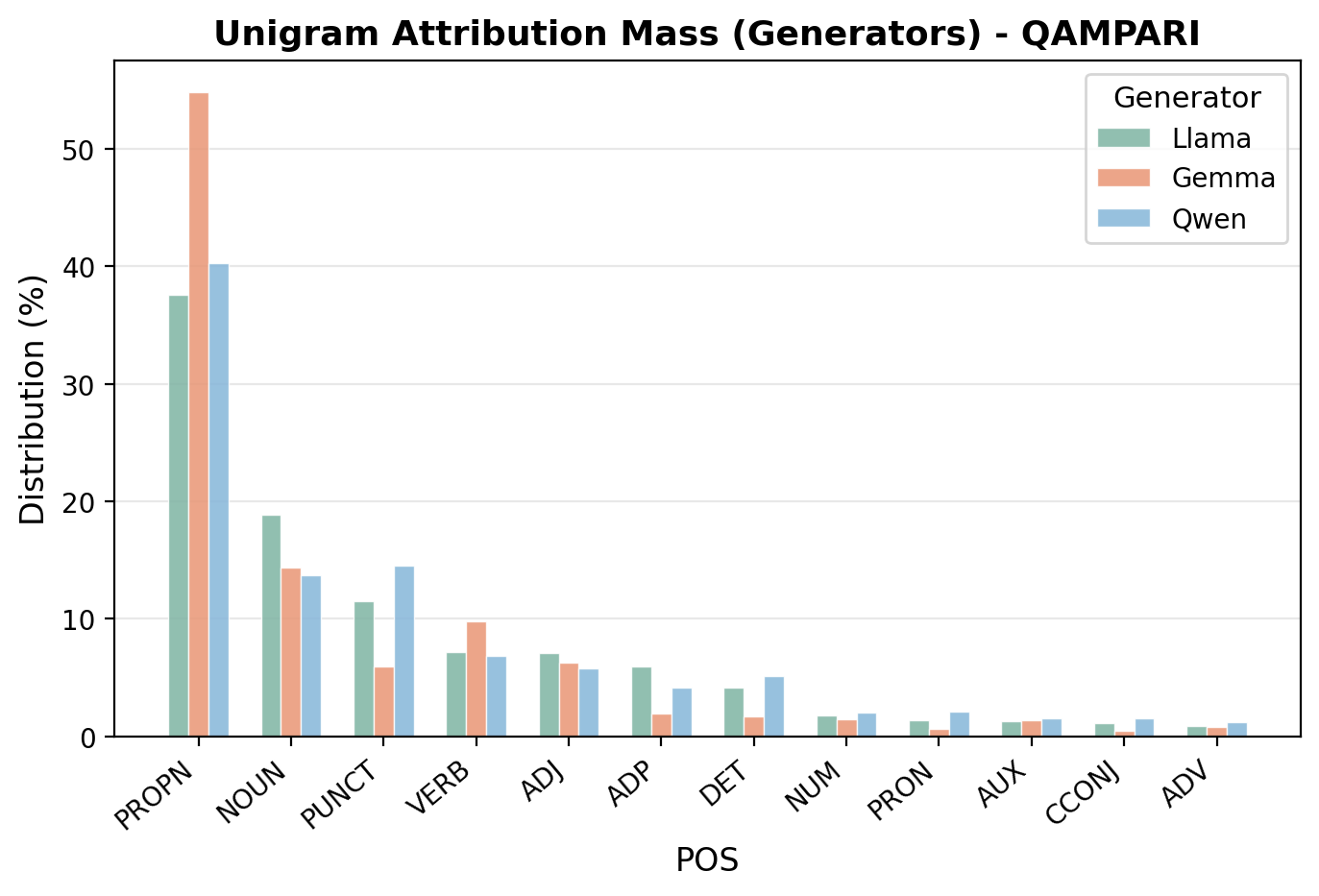}
        \caption{Unigram POS Attribution\\for \textit{Llama}, \textit{Gemma}, and \textit{Qwen}}
        \label{fig:pos_unigram_generators}
    \end{subfigure}

    \begin{subfigure}{0.35\linewidth}
        \centering
        \includegraphics[trim={0 18 0 18},clip,height=5.51cm]{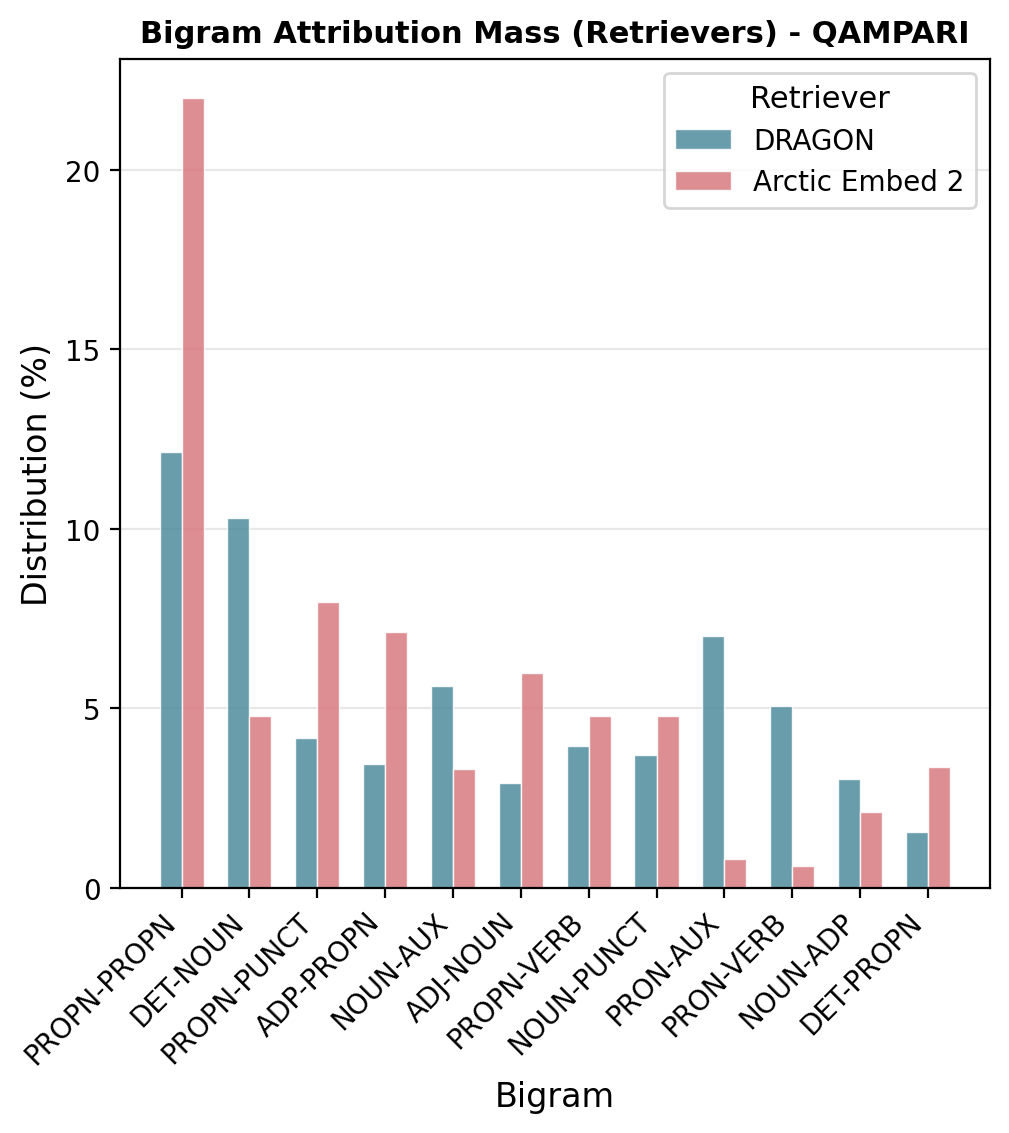}
        \caption{Bigram POS Attribution\\for \textit{Arctic Embed 2} and \textit{DRAGON}}
        \label{fig:pos_bigram_retrievers}
    \end{subfigure}
    \hfill
    \begin{subfigure}{0.60\linewidth}
        \centering
        \includegraphics[trim={0 25 0 20},clip,height=5.51cm]{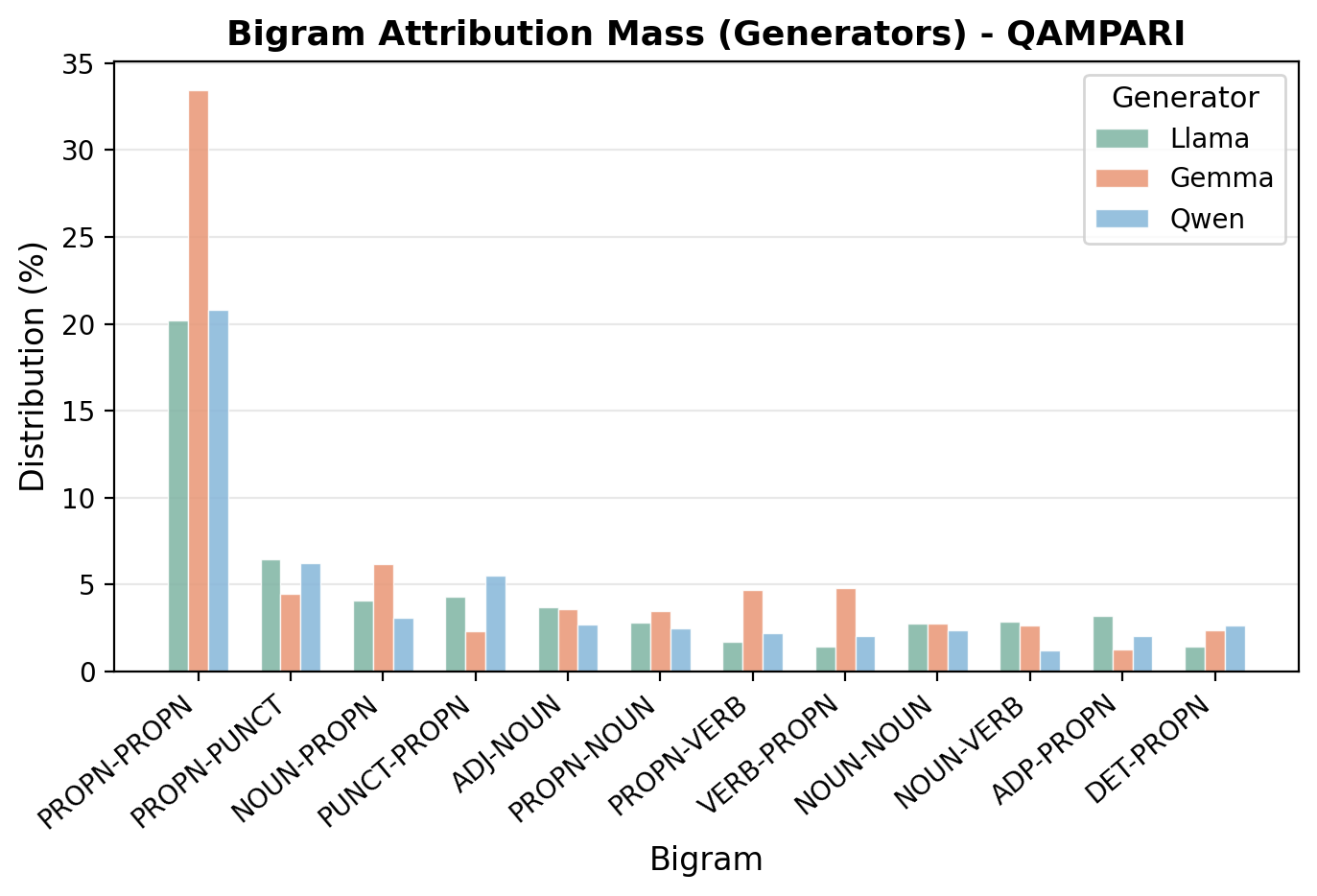}
        \caption{Bigram POS Attribution\\for \textit{Llama}, \textit{Gemma}, and \textit{Qwen}}
        \label{fig:pos_bigram_generators}
    \end{subfigure}
    \caption{POS attribution results on QAMPARI.}
    \label{fig:pos_combined}
\end{figure*}

In this section, we report some analysis regarding the attribution given by the models to the different parts of speech.
We analysed the distribution of attribution weights across Part-of-Speech (POS) tags and POS bigrams (word pairs) for both \gen{}s and \ret{}s, focusing on the top 30\% of attributed tokens (Figure~\ref {fig:pos_combined}). Our analysis reveals several key similarities and distinctions in how these models attend to linguistic features.
To be able to compute this, we first reconstructed the words from the tokens and successively summed the attributions outputted by the models.

Across all five models, PROPNs consistently receive the highest attribution mass. This indicates a shared reliance on substantive content words as the primary elements of meaning and relevance. Furthermore, PROPN + PROPN is overall the most significant bigram, which highlights the importance of identifiers in both generation and retrieval contexts.
While \gen{}s prioritize nouns, their secondary focuses diverge significantly. Gemma and Qwen exhibit a high sensitivity to punctuation and structural markers. PUNCT is the second most attributed tag by Qwen, and the PROPN + PUNCT values are also prominent for LLama. This might suggest that the attention mechanism relies heavily on sentence boundaries and delimiters to organize context.
On the other hand, Gemma focuses more on VERBs, while LLama focuses on function words like ADP (adpositions). Its bigrams, such as NOUN + ADP, reflect a deeper engagement with the grammatical structure and prepositional relationships within the text, rather than just isolated keywords or delimiters.

\paragraph{\ret{} Comparison: Arctic-Embed 2 vs. DRAGON}
The \ret{}s show a stronger alignment with entity-centric processing compared to the \gen{}s, but with nuanced differences.
Arctic-Embed 2 shows a distinct preference for \texttt{PROPN}~(Proper Nouns). Its top bigrams, like \texttt{PROPN + NOUN}, suggest it functions closer to a traditional keyword-based \ret{}, heavily prioritizing named entities and specific terminology to match queries.
DRAGON shares the entity focus (high {PROPN}) but also places significant weight on \texttt{PRON}~(Pronouns). This may suggest a more semantic ``understanding" of the query, attending to descriptive qualifiers and anaphoric references (e.g., \texttt{PRON + AUX}) which are typical of dense retrieval models that capture "meaning" beyond exact keyword matches.

\paragraph{\gen{} vs. \ret{}}
A broad distinction emerges between the two classes of models. On the one hand, \gen{}s distribute attention more broadly across syntactic elements (\texttt{VERB}, \texttt{ADP}, \texttt{DET}) required to construct coherent sentences and follow narrative flow. On the other hand, \ret{}s concentrate their mass more narrowly on content-bearing classes (\texttt{NOUN}, \texttt{PROPN}, \texttt{ADJ}), filtering out much of the connectors to maximize relevance attribution.

\subsection{Primacy Bias}\label{app:primacy_bias}
We evaluate the stability of \gen{} document attributions with respect to document order in the prompt.
Specifically, we compute the mean attribution score for each document included in the prompt under the \textsc{c1} and \textsc{c4} orderings across all tested \gen{}'s and Arctic Embed 2.
Under a generic prompting setup, a document’s importance should depend on its content rather than its position. Note that in our case, the prompt explicitly specifies the ranked nature of the provided documents, which is expected to influence the attributions assigned by the models (Appendix~\ref{app:prompts}).

\begin{figure}[h]
\centering
\includegraphics[width=\linewidth,keepaspectratio]{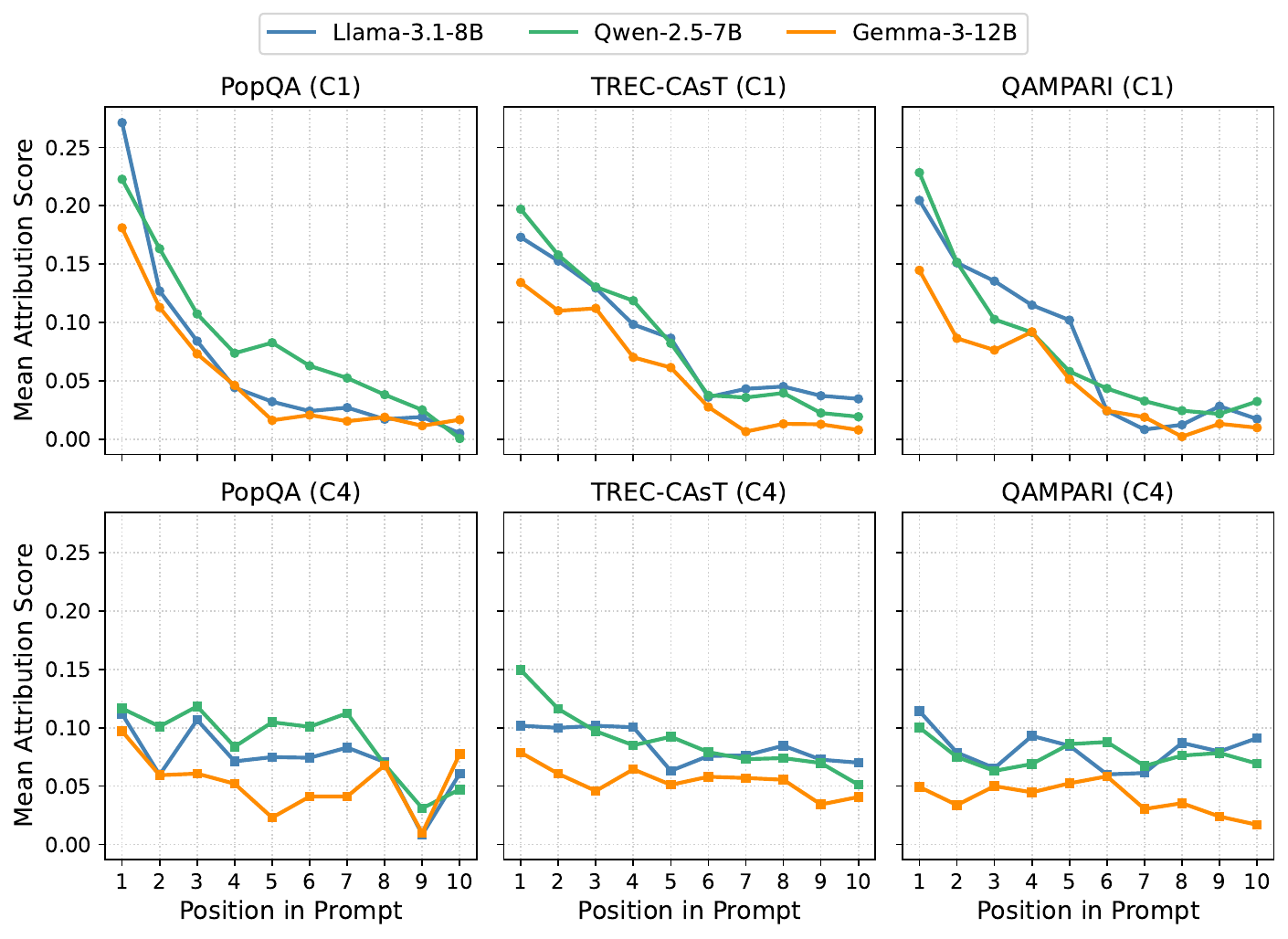}
\caption{Attribution Instability for Arctic Embed 2.}
\label{fig:stability_snowflake}
\end{figure}

As shown in Fig.~\ref{fig:stability_snowflake}, all models tend to attribute more to the first documents when prompted with the ranked documents (\textsc{c1}). 
When shuffling the prompt documents (\textsc{c4}), we observe that for all \gen{}s the attributions distribute uniformly throughout the documents. Contrary to literature~\citep{Liu2024_lostinthemiddle}, this suggests that the tested models can avoid primacy bias and detect important information regardless of document position. We assume that, due to the relatively short context sizes we use, the prompts are short enough for the LLMs to capture them entirely.

\subsection{WARG Analysis}\label{app:warg+}
\begin{figure*}[p]
    \centering
    \includegraphics[trim={0 270pt 0 4pt},clip,width=\linewidth]{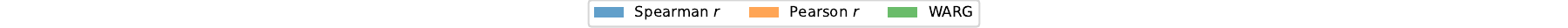}

    {\tiny\hspace{.9cm}\hfill \textsf{DRAGON / Llama} \hfill\hfill \textsf{DRAGON / Qwen} \hfill\hfill \textsf{DRAGON / Gemma} \hfill\textcolor{white}{ }}
    
    \includegraphics[trim={0 145pt 0 45pt},clip,width=\linewidth]{figures/boxplots_warg.pdf}

    {\tiny\hspace{.9cm}\hfill \textsf{Arctic Embed 2 / Llama} \hfill\hfill \textsf{Arctic Embed 2 / Qwen} \hfill\hfill \textsf{Arctic Embed 2 / Gemma} \hfill\textcolor{white}{ }}
    
    \includegraphics[trim={0 10pt 0 178.5pt},clip,width=\linewidth]{figures/boxplots_warg.pdf}
    
    \caption{WARG vs Spearman and Pearson correlation under prompt conditions \textsc{c2}~(\textit{top-k first}), \textsc{c3}~(\textit{random first}), and \textsc{c4}~(\textit{shuffled}) for all \ret{}-\gen{} combinations on PopQA.}
    \label{fig:boxplots_warg_popqa}
\end{figure*}

\begin{figure*}[p]
    \centering
    \includegraphics[trim={0 270pt 0 4pt},clip,width=\linewidth]{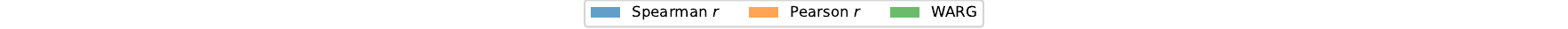}

    {\tiny\hspace{.7cm}\hfill \textsf{DRAGON / Llama} \hfill\hfill \textsf{DRAGON / Qwen} \hfill\hfill \textsf{DRAGON / Gemma} \hfill\textcolor{white}{ }}
    
    \includegraphics[trim={0 125pt 0 66pt},clip,width=\linewidth]{figures/boxplots_perturbation_qampari.pdf}

    {\tiny\hspace{.7cm}\hfill \textsf{Arctic Embed 2 / Llama} \hfill\hfill \textsf{Arctic Embed 2 / Qwen} \hfill\hfill \textsf{Arctic Embed 2 / Gemma} \hfill\textcolor{white}{ }}
    
    \includegraphics[trim={0 5pt 0 190pt},clip,width=\linewidth]{figures/boxplots_perturbation_qampari.pdf}
    
    \caption{WARG vs Spearman and Pearson correlation under prompt conditions \textsc{c2}~(\textit{top-k first}), \textsc{c3}~(\textit{random first}), and \textsc{c4}~(\textit{shuffled}) for all \ret{}-\gen{} combinations on QAMPARI.}
    \label{fig:boxplots_warg_qampari}
\end{figure*}

\begin{figure*}[p]
    \centering
    \includegraphics[trim={0 270pt 0 4pt},clip,width=\linewidth]{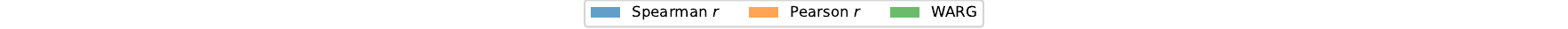}

    {\tiny\hspace{.9cm}\hfill \textsf{DRAGON / Llama} \hfill\hfill \textsf{DRAGON / Qwen} \hfill\hfill \textsf{DRAGON / Gemma} \hfill\textcolor{white}{ }}
    
    \includegraphics[trim={0 125pt 0 55pt},clip,width=\linewidth]{figures/boxplots_perturbation_trec.pdf}

    {\tiny\hspace{.9cm}\hfill \textsf{Arctic Embed 2 / Llama} \hfill\hfill \textsf{Arctic Embed 2 / Qwen} \hfill\hfill \textsf{Arctic Embed 2 / Gemma} \hfill\textcolor{white}{ }}
    
    \includegraphics[trim={0 8pt 0 178.5pt},clip,width=\linewidth]{figures/boxplots_perturbation_trec.pdf}
    
    \caption{WARG vs Spearman and Pearson correlation under prompt conditions \textsc{c2}~(\textit{top-k first}), \textsc{c3}~(\textit{random first}), and \textsc{c4}~(\textit{shuffled}) for all \ret{}-\gen{} combinations on TREC.}
    \label{fig:boxplots_warg_trec}
\end{figure*}

\begin{figure*}[p]
    \centering
    \begin{subfigure}{.35\textwidth}
        \centering
        \includegraphics[width=\linewidth]{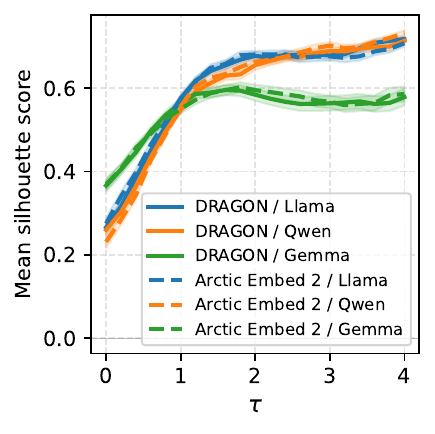}

        \hspace{10pt}
        
        \caption{}
        \label{fig:silhouette_vs_tau_popqa}
    \end{subfigure}
    \hfill
    \begin{subfigure}{.6\textwidth}
        \centering
        \includegraphics[width=\linewidth]{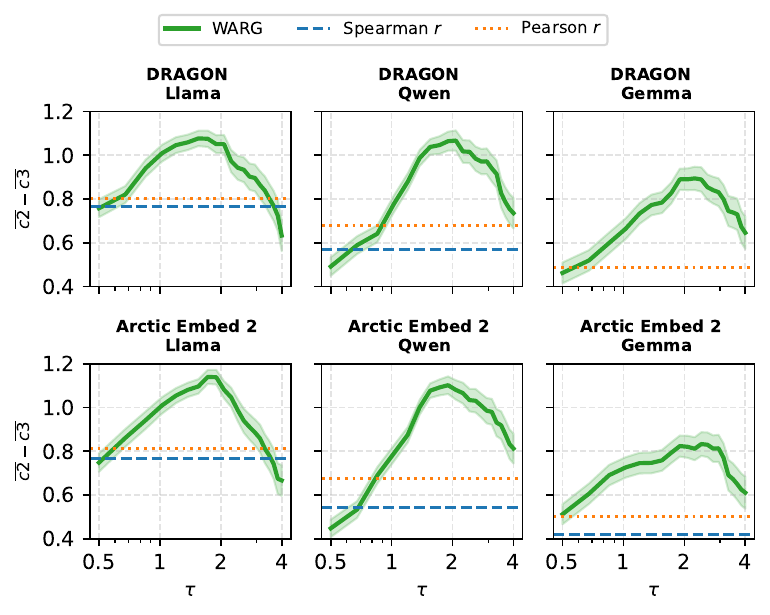}
        \caption{}
        \label{fig:warg_vs_tau_popqa}
    \end{subfigure}
    \caption[Clustering of \expandafter{\gen{}} attributions using the threshold parameter $\tau$ on PopQA.]{Clustering of \expandafter{\gen{}} attributions using the threshold parameter $\tau$ on PopQA.
    \textbf{(a):} Mean silhouette-score of the resulting clusters over $\tau$.
    \textbf{(b):} Mean difference $\overline{\textsc{c2}} - \overline{\textsc{c3}}$ of WARG as a function of $\tau$. Pearson and Spearman correlation computed with the same position-based oracle serve as $\tau$-independent baselines. A larger positive difference indicates that the metric successfully distinguishes the two conditions; a difference near zero would imply that the metric is equally satisfied by random documents in the early positions as by relevant ones.}
\end{figure*}

\begin{figure*}[p]
    \centering
    \includegraphics[width=.55\linewidth]{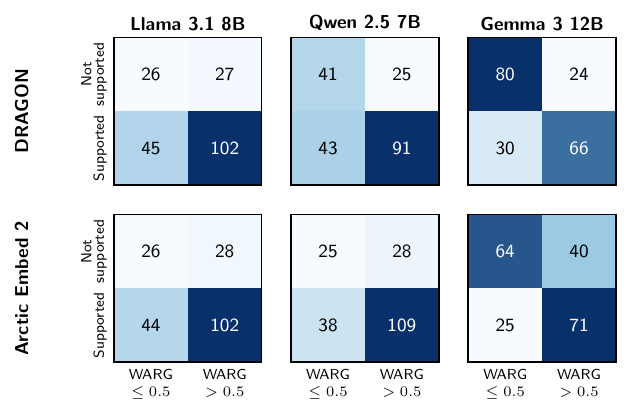}
    \caption{WARG vs QA downstream performance for all \ret{}-\gen{} combinations on PopQA.}
    \label{fig:confusion_warg_popqa}
\end{figure*}

\begin{figure*}[p]
    \centering
    \includegraphics[width=.55\linewidth]{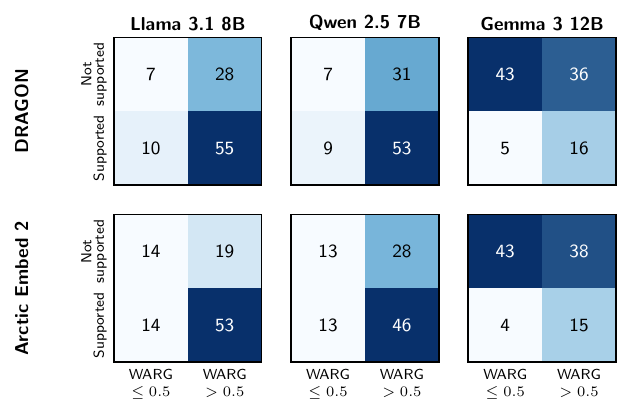}
    \caption{WARG vs QA downstream performance for all \ret{}-\gen{} combinations on QAMPARI.}
    \label{fig:confusion_warg_qampari}
\end{figure*}

In this section, we provide more detailed results of the WARG analysis. Figures~\ref{fig:boxplots_warg_popqa} to \ref{fig:boxplots_warg_trec} compare WARG, Spearman, and Pearson correlation across perturbation conditions and complements the results in Table~\ref{tab:agreement}. WARG assigns the highest scores under condition~\textsc{c2}, where the top-$k$ retrieved documents are placed first in the context, and the lowest scores under condition~\textsc{c3}, where random-corpus documents occupy those positions instead; mean values under the shuffled condition~\textsc{c4} are close to zero. Spearman and Pearson correlation follow the same ordering but exhibit a substantially narrower range across conditions.

We attribute this difference to the assumptions underlying each metric. Pearson correlation assumes a linear relationship between \ret{} and \gen{} scores, which does not fit the near-binary structure of \gen{} importance: documents are typically either relevant to or not, with few intermediate values. Spearman correlation avoids the linearity assumption by operating on ranks, but this introduces a different problem: the ranking step spreads the probability mass of irrelevant documents across many distinct positions, amplifying small differences among documents that the generator \gen{} ignores. WARG sidesteps both issues by thresholding the generator scores into a high-attention set and measuring retriever alignment only within that set, making it robust to the distribution of importance scores among irrelevant documents.

Figure~\ref{fig:warg_vs_tau_popqa} provides an analysis of WARG's sensitivity to the threshold parameter~$\tau$. It compares WARG under high alignment of \ret{} and \gen{} (condition \textsc{c2}) and low alignment (condition \textsc{c3}). Oracle retriever scores are assigned by context position~($k{-}1, \ldots, 0$), so the difference measures how well WARG distinguishes between alignment and disalignment. While WARG generally outperforms baselines for $\tau \in [1,3]$, WARG usually peaks for $\tau \in [1.5,2]$. This is supported by Figure~\ref{fig:silhouette_vs_tau_popqa}, which shows no meaningful improvement of cluster quality after $\tau = 1.5$. Therefore, the choice of $\tau = 1.5$ made in this paper is sensible.

Lastly, Figures~\ref{fig:confusion_warg_popqa} and~\ref{fig:confusion_warg_qampari} show more detailed versions of Figure~\ref{fig:warg_confusion_combined} with individual confusion matrices for all \ret{}-\gen{} combinations. Again, they support the findings discussed in the main text: generated answers in agreement with the ground truth are more than twice (PopQA) and four times (QAMPARI) as likely to achieve a $\text{WARG} > 0.5$.

\subsection{Qualitative Analysis}\label{app:showcase}
In this section, we showcase examples for the PopQA dataset, highlighting the extremes of our WARG attribution. We report, for every query, the attribution for the query and the retrieved documents, as well as the impact of each document on the generated text (\textit{Assistant}) computed with \xrag{}.
As a case study, we present two QA-pairs with correct answers (Figures~\ref{fig:attribution-popqa-high-true} \& \ref{fig:attribution-popqa-low-true}), as well as two QA-pairs with incorrect answers (Figures~\ref{fig:attribution-popqa-high-false} \& \ref{fig:attribution-popqa-low-false}). Nevertheless, we found similar patterns all over our tested data.
Overall, qualitative analysis of the QA pairs reveals two trends:

\noindent\textbf{(i)~Keyword search by the \ret{}:} The \ret{} seems to rank documents mainly by the occurrence of keywords in the query:
in Figure~\ref{fig:attribution-popqa-high-true} the \ret{} focuses on the name ``May'';
in Figure~\ref{fig:attribution-popqa-high-false} it focuses on the name ``Fat'';
and in Figure~\ref{fig:attribution-popqa-low-true}, retrieval fails to produce relevant documents while focusing on the phrase ``ancient time'' and punctuation.
While less apparent, Figure~\ref{fig:attribution-popqa-low-false} also shows signs of keyword matching as documents~1 and 8 are directly retrieved for the query words ``genre'' and ``Emmeline''. For the other documents, the \ret{} falls back to versions of these keywords, like variations of the name ``Emmeline'' and specific genres.

A POS analysis of queries and retrieved documents (Appendix~\ref{app:pos_analysis}) shows that attribution in the retrieval phase is dominated by nouns~(NOUN) and proper nouns~(PROPN): together they account for over $70\%$ of the top-30\% attributed tokens for Arctic Embed~2 and approximately $50\%$ for DRAGON, while all other POS categories individually contribute $10\%$ or less. This pattern supports that \ret{} prioritises entity and concept matching over structural or function words.

\noindent\textbf{(ii)~Binary document importance patterns for the \gen{}:} As a second trend, we see a high difference in document importance between relevant and irrelevant documents. In all four Figures, we see relevant documents at $33\%$ to $45\%$ document importance, while the remaining irrelevant documents are at $\le15\%$ (chance level: $10\% = \frac{1}{|\mathcal{D}|}$).

Figure~\ref{fig:silhouette_vs_tau_popqa} shows an analysis of cluster quality on PopQA in terms of silhouette score over a threshold~$\tau$, splitting the documents into a cluster of relevant documents with attribution scores $> \frac{\tau}{|\mathcal{D}|}$ and a cluster of irrelevant documents with attribution scores $\le \frac{\tau}{|\mathcal{D}|}$. The cluster quality improves linearly until  $\tau \approx 1.5$ and then stabilises, supporting our qualitative findings.

\begin{figure*}[p]
\centering

\definecolor{cl0}{RGB}{0,221,221}
\definecolor{cl1}{RGB}{221,0,221}
\definecolor{cl2}{RGB}{31,119,180}
\definecolor{cl3}{RGB}{255,127,14}
\definecolor{cl4}{RGB}{44,160,44}
\definecolor{cl5}{RGB}{214,39,40}
\definecolor{cl6}{RGB}{148,103,189}
\definecolor{cl7}{RGB}{140,86,75}
\definecolor{cl8}{RGB}{227,119,194}
\definecolor{cl9}{RGB}{127,127,127}
\definecolor{cl10}{RGB}{188,189,34}
\definecolor{cl11}{RGB}{23,190,207}

\resizetable{\linewidth}{
\begin{tabular}{rl}
\midrule

\small\tok{cl0!0}{\tok{cl0!0}{\textbf{Query:}}} & {\small
\tok{cl0!8}{Who}
\tok{cl0!2}{ was}
\tok{cl1!14}{ the}
\tok{cl1!70}{ screen}
\tok{cl1!72}{writer}
\tok{cl1!22}{ for}
\tok{cl0!100}{ May}
\tok{cl1!12}{?}
}
\\
\midrule

\small\tok{cl0!0}{\textbf{Doc.~1:}} & {\small
\tok{cl0!3}{ **}
\tok{cl0!28}{May}
\tok{cl0!4}{ (}
\tok{cl0!2}{film}
\tok{cl0!2}{):}
\tok{cl0!0}{**}
\tok{cl0!0}{ 2002}
\tok{cl0!0}{ film}
\tok{cl0!0}{ by}
\tok{cl0!0}{ Lucky}
\tok{cl0!0}{ Mc}
\tok{cl0!0}{Ke}
\tok{cl0!0}{e}
\tok{cl0!0}{ **}
\tok{cl0!10}{May}
\tok{cl0!0}{***}
\tok{cl0!0}{ is}
\tok{cl0!0}{ a}
\tok{cl0!0}{ 2002}
\tok{cl0!0}{ American}
\tok{cl0!1}{ psycholog}
\tok{cl0!0}{ical}
\tok{cl0!1}{ horror}
\tok{cl0!1}{ film}
\tok{cl0!4}{ written}
\tok{cl0!1}{ and}
\tok{cl0!1}{ direct}
\tok{cl0!0}{ed}
\tok{cl0!0}{ by}
\tok{cl0!0}{ Lucky}
\tok{cl0!0}{ Mc}
\tok{cl0!0}{Ke}
\tok{cl0!1}{e}
\tok{cl0!0}{ [...]}} \\

\midrule

\small\tok{cl0!0}{\textbf{Doc.~2:}} & {\small
\tok{cl0!1}{ **}
\tok{cl0!20}{May}
\tok{cl0!2}{ (}
\tok{cl0!7}{film}
\tok{cl0!1}{):}
\tok{cl0!0}{**}
\tok{cl0!0}{ "}
\tok{cl0!2}{Movie}
\tok{cl0!2}{ Review}
\tok{cl0!1}{ –}
\tok{cl0!11}{ May}
\tok{cl0!0}{".}
\tok{cl0!0}{ *}
\tok{cl0!0}{ The}
\tok{cl0!0}{ New}
\tok{cl0!0}{ York}
\tok{cl0!4}{ Times}
\tok{cl0!0}{*}
\tok{cl0!0}{.}
\tok{cl0!0}{ Retrieve}
\tok{cl0!0}{d}
\tok{cl0!1}{ July}
\tok{cl0!1}{ 9}
\tok{cl0!0}{,}
\tok{cl0!1}{ 2011.}
\tok{cl0!2}{ -}
\tok{cl0!0}{ **}
\tok{cl0!0}{ \^{}}
\tok{cl0!0}{**}
\tok{cl0!1}{ Brown}
\tok{cl0!0}{,}
\tok{cl0!5}{ Scott}
\tok{cl0!1}{ (}
\tok{cl0!5}{Jun}
\tok{cl0!0}{e}
\tok{cl0!3}{ 20}
\tok{cl0!6}{,}
\tok{cl0!3}{ 2003}
\tok{cl0!5}{).}
\tok{cl0!2}{ "}
\tok{cl0!6}{May}
\tok{cl0!5}{".}
\tok{cl0!1}{ *}
\tok{cl0!0}{ [...]}} \\

\midrule

\small\tok{cl0!0}{\textbf{Doc.~3:}} & {\small
\tok{cl0!0}{ **}
\tok{cl0!25}{May}
\tok{cl0!4}{ (}
\tok{cl0!10}{film}
\tok{cl0!1}{):}
\tok{cl0!0}{**}
\tok{cl0!1}{ -}
\tok{cl0!0}{ **}
\tok{cl0!0}{ \^{}}
\tok{cl0!0}{**}
\tok{cl0!0}{ Thomas}
\tok{cl0!0}{,}
\tok{cl0!4}{ Kevin}
\tok{cl0!1}{ (}
\tok{cl0!5}{Jun}
\tok{cl0!1}{e}
\tok{cl0!2}{ 20}
\tok{cl0!0}{,}
\tok{cl0!2}{ 2003}
\tok{cl0!0}{).}
\tok{cl0!0}{ "}
\tok{cl0!0}{ '}
\tok{cl0!13}{May}
\tok{cl0!0}{'}
\tok{cl0!0}{ just}
\tok{cl0!0}{ might}
\tok{cl0!0}{ be}
\tok{cl0!0}{ a}
\tok{cl0!0}{ cult}
\tok{cl0!0}{ classic}
\tok{cl0!3}{".}
\tok{cl0!1}{ *}
\tok{cl0!2}{ Los}
\tok{cl0!4}{ Angeles}
\tok{cl0!3}{ Times}
\tok{cl0!0}{*}
\tok{cl0!9}{.}
\tok{cl0!0}{ p}
\tok{cl0!4}{.}
\tok{cl0!1}{ E}
\tok{cl0!0}{8}
\tok{cl0!0}{ –}
\tok{cl0!0}{ via}
\tok{cl0!2}{ Newspaper}
\tok{cl0!0}{s}
\tok{cl0!0}{.}
\tok{cl0!1}{com}
\tok{cl0!4}{.}
\tok{cl0!0}{ [...]}} \\

\midrule

\small\tok{cl0!0}{\textbf{Doc.~4:}} & {\small
\tok{cl0!0}{ **}
\tok{cl0!37}{May}
\tok{cl0!3}{ (}
\tok{cl0!8}{film}
\tok{cl0!2}{):}
\tok{cl0!1}{**}
\tok{cl0!1}{ The}
\tok{cl0!4}{ film}
\tok{cl0!0}{ follow}
\tok{cl0!0}{s}
\tok{cl0!0}{ a}
\tok{cl0!0}{ lo}
\tok{cl0!1}{ne}
\tok{cl0!0}{ly}
\tok{cl0!1}{ young}
\tok{cl0!1}{ woman}
\tok{cl0!0}{ (}
\tok{cl0!1}{Bet}
\tok{cl0!0}{tis}
\tok{cl0!1}{)}
\tok{cl0!0}{ traumat}
\tok{cl0!0}{ized}
\tok{cl0!0}{ by}
\tok{cl0!0}{ a}
\tok{cl0!0}{ difficult}
\tok{cl0!0}{ child}
\tok{cl0!0}{hood}
\tok{cl0!0}{,}
\tok{cl0!0}{ and}
\tok{cl0!0}{ her}
\tok{cl0!0}{ increasing}
\tok{cl0!0}{ly}
\tok{cl0!0}{ desperat}
\tok{cl0!0}{e}
\tok{cl0!0}{ attempt}
\tok{cl0!0}{s}
\tok{cl0!0}{ to}
\tok{cl0!0}{ [...]}} \\

\midrule

\small\tok{cl0!0}{\textbf{Doc.~5:}} & {\small
\tok{cl0!1}{ **}
\tok{cl0!43}{May}
\tok{cl0!3}{ (}
\tok{cl0!8}{film}
\tok{cl0!1}{):}
\tok{cl0!1}{**}
\tok{cl0!3}{ Writing}
\tok{cl0!0}{ for}
\tok{cl0!0}{ the}
\tok{cl0!1}{ *}
\tok{cl0!0}{ Chicago}
\tok{cl0!0}{ Tribun}
\tok{cl0!0}{e}
\tok{cl0!0}{*}
\tok{cl0!0}{,}
\tok{cl0!0}{ Robert}
\tok{cl0!0}{ K}
\tok{cl0!0}{.}
\tok{cl0!2}{ El}
\tok{cl0!0}{der}
\tok{cl0!1}{ pra}
\tok{cl0!0}{ised}
\tok{cl0!1}{ the}
\tok{cl0!5}{ film}
\tok{cl0!0}{ as}
\tok{cl0!0}{ "}
\tok{cl0!0}{a}
\tok{cl0!0}{ refresh}
\tok{cl0!0}{ing}
\tok{cl0!1}{,}
\tok{cl0!1}{ maca}
\tok{cl0!0}{bre}
\tok{cl0!0}{ tale}
\tok{cl0!0}{."}
\tok{cl0!0}{ [...]}} \\

\midrule

\small\tok{cl0!0}{\textbf{Doc.~6:}} & {\small
\tok{cl0!1}{ **}
\tok{cl0!22}{May}
\tok{cl0!4}{ (}
\tok{cl0!7}{film}
\tok{cl0!1}{):}
\tok{cl0!0}{**}
\tok{cl0!0}{ "}
\tok{cl0!12}{May}
\tok{cl0!0}{ –}
\tok{cl0!1}{ Movie}
\tok{cl0!1}{ Review}
\tok{cl0!1}{".}
\tok{cl0!1}{ *}
\tok{cl0!0}{ The}
\tok{cl0!2}{ Austin}
\tok{cl0!4}{ Chro}
\tok{cl0!2}{nic}
\tok{cl0!0}{le}
\tok{cl0!0}{*}
\tok{cl0!1}{.}
\tok{cl0!0}{ Archive}
\tok{cl0!0}{d}
\tok{cl0!0}{ from}
\tok{cl0!0}{ the}
\tok{cl0!0}{ original}
\tok{cl0!0}{ on}
\tok{cl0!2}{ February}
\tok{cl0!1}{ 21,}
\tok{cl0!0}{ 20}
\tok{cl0!0}{24.}
\tok{cl0!0}{ [...]}} \\

\midrule

\small\tok{cl0!0}{\textbf{Doc.~7:}} & {\small
\tok{cl0!0}{ **}
\tok{cl0!22}{May}
\tok{cl0!3}{ (}
\tok{cl0!8}{film}
\tok{cl0!0}{):}
\tok{cl0!0}{**}
\tok{cl0!0}{ *}
\tok{cl0!0}{i}
\tok{cl0!2}{Hor}
\tok{cl0!2}{ror}
\tok{cl0!0}{*}
\tok{cl0!1}{.}
\tok{cl0!3}{ September}
\tok{cl0!2}{ 27,}
\tok{cl0!3}{ 2015.}
\tok{cl0!2}{ Archive}
\tok{cl0!0}{d}
\tok{cl0!0}{ from}
\tok{cl0!0}{ the}
\tok{cl0!1}{ original}
\tok{cl0!0}{ on}
\tok{cl0!1}{ October}
\tok{cl0!0}{ 8}
\tok{cl0!0}{,}
\tok{cl0!0}{ 2015.}
\tok{cl0!1}{ -}
\tok{cl0!0}{ **}
\tok{cl0!0}{ \^{}}
\tok{cl0!0}{**}
\tok{cl0!0}{ Pa}
\tok{cl0!2}{tter}
\tok{cl0!2}{son}
\tok{cl0!2}{,}
\tok{cl0!0}{ Robert}
\tok{cl0!0}{ (}
\tok{cl0!1}{Jan}
\tok{cl0!1}{uar}
\tok{cl0!0}{y}
\tok{cl0!0}{ 13,}
\tok{cl0!1}{ 2021}
\tok{cl0!2}{).}
\tok{cl0!0}{ [...]}} \\
\midrule

\small\tok{cl0!0}{\textbf{Doc.~8:}} & {\small
\tok{cl0!1}{ **}
\tok{cl0!25}{May}
\tok{cl0!3}{ (}
\tok{cl0!14}{film}
\tok{cl0!1}{):}
\tok{cl0!0}{**}
\tok{cl0!2}{ -}
\tok{cl0!0}{ **}
\tok{cl0!0}{ \^{}}
\tok{cl0!0}{**}
\tok{cl0!0}{ E}
\tok{cl0!2}{bert}
\tok{cl0!0}{,}
\tok{cl0!1}{ Roger}
\tok{cl0!0}{ (}
\tok{cl0!4}{Jun}
\tok{cl0!0}{e}
\tok{cl0!0}{ 6}
\tok{cl0!0}{,}
\tok{cl0!2}{ 2003}
\tok{cl0!1}{).}
\tok{cl0!1}{ "}
\tok{cl0!13}{May}
\tok{cl0!0}{".}
\tok{cl0!0}{ *}
\tok{cl0!1}{ Chicago}
\tok{cl0!1}{ Sun}
\tok{cl0!0}{-}
\tok{cl0!0}{Time}
\tok{cl0!0}{s}
\tok{cl0!0}{*}
\tok{cl0!10}{.}
\tok{cl0!1}{ Archive}
\tok{cl0!1}{d}
\tok{cl0!1}{ from}
\tok{cl0!0}{ the}
\tok{cl0!1}{ original}
\tok{cl0!1}{ on}
\tok{cl0!1}{ December}
\tok{cl0!3}{ 8}
\tok{cl0!8}{,}
\tok{cl0!1}{ 2005.}
\tok{cl0!0}{ [...]}} \\

\midrule

\small\tok{cl0!0}{\textbf{Doc.~9:}} & {\small
\tok{cl0!0}{ **}
\tok{cl0!24}{May}
\tok{cl0!4}{ (}
\tok{cl0!10}{film}
\tok{cl0!0}{):}
\tok{cl0!0}{**}
\tok{cl0!2}{ "}
\tok{cl0!1}{My}
\tok{cl0!0}{ Halloween}
\tok{cl0!3}{ season}
\tok{cl0!0}{ view}
\tok{cl0!0}{ing}
\tok{cl0!2}{ recommendations}
\tok{cl0!0}{:}
\tok{cl0!0}{ Lucky}
\tok{cl0!1}{ Mc}
\tok{cl0!1}{Ke}
\tok{cl0!0}{e}
\tok{cl0!0}{'}
\tok{cl0!0}{s}
\tok{cl0!6}{ MAY}
\tok{cl0!0}{,}
\tok{cl0!0}{ with}
\tok{cl0!0}{ the}
\tok{cl0!0}{ brilliant}
\tok{cl0!1}{ Angela}
\tok{cl0!1}{ Bett}
\tok{cl0!0}{is}
\tok{cl0!0}{ and}
\tok{cl0!2}{ Anna}
\tok{cl0!2}{ Far}
\tok{cl0!0}{is}
\tok{cl0!0}{ (...)}
\tok{cl0!1}{".}
\tok{cl0!0}{ [...]}} \\
\midrule

\small\tok{cl0!0}{\textbf{Doc.~10:}} & {\small
\tok{cl0!1}{ **}
\tok{cl0!46}{May}
\tok{cl0!6}{ (}
\tok{cl0!8}{film}
\tok{cl0!0}{):}
\tok{cl0!1}{**}
\tok{cl0!0}{ It}
\tok{cl0!0}{ eventually}
\tok{cl0!1}{ grosse}
\tok{cl0!0}{d}
\tok{cl0!0}{ \$}
\tok{cl0!0}{63}
\tok{cl0!1}{4,8}
\tok{cl0!0}{03}
\tok{cl0!0}{ worldwide}
\tok{cl0!1}{ on}
\tok{cl0!0}{ its}
\tok{cl0!0}{ \$}
\tok{cl0!0}{1.7}
\tok{cl0!1}{ million}
\tok{cl0!2}{ budget}
\tok{cl0!2}{.}
\tok{cl0!2}{ **}
\tok{cl0!1}{C}
\tok{cl0!1}{ri}
\tok{cl0!1}{tical}
\tok{cl0!0}{ response}
\tok{cl0!0}{**}
\tok{cl0!0}{ The}
\tok{cl0!3}{ film}
\tok{cl0!0}{ received}
\tok{cl0!0}{ favorable}
\tok{cl0!1}{ reviews}
\tok{cl0!0}{ from}
\tok{cl0!0}{ [...]}} \\
\\
\midrule

\small\tok{cl0!0}{\textbf{Assistant:}} & {\small
\tok{cl11!3}{}
\tok{cl6!30}{According}
\tok{cl5!15}{ to}
\tok{cl2!30}{ Document}
\tok{cl2!24}{ }
\tok{cl2!48}{1}
\tok{cl2!24}{,}
\tok{cl2!45}{ the}
\tok{cl2!67}{ screen}
\tok{cl2!53}{writer}
\tok{cl2!55}{ and}
\tok{cl2!64}{ director}
\tok{cl2!52}{ for}
\tok{cl2!12}{ the}
\tok{cl2!17}{ film}
\tok{cl2!8}{ "}
\tok{cl2!18}{May}
\tok{cl2!10}{"}
\tok{cl2!2}{ was}
\tok{cl2!100}{ Lucky}
\tok{cl2!64}{ McK}
\tok{cl2!42}{ee}
\tok{cl2!23}{.}
}
\\[-2pt]
& {\scriptsize\quad
\docchip{cl2}{Doc.~1}{45\%}\hspace{3pt}
\docchip{cl3}{Doc.~2}{-1\%}\hspace{3pt}
\docchip{cl4}{Doc.~3}{-10\%}\hspace{3pt}
\docchip{cl5}{Doc.~4}{4\%}\hspace{3pt}
\docchip{cl6}{Doc.~5}{5\%}\hspace{3pt}
\docchip{cl7}{Doc.~6}{8\%}\hspace{3pt}
\docchip{cl8}{Doc.~7}{-5\%}\hspace{3pt}
\docchip{cl9}{Doc.~8}{2\%}\hspace{3pt}
\docchip{cl10}{Doc.~9}{15\%}\hspace{3pt}
\docchip{cl11}{Doc.~10}{-6\%}
}
\\[-2pt]
\bottomrule
\end{tabular}
}

\caption{Visual example of token-level attributions on PopQA (\expandafter{\ret{}}: Arctic Embed 2, \expandafter{\gen{}}: Qwen, WARG: 0.477).
The correct answer is ``Edward Lucky McKee''.
Attribution for the query is coded as follows: \gradlegend{}.}
\label{fig:attribution-popqa-high-true}
\end{figure*}

\begin{figure*}[p]
\centering

\definecolor{cl0}{RGB}{0,221,221}
\definecolor{cl1}{RGB}{221,0,221}
\definecolor{cl2}{RGB}{31,119,180}
\definecolor{cl3}{RGB}{255,127,14}
\definecolor{cl4}{RGB}{44,160,44}
\definecolor{cl5}{RGB}{214,39,40}
\definecolor{cl6}{RGB}{148,103,189}
\definecolor{cl7}{RGB}{140,86,75}
\definecolor{cl8}{RGB}{227,119,194}
\definecolor{cl9}{RGB}{127,127,127}
\definecolor{cl10}{RGB}{188,189,34}
\definecolor{cl11}{RGB}{23,190,207}

\setlength{\tabcolsep}{2pt}
\renewcommand{\arraystretch}{1.1}

\resizetable{\linewidth}{
\begin{tabular}{rl}
\midrule

\small\tok{cl0!0}{\textbf{Query:}} & {\small
\tok{cl1!26}{Who}
\tok{cl0!9}{ is}
\tok{cl0!14}{ the}
\tok{cl1!80}{ author}
\tok{cl1!45}{ of}
\tok{cl0!100}{ Fat}
\tok{cl1!20}{?}
}
\\
\midrule

\small\tok{cl0!0}{\textbf{Doc.~1:}} & {\small
\tok{cl0!3}{ **}
\tok{cl0!12}{F}
\tok{cl0!13}{at}
\tok{cl0!1}{ (}
\tok{cl0!4}{no}
\tok{cl0!4}{vel}
\tok{cl0!0}{):}
\tok{cl0!0}{**}
\tok{cl0!2}{ 1969}
\tok{cl0!3}{ novel}
\tok{cl0!3}{ by}
\tok{cl0!1}{ Leonard}
\tok{cl0!0}{ Gard}
\tok{cl0!1}{ner}
\tok{cl0!0}{ **}
\tok{cl0!6}{F}
\tok{cl0!7}{at}
\tok{cl0!0}{ City}
\tok{cl0!1}{***}
\tok{cl0!2}{ is}
\tok{cl0!1}{ a}
\tok{cl0!4}{ novel}
\tok{cl0!2}{ by}
\tok{cl0!1}{ Leonard}
\tok{cl0!0}{ Gard}
\tok{cl0!1}{ner}
\tok{cl0!5}{ published}
\tok{cl0!3}{ in}
\tok{cl0!3}{ 1969}
\tok{cl0!1}{.}
}
\\
\midrule

\small\tok{cl0!0}{\textbf{Doc.~2:}} & {\small
\tok{cl0!0}{ **}
\tok{cl0!21}{F}
\tok{cl0!19}{at}
\tok{cl0!4}{ (}
\tok{cl0!5}{no}
\tok{cl0!9}{vel}
\tok{cl0!0}{):}
\tok{cl0!0}{**}
\tok{cl0!0}{ *}
\tok{cl0!0}{Sal}
\tok{cl0!0}{on}
\tok{cl0!0}{*}
\tok{cl0!2}{.}
\tok{cl0!0}{ Retrieve}
\tok{cl0!0}{d}
\tok{cl0!0}{ September}
\tok{cl0!0}{ 8}
\tok{cl0!0}{,}
\tok{cl0!2}{ 2021}
\tok{cl0!1}{.}
\tok{cl0!2}{ -}
\tok{cl0!0}{ **}
\tok{cl0!0}{ \^{}}
\tok{cl0!0}{**}
\tok{cl0!1}{ Dur}
\tok{cl0!3}{ham}
\tok{cl0!0}{,}
\tok{cl0!1}{ Michael}
\tok{cl0!1}{ (}
\tok{cl0!1}{August}
\tok{cl0!0}{ 29,}
\tok{cl0!2}{ 1969}
\tok{cl0!2}{).}
\tok{cl0!0}{ "}
\tok{cl0!2}{A}
\tok{cl0!1}{ short}
\tok{cl0!1}{ talk}
\tok{cl0!1}{ with}
\tok{cl0!1}{ a}
\tok{cl0!0}{ first}
\tok{cl0!5}{ novel}
\tok{cl0!3}{ist}
\tok{cl0!0}{".}
\tok{cl0!0}{ [...]}} \\

\midrule

\small\tok{cl0!0}{\textbf{Doc.~3:}} & {\small
\tok{cl0!0}{ **}
\tok{cl0!21}{F}
\tok{cl0!21}{at}
\tok{cl0!1}{ (}
\tok{cl0!3}{no}
\tok{cl0!5}{vel}
\tok{cl0!1}{):}
\tok{cl0!1}{**}
\tok{cl0!0}{ It}
\tok{cl0!0}{ is}
\tok{cl0!0}{ his}
\tok{cl0!0}{ only}
\tok{cl0!4}{ novel}
\tok{cl0!0}{.}
\tok{cl0!1}{ It}
\tok{cl0!0}{s}
\tok{cl0!1}{ pre}
\tok{cl0!1}{stige}
\tok{cl0!1}{ has}
\tok{cl0!1}{ grow}
\tok{cl0!0}{n}
\tok{cl0!0}{ since}
\tok{cl0!0}{ its}
\tok{cl0!2}{ publication}
\tok{cl0!0}{,}
\tok{cl0!1}{ due}
\tok{cl0!0}{ to}
\tok{cl0!0}{ critical}
\tok{cl0!0}{ ac}
\tok{cl0!0}{claim}
\tok{cl0!0}{ from}
\tok{cl0!1}{ Joan}
\tok{cl0!0}{ Did}
\tok{cl0!0}{ion}
\tok{cl0!0}{ and}
\tok{cl0!0}{ Walker}
\tok{cl0!0}{ Per}
\tok{cl0!0}{cy}
\tok{cl0!4}{,}
\tok{cl0!0}{ [...]}} \\

\midrule

\small\tok{cl0!0}{\textbf{Doc.~4:}} & {\small
\tok{cl0!0}{ **}
\tok{cl0!21}{F}
\tok{cl0!21}{at}
\tok{cl0!2}{ (}
\tok{cl0!3}{no}
\tok{cl0!6}{vel}
\tok{cl0!0}{):}
\tok{cl0!1}{**}
\tok{cl0!0}{ Of}
\tok{cl0!5}{ the}
\tok{cl0!4}{ novel}
\tok{cl0!2}{,}
\tok{cl0!1}{ Denis}
\tok{cl0!0}{ Johnson}
\tok{cl0!0}{ has}
\tok{cl0!0}{ state}
\tok{cl0!1}{d}
\tok{cl0!0}{ that}
\tok{cl0!1}{ "}
\tok{cl0!0}{I}
\tok{cl0!3}{ studie}
\tok{cl0!0}{d}
\tok{cl0!1}{ Leonard}
\tok{cl0!0}{ Gard}
\tok{cl0!0}{ner}
\tok{cl0!0}{'}
\tok{cl0!1}{s}
\tok{cl0!5}{ book}
\tok{cl0!0}{ so}
\tok{cl0!0}{ close}
\tok{cl0!0}{ly}
\tok{cl0!0}{ that}
\tok{cl0!0}{ I}
\tok{cl0!0}{ began}
\tok{cl0!0}{ to}
\tok{cl0!0}{ fear}
\tok{cl0!0}{ I}
\tok{cl0!0}{'}
\tok{cl0!1}{d}
\tok{cl0!1}{ never}
\tok{cl0!7}{ be}
\tok{cl0!0}{ able}
\tok{cl0!0}{ to}
\tok{cl0!0}{ [...]}} \\

\midrule

\small\tok{cl0!0}{\textbf{Doc.~5:}} & {\small
\tok{cl0!0}{ **}
\tok{cl0!10}{F}
\tok{cl0!15}{at}
\tok{cl0!2}{ (}
\tok{cl0!4}{no}
\tok{cl0!7}{vel}
\tok{cl0!0}{):}
\tok{cl0!0}{**}
\tok{cl0!0}{ -}
\tok{cl0!0}{ **}
\tok{cl0!0}{ \^{}}
\tok{cl0!0}{**}
\tok{cl0!0}{ "}
\tok{cl0!3}{Book}
\tok{cl0!1}{ Review}
\tok{cl0!0}{s}
\tok{cl0!0}{,}
\tok{cl0!0}{ Site}
\tok{cl0!0}{s}
\tok{cl0!0}{,}
\tok{cl0!1}{ Roman}
\tok{cl0!0}{ce}
\tok{cl0!0}{,}
\tok{cl0!0}{ Fantasy}
\tok{cl0!0}{,}
\tok{cl0!2}{ Fi}
\tok{cl0!1}{ction}
\tok{cl0!0}{".}
\tok{cl0!1}{ *}
\tok{cl0!0}{Kir}
\tok{cl0!0}{kus}
\tok{cl0!0}{ Review}
\tok{cl0!1}{s}
\tok{cl0!0}{*}
\tok{cl0!2}{.}
\tok{cl0!0}{ -}
\tok{cl0!0}{ **}
\tok{cl0!0}{ \^{}}
\tok{cl0!0}{**}
\tok{cl0!0}{ "}
\tok{cl0!2}{F}
\tok{cl0!2}{at}
\tok{cl0!2}{ City}
\tok{cl0!0}{".}
\tok{cl0!0}{ *}
\tok{cl0!0}{Pen}
\tok{cl0!0}{guin}
\tok{cl0!0}{ Rand}
\tok{cl0!0}{om}
\tok{cl0!0}{ House}
\tok{cl0!0}{*}
\tok{cl0!0}{.}
\tok{cl0!0}{ [...]}} \\

\midrule

\small\tok{cl0!0}{\textbf{Doc.~6:}} & {\small
\tok{cl0!0}{ **}
\tok{cl0!18}{F}
\tok{cl0!18}{at}
\tok{cl0!2}{ (}
\tok{cl0!8}{no}
\tok{cl0!11}{vel}
\tok{cl0!0}{):}
\tok{cl0!0}{**}
\tok{cl0!1}{ **}
\tok{cl0!0}{Con}
\tok{cl0!1}{tent}
\tok{cl0!0}{s}
\tok{cl0!0}{**}
\tok{cl0!0}{ -}
\tok{cl0!1}{ 1}
\tok{cl0!0}{ Plot}
\tok{cl0!0}{ summary}
\tok{cl0!0}{ -}
\tok{cl0!0}{ 2}
\tok{cl0!0}{ Me}
\tok{cl0!0}{a}
\tok{cl0!1}{ning}
\tok{cl0!0}{ of}
\tok{cl0!0}{ title}
\tok{cl0!0}{ -}
\tok{cl0!0}{ 3}
\tok{cl0!0}{ Critic}
\tok{cl0!0}{al}
\tok{cl0!0}{ reception}
\tok{cl0!0}{ and}
\tok{cl0!0}{ influence}
\tok{cl0!0}{ -}
\tok{cl0!0}{ 4}
\tok{cl0!0}{ Film}
\tok{cl0!8}{ }
\tok{cl0!0}{adaptation}
\tok{cl0!0}{ -}
\tok{cl0!0}{ 5}
\tok{cl0!0}{ Reference}
\tok{cl0!0}{s}
\tok{cl0!0}{ [...]}} \\

\midrule

\small\tok{cl0!0}{\textbf{Doc.~7:}} & {\small
\tok{cl0!0}{ **}
\tok{cl0!10}{F}
\tok{cl0!13}{at}
\tok{cl0!1}{ (}
\tok{cl0!4}{no}
\tok{cl0!6}{vel}
\tok{cl0!1}{):}
\tok{cl0!0}{**}
\tok{cl0!0}{ "}
\tok{cl0!2}{In}
\tok{cl0!1}{flu}
\tok{cl0!1}{ence}
\tok{cl0!2}{ Loo}
\tok{cl0!0}{ming}
\tok{cl0!0}{".}
\tok{cl0!0}{ *}
\tok{cl0!0}{The}
\tok{cl0!1}{ New}
\tok{cl0!2}{ York}
\tok{cl0!2}{ Times}
\tok{cl0!0}{*}
\tok{cl0!0}{.}
\tok{cl0!2}{ ISSN}
\tok{cl0!0}{ 03}
\tok{cl0!0}{62}
\tok{cl0!0}{-}
\tok{cl0!0}{43}
\tok{cl0!0}{31}
\tok{cl0!0}{.}
\tok{cl0!0}{ Retrieve}
\tok{cl0!0}{d}
\tok{cl0!0}{ September}
\tok{cl0!0}{ 8}
\tok{cl0!4}{,}
\tok{cl0!1}{ 2021}
\tok{cl0!5}{.}
\tok{cl0!0}{ [...]}} \\

\midrule

\small\tok{cl0!0}{\textbf{Doc.~8:}} & {\small
\tok{cl0!0}{ **}
\tok{cl0!13}{F}
\tok{cl0!14}{at}
\tok{cl0!3}{ (}
\tok{cl0!4}{no}
\tok{cl0!5}{vel}
\tok{cl0!1}{):}
\tok{cl0!0}{**}
\tok{cl0!3}{ Joan}
\tok{cl0!0}{ Did}
\tok{cl0!0}{ion}
\tok{cl0!0}{ said}
\tok{cl0!0}{ that}
\tok{cl0!0}{ "}
\tok{cl0!2}{ *}
\tok{cl0!3}{F}
\tok{cl0!3}{at}
\tok{cl0!1}{ City}
\tok{cl0!0}{*}
\tok{cl0!0}{ affected}
\tok{cl0!0}{ me}
\tok{cl0!0}{ more}
\tok{cl0!0}{ than}
\tok{cl0!0}{ any}
\tok{cl0!0}{ new}
\tok{cl0!2}{ fiction}
\tok{cl0!0}{ I}
\tok{cl0!0}{ have}
\tok{cl0!1}{ read}
\tok{cl0!0}{ in}
\tok{cl0!1}{ a}
\tok{cl0!1}{ long}
\tok{cl0!0}{ while}
\tok{cl0!95}{,}
\tok{cl0!10}{ and}
\tok{cl0!2}{ I}
\tok{cl0!2}{ do}
\tok{cl0!0}{ not}
\tok{cl0!0}{ think}
\tok{cl0!0}{ it}
\tok{cl0!0}{ affected}
\tok{cl0!0}{ [...]}} \\

\midrule

\small\tok{cl0!0}{\textbf{Doc.~9:}} & {\small
\tok{cl0!0}{ **}
\tok{cl0!10}{F}
\tok{cl0!13}{at}
\tok{cl0!2}{ (}
\tok{cl0!3}{no}
\tok{cl0!3}{vel}
\tok{cl0!1}{):}
\tok{cl0!0}{**}
\tok{cl0!1}{ Johnson}
\tok{cl0!0}{'}
\tok{cl0!0}{s}
\tok{cl0!6}{ book}
\tok{cl0!0}{ *}
\tok{cl0!1}{ Jesus}
\tok{cl0!0}{'}
\tok{cl0!0}{ Son}
\tok{cl0!0}{*}
\tok{cl0!0}{ has}
\tok{cl0!0}{ been}
\tok{cl0!0}{ compared}
\tok{cl0!1}{ to}
\tok{cl0!19}{ Fat}
\tok{cl0!4}{ City}
\tok{cl0!0}{ in}
\tok{cl0!0}{ both}
\tok{cl0!0}{ its}
\tok{cl0!0}{ content}
\tok{cl0!0}{ and}
\tok{cl0!0}{ style}
\tok{cl0!1}{.}
\tok{cl0!0}{ **}
\tok{cl0!1}{Film}
\tok{cl0!6}{ }
\tok{cl0!3}{adaptation}
\tok{cl0!0}{**}
\tok{cl0!0}{ The}
\tok{cl0!1}{ novel}
\tok{cl0!0}{ was}
\tok{cl0!1}{ adapt}
\tok{cl0!0}{ed}
\tok{cl0!0}{ [...]}} \\

\midrule

\small\tok{cl0!0}{\textbf{Doc.~10:}} & {\small
\tok{cl0!0}{ **}
\tok{cl0!13}{F}
\tok{cl0!14}{at}
\tok{cl0!2}{ (}
\tok{cl0!4}{no}
\tok{cl0!6}{vel}
\tok{cl0!0}{):}
\tok{cl0!0}{**}
\tok{cl0!0}{ The}
\tok{cl0!1}{ title}
\tok{cl0!0}{ is}
\tok{cl0!0}{ ironi}
\tok{cl0!0}{c}
\tok{cl0!1}{:}
\tok{cl0!27}{ Fat}
\tok{cl0!3}{ City}
\tok{cl0!1}{ is}
\tok{cl0!0}{ a}
\tok{cl0!0}{ crazy}
\tok{cl0!0}{ goal}
\tok{cl0!0}{ no}
\tok{cl0!0}{ one}
\tok{cl0!0}{ is}
\tok{cl0!0}{ ever}
\tok{cl0!0}{ going}
\tok{cl0!0}{ to}
\tok{cl0!0}{ reach}
\tok{cl0!3}{."}
\tok{cl0!1}{ **}
\tok{cl0!0}{C}
\tok{cl0!0}{ri}
\tok{cl0!0}{tical}
\tok{cl0!0}{ reception}
\tok{cl0!0}{ and}
\tok{cl0!1}{ influence}
\tok{cl0!0}{**}
\tok{cl0!0}{ Kirk}
\tok{cl0!0}{us}
\tok{cl0!0}{ Review}
\tok{cl0!0}{s}
\tok{cl0!0}{ [...]}} \\

\midrule

\small\tok{cl0!0}{\textbf{Assistant:}} & {\small
\tok{cl5!5}{urnished}
\tok{cl7!9}{}
\tok{cl11!26}{The}
\tok{cl2!27}{ author}
\tok{cl2!16}{ of}
\tok{cl2!7}{ *}
\tok{cl9!6}{Fat}
\tok{cl2!43}{ City}
\tok{cl9!6}{*}
\tok{cl9!4}{ is}
\tok{cl2!100}{ Leonard}
\tok{cl2!76}{ Gardner}
\tok{cl2!9}{.}
\tok{cl2!4}{ This}
\tok{cl3!4}{ information}
\tok{cl2!6}{ is}
\tok{cl2!19}{ directly}
\tok{cl2!46}{ stated}
\tok{cl2!17}{ in}
\tok{cl2!34}{ Document}
\tok{cl2!34}{ }
\tok{cl2!47}{1}
\tok{cl2!6}{ and}
\tok{cl2!9}{ supported}
\tok{cl2!7}{ by}
\tok{cl2!10}{ other}
\tok{cl2!20}{ documents}
\tok{cl7!6}{ that}
\tok{cl5!5}{ discuss}
} \\
& {\small
\tok{cl2!11}{ the}
\tok{cl2!13}{ novel}
\tok{cl2!9}{'s}
\tok{cl4!21}{ critical}
\tok{cl2!8}{ reception}
\tok{cl2!10}{ and}
\tok{cl5!15}{ influence}
\tok{cl11!5}{.}
}
\\[-2pt]
& {\scriptsize\quad
\docchip{cl2}{Doc.~1}{33\%}\hspace{3pt}
\docchip{cl3}{Doc.~2}{9\%}\hspace{3pt}
\docchip{cl4}{Doc.~3}{9\%}\hspace{3pt}
\docchip{cl5}{Doc.~4}{10\%}\hspace{3pt}
\docchip{cl6}{Doc.~5}{3\%}\hspace{3pt}
\docchip{cl7}{Doc.~6}{10\%}\hspace{3pt}
\docchip{cl8}{Doc.~7}{8\%}\hspace{3pt}
\docchip{cl9}{Doc.~8}{9\%}\hspace{3pt}
\docchip{cl10}{Doc.~9}{-0\%}\hspace{3pt}
\docchip{cl11}{Doc.~10}{8\%}
}
\\[-2pt]
\bottomrule
\end{tabular}
}

\caption{Visual example of token-level attributions on PopQA (\expandafter{\ret{}}: Arctic Embed 2, \expandafter{\gen{}}: Qwen, WARG: 1.000).
The correct answer is "Robert Grant".
Attribution for the query is coded as follows: \gradlegend{}.}
\label{fig:attribution-popqa-high-false}
\end{figure*}

\begin{figure*}[p]
\centering

\definecolor{cl0}{RGB}{0,221,221}
\definecolor{cl1}{RGB}{221,0,221}
\definecolor{cl2}{RGB}{31,119,180}
\definecolor{cl3}{RGB}{255,127,14}
\definecolor{cl4}{RGB}{44,160,44}
\definecolor{cl5}{RGB}{214,39,40}
\definecolor{cl6}{RGB}{148,103,189}
\definecolor{cl7}{RGB}{140,86,75}
\definecolor{cl8}{RGB}{227,119,194}
\definecolor{cl9}{RGB}{127,127,127}
\definecolor{cl10}{RGB}{188,189,34}
\definecolor{cl11}{RGB}{23,190,207}

\setlength{\tabcolsep}{2pt}
\renewcommand{\arraystretch}{1.1}

\resizetable{\linewidth}{
\begin{tabular}{rl}
\midrule

\small\tok{cl0!0}{\textbf{Query:}} & {\small
\tok{cl0!2}{Who}
\tok{cl1!12}{ is}
\tok{cl1!13}{ the}
\tok{cl1!100}{ author}
\tok{cl1!33}{ of}
\tok{cl1!81}{ And}
\tok{cl1!36}{ did}
\tok{cl1!33}{ those}
\tok{cl1!56}{ feet}
\tok{cl0!5}{ in}
\tok{cl0!39}{ an}
\tok{cl0!38}{cient}
\tok{cl0!27}{ time}
\tok{cl0!28}{?}
}
\\
\midrule

\small\tok{cl0!0}{\textbf{Doc.~1:}} & {\small
\tok{cl0!1}{ **}
\tok{cl0!3}{A}
\tok{cl0!1}{men}
\tok{cl0!2}{hot}
\tok{cl0!0}{ep}
\tok{cl0!0}{ II}
\tok{cl0!0}{:}
\tok{cl0!0}{**}
\tok{cl0!0}{ -}
\tok{cl0!0}{ \^{}}
\tok{cl0!0}{ **}
\tok{cl0!0}{a}
\tok{cl0!0}{***}
\tok{cl0!0}{ **}
\tok{cl0!0}{b}
\tok{cl0!0}{***}
\tok{cl0!0}{ Gard}
\tok{cl0!0}{iner}
\tok{cl0!0}{,}
\tok{cl0!0}{ p}
\tok{cl0!0}{.}
\tok{cl0!0}{200}
\tok{cl0!0}{ -}
\tok{cl0!0}{ **}
\tok{cl0!0}{ \^{}}
\tok{cl0!0}{**}
\tok{cl0!0}{ Red}
\tok{cl0!2}{ford}
\tok{cl0!1}{,}
\tok{cl0!0}{ Donald}
\tok{cl0!8}{ B}
\tok{cl0!10}{.}
\tok{cl0!13}{ *}
\tok{cl0!7}{Egy}
\tok{cl0!2}{pt}
\tok{cl0!4}{,}
\tok{cl0!3}{ Cana}
\tok{cl0!0}{an}
\tok{cl0!2}{,}
\tok{cl0!2}{ and}
\tok{cl0!2}{ Israel}
\tok{cl0!2}{ in}
\tok{cl0!6}{ An}
\tok{cl0!3}{cient}
\tok{cl0!0}{ Times}
\tok{cl0!100}{.}
\tok{cl0!10}{*}
\tok{cl0!7}{ p}
\tok{cl0!5}{.}
\tok{cl0!2}{ 162}
\tok{cl0!4}{.}
\tok{cl0!0}{ [...]}} \\

\midrule

\small\tok{cl0!0}{\textbf{Doc.~2:}} & {\small
\tok{cl0!1}{ **}
\tok{cl0!2}{A}
\tok{cl0!1}{men}
\tok{cl0!2}{hot}
\tok{cl0!0}{ep}
\tok{cl0!1}{ II}
\tok{cl0!1}{:}
\tok{cl0!0}{**}
\tok{cl0!0}{ -}
\tok{cl0!0}{ **}
\tok{cl0!0}{ \^{}}
\tok{cl0!0}{**}
\tok{cl0!0}{ Red}
\tok{cl0!3}{ford}
\tok{cl0!0}{,}
\tok{cl0!0}{ Donald}
\tok{cl0!0}{ B}
\tok{cl0!0}{.}
\tok{cl0!0}{ *}
\tok{cl0!2}{Egy}
\tok{cl0!0}{pt}
\tok{cl0!0}{,}
\tok{cl0!2}{ Cana}
\tok{cl0!0}{an}
\tok{cl0!1}{,}
\tok{cl0!3}{ and}
\tok{cl0!2}{ Israel}
\tok{cl0!2}{ in}
\tok{cl0!5}{ An}
\tok{cl0!4}{cient}
\tok{cl0!6}{ Times}
\tok{cl0!8}{.}
\tok{cl0!1}{*}
\tok{cl0!2}{ p}
\tok{cl0!35}{.}
\tok{cl0!2}{ 163}
\tok{cl0!2}{.}
\tok{cl0!0}{ Prince}
\tok{cl0!0}{ton}
\tok{cl0!2}{ University}
\tok{cl0!4}{ Press}
\tok{cl0!1}{,}
\tok{cl0!0}{ Prince}
\tok{cl0!0}{ton}
\tok{cl0!7}{ }
\tok{cl0!0}{NJ}
\tok{cl0!4}{,}
\tok{cl0!0}{ 1992.}
\tok{cl0!0}{ [...]}} \\

\midrule

\small\tok{cl0!0}{\textbf{Doc.~3:}} & {\small
\tok{cl0!1}{ **}
\tok{cl0!1}{A}
\tok{cl0!1}{men}
\tok{cl0!2}{hot}
\tok{cl0!0}{ep}
\tok{cl0!0}{ II}
\tok{cl0!0}{:}
\tok{cl0!0}{**}
\tok{cl0!0}{ *}
\tok{cl0!1}{Egy}
\tok{cl0!0}{pt}
\tok{cl0!0}{,}
\tok{cl0!2}{ Cana}
\tok{cl0!0}{an}
\tok{cl0!0}{,}
\tok{cl0!3}{ and}
\tok{cl0!2}{ Israel}
\tok{cl0!1}{ in}
\tok{cl0!3}{ An}
\tok{cl0!2}{cient}
\tok{cl0!6}{ Times}
\tok{cl0!0}{.}
\tok{cl0!0}{*}
\tok{cl0!0}{ p}
\tok{cl0!1}{.}
\tok{cl0!1}{ 164}
\tok{cl0!1}{.}
\tok{cl0!1}{ Prince}
\tok{cl0!0}{ton}
\tok{cl0!2}{ University}
\tok{cl0!3}{ Press}
\tok{cl0!0}{,}
\tok{cl0!1}{ Prince}
\tok{cl0!0}{ton}
\tok{cl0!2}{ }
\tok{cl0!1}{NJ}
\tok{cl0!4}{,}
\tok{cl0!0}{ 1992.}
\tok{cl0!2}{ -}
\tok{cl0!0}{ **}
\tok{cl0!0}{ [...]}} \\

\midrule

\small\tok{cl0!0}{\textbf{Doc.~4:}} & {\small
\tok{cl0!0}{ **}
\tok{cl0!0}{Association}
\tok{cl0!2}{ football}
\tok{cl0!0}{:}
\tok{cl0!0}{**}
\tok{cl0!0}{ Retrieve}
\tok{cl0!2}{d}
\tok{cl0!0}{ 2}
\tok{cl0!0}{ March}
\tok{cl0!0}{ 2010.}
\tok{cl0!1}{ -}
\tok{cl0!0}{ **}
\tok{cl0!0}{ \^{}}
\tok{cl0!0}{**}
\tok{cl0!4}{ Austin}
\tok{cl0!1}{,}
\tok{cl0!1}{ Mer}
\tok{cl0!0}{r}
\tok{cl0!1}{ill}
\tok{cl0!0}{ (10}
\tok{cl0!0}{ July}
\tok{cl0!2}{ 2007}
\tok{cl0!1}{).}
\tok{cl0!2}{ "}
\tok{cl0!1}{Best}
\tok{cl0!7}{ Fe}
\tok{cl0!8}{et}
\tok{cl0!0}{ For}
\tok{cl0!0}{ward}
\tok{cl0!0}{".}
\tok{cl0!0}{ *}
\tok{cl0!1}{Van}
\tok{cl0!0}{ity}
\tok{cl0!0}{ Fair}
\tok{cl0!0}{*}
\tok{cl0!5}{.}
\tok{cl0!0}{ [...]}} \\

\midrule

\small\tok{cl0!0}{\textbf{Doc.~5:}} & {\small
\tok{cl0!0}{ **}
\tok{cl0!1}{John}
\tok{cl0!0}{ Wy}
\tok{cl0!0}{n}
\tok{cl0!1}{dham}
\tok{cl0!0}{:}
\tok{cl0!0}{**}
\tok{cl0!0}{ -}
\tok{cl0!0}{ **}
\tok{cl0!0}{ \^{}}
\tok{cl0!0}{**}
\tok{cl0!0}{ Al}
\tok{cl0!2}{dis}
\tok{cl0!0}{s}
\tok{cl0!1}{ 1973}
\tok{cl0!0}{,}
\tok{cl0!0}{ p}
\tok{cl0!0}{.}
\tok{cl0!1}{ 254}
\tok{cl0!0}{.}
\tok{cl0!1}{ -}
\tok{cl0!0}{ **}
\tok{cl0!0}{ \^{}}
\tok{cl0!0}{**}
\tok{cl0!1}{ Walt}
\tok{cl0!0}{on}
\tok{cl0!0}{,}
\tok{cl0!0}{ Jo}
\tok{cl0!0}{ (27}
\tok{cl0!0}{ October}
\tok{cl0!0}{ 2008)}
\tok{cl0!4}{.}
\tok{cl0!0}{ "}
\tok{cl0!0}{Tele}
\tok{cl0!3}{pathy}
\tok{cl0!3}{ and}
\tok{cl0!2}{ Tri}
\tok{cl0!0}{b}
\tok{cl0!0}{ulation}
\tok{cl0!0}{:}
\tok{cl0!2}{ John}
\tok{cl0!1}{ Wy}
\tok{cl0!0}{n}
\tok{cl0!2}{dham}
\tok{cl0!2}{'}
\tok{cl0!1}{s}
\tok{cl0!5}{ The}
\tok{cl0!1}{ Chr}
\tok{cl0!2}{ysa}
\tok{cl0!1}{lid}
\tok{cl0!0}{s}
\tok{cl0!1}{".}
\tok{cl0!0}{ [...]}} \\

\midrule

\small\tok{cl0!0}{\textbf{Doc.~6:}} & {\small
\tok{cl0!0}{ **}
\tok{cl0!2}{Poli}
\tok{cl0!4}{tici}
\tok{cl0!2}{an}
\tok{cl0!0}{:}
\tok{cl0!0}{**}
\tok{cl0!1}{ *}
\tok{cl0!0}{Mo}
\tok{cl0!0}{dern}
\tok{cl0!1}{ political}
\tok{cl0!1}{ analysis}
\tok{cl0!0}{*}
\tok{cl0!0}{.}
\tok{cl0!0}{ Pre}
\tok{cl0!0}{n}
\tok{cl0!2}{tice}
\tok{cl0!1}{ Hall}
\tok{cl0!0}{.}
\tok{cl0!3}{ ISBN}
\tok{cl0!0}{ 0}
\tok{cl0!0}{-13}
\tok{cl0!0}{-0}
\tok{cl0!0}{497}
\tok{cl0!0}{02}
\tok{cl0!0}{-9}
\tok{cl0!1}{.}
\tok{cl0!2}{ OC}
\tok{cl0!1}{LC}
\tok{cl0!1}{ }
\tok{cl0!0}{496}
\tok{cl0!0}{11}
\tok{cl0!0}{149}
\tok{cl0!1}{.}
\tok{cl0!2}{ -}
\tok{cl0!6}{ Daniel}
\tok{cl0!0}{,}
\tok{cl0!0}{ G}
\tok{cl0!0}{lyn}
\tok{cl0!0}{ (2003)}
\tok{cl0!0}{ [}
\tok{cl0!0}{1968}
\tok{cl0!0}{]}
\tok{cl0!2}{.}
\tok{cl0!1}{ The}
\tok{cl0!2}{ First}
\tok{cl0!3}{ Civil}
\tok{cl0!2}{ization}
\tok{cl0!0}{s}
\tok{cl0!0}{:}
\tok{cl0!0}{ [...]}} \\

\midrule

\small\tok{cl0!0}{\textbf{Doc.~7:}} & {\small
\tok{cl0!0}{ **}
\tok{cl0!1}{T}
\tok{cl0!2}{hut}
\tok{cl0!0}{mos}
\tok{cl0!0}{e}
\tok{cl0!3}{ III}
\tok{cl0!0}{:}
\tok{cl0!0}{**}
\tok{cl0!0}{ p}
\tok{cl0!0}{.}
\tok{cl0!0}{ 340}
\tok{cl0!0}{.}
\tok{cl0!4}{ ISBN}
\tok{cl0!1}{ 97}
\tok{cl0!0}{805}
\tok{cl0!0}{20}
\tok{cl0!0}{30}
\tok{cl0!0}{58}
\tok{cl0!0}{47}
\tok{cl0!0}{.}
\tok{cl0!1}{ Retrieve}
\tok{cl0!0}{d}
\tok{cl0!0}{ 22}
\tok{cl0!0}{ October}
\tok{cl0!0}{ 2019.}
\tok{cl0!2}{ -}
\tok{cl0!1}{ **}
\tok{cl0!1}{ \^{}}
\tok{cl0!0}{**}
\tok{cl0!0}{ J}
\tok{cl0!2}{.}
\tok{cl0!0}{H}
\tok{cl0!20}{.}
\tok{cl0!1}{ Breast}
\tok{cl0!0}{ed}
\tok{cl0!0}{,}
\tok{cl0!3}{ An}
\tok{cl0!1}{cient}
\tok{cl0!7}{ Times}
\tok{cl0!1}{:}
\tok{cl0!1}{ A}
\tok{cl0!2}{ History}
\tok{cl0!0}{ of}
\tok{cl0!0}{ the}
\tok{cl0!1}{ E}
\tok{cl0!1}{ar}
\tok{cl0!1}{ly}
\tok{cl0!1}{ World}
\tok{cl0!2}{;}
\tok{cl0!0}{ [...]}} \\

\midrule

\small\tok{cl0!0}{\textbf{Doc.~8:}} & {\small
\tok{cl0!1}{ **}
\tok{cl0!0}{Em}
\tok{cl0!0}{ily}
\tok{cl0!0}{ Dick}
\tok{cl0!0}{in}
\tok{cl0!0}{son}
\tok{cl0!1}{:}
\tok{cl0!0}{**}
\tok{cl0!2}{ -}
\tok{cl0!1}{ Parker}
\tok{cl0!0}{,}
\tok{cl0!0}{ Peter}
\tok{cl0!0}{.}
\tok{cl0!1}{ 2007.}
\tok{cl0!0}{ "}
\tok{cl0!0}{New}
\tok{cl0!6}{ Fe}
\tok{cl0!7}{et}
\tok{cl0!1}{ With}
\tok{cl0!0}{in}
\tok{cl0!0}{ My}
\tok{cl0!1}{ Garden}
\tok{cl0!0}{ Go}
\tok{cl0!1}{:}
\tok{cl0!0}{ Emily}
\tok{cl0!2}{ Dick}
\tok{cl0!0}{in}
\tok{cl0!0}{son}
\tok{cl0!1}{'}
\tok{cl0!0}{s}
\tok{cl0!2}{ Her}
\tok{cl0!0}{ba}
\tok{cl0!0}{rium}
\tok{cl0!1}{",}
\tok{cl0!0}{ *}
\tok{cl0!0}{ The}
\tok{cl0!0}{ Daily}
\tok{cl0!0}{ Telegraph}
\tok{cl0!0}{*}
\tok{cl0!2}{,}
\tok{cl0!0}{ June}
\tok{cl0!0}{ 29,}
\tok{cl0!1}{ 2007.}
\tok{cl0!0}{ [...]}} \\

\midrule

\small\tok{cl0!0}{\textbf{Doc.~9:}} & {\small
\tok{cl0!0}{ **}
\tok{cl0!0}{India}
\tok{cl0!0}{:}
\tok{cl0!0}{**}
\tok{cl0!1}{ 16}
\tok{cl0!0}{*}
\tok{cl0!0}{.}
\tok{cl0!0}{ Madrid}
\tok{cl0!1}{:}
\tok{cl0!1}{ Vidu}
\tok{cl0!0}{a}
\tok{cl0!0}{ de}
\tok{cl0!1}{ Cal}
\tok{cl0!0}{ero}
\tok{cl0!0}{.}
\tok{cl0!0}{ p}
\tok{cl0!0}{.}
\tok{cl0!0}{ 5}
\tok{cl0!0}{16}
\tok{cl0!0}{.}
\tok{cl0!2}{ -}
\tok{cl0!0}{ **}
\tok{cl0!0}{ \^{}}
\tok{cl0!0}{**}
\tok{cl0!0}{ Marti}
\tok{cl0!0}{re}
\tok{cl0!0}{ d}
\tok{cl0!0}{’}
\tok{cl0!2}{Ang}
\tok{cl0!0}{hier}
\tok{cl0!0}{a}
\tok{cl0!1}{,}
\tok{cl0!0}{ Pietro}
\tok{cl0!0}{ (18}
\tok{cl0!0}{92)}
\tok{cl0!11}{.}
\tok{cl0!1}{ Jo}
\tok{cl0!0}{a}
\tok{cl0!0}{quin}
\tok{cl0!1}{ As}
\tok{cl0!1}{en}
\tok{cl0!0}{cio}
\tok{cl0!4}{ Torres}
\tok{cl0!1}{ (}
\tok{cl0!2}{ed}
\tok{cl0!9}{.}
\tok{cl0!2}{).}
\tok{cl0!0}{ [...]}} \\

\midrule

\small\tok{cl0!0}{\textbf{Doc.~10:}} & {\small
\tok{cl0!0}{ **}
\tok{cl0!3}{India}
\tok{cl0!0}{:}
\tok{cl0!0}{**}
\tok{cl0!0}{ Re}
\tok{cl0!1}{sili}
\tok{cl0!1}{ent}
\tok{cl0!0}{ Culture}
\tok{cl0!0}{s}
\tok{cl0!0}{:}
\tok{cl0!2}{ America}
\tok{cl0!0}{'}
\tok{cl0!0}{s}
\tok{cl0!0}{ Na}
\tok{cl0!1}{tive}
\tok{cl0!1}{ People}
\tok{cl0!0}{s}
\tok{cl0!1}{ Con}
\tok{cl0!0}{front}
\tok{cl0!0}{ European}
\tok{cl0!0}{ Colonia}
\tok{cl0!0}{l}
\tok{cl0!0}{ization}
\tok{cl0!1}{ 1500}
\tok{cl0!0}{-}
\tok{cl0!0}{1800}
\tok{cl0!0}{*}
\tok{cl0!1}{ (2}
\tok{cl0!1}{ ed}
\tok{cl0!1}{.}
\tok{cl0!2}{).}
\tok{cl0!1}{ Rou}
\tok{cl0!0}{t}
\tok{cl0!0}{ledge}
\tok{cl0!4}{.}
\tok{cl0!2}{ ISBN}
\tok{cl0!1}{ 978-}
\tok{cl0!0}{1-3}
\tok{cl0!0}{15}
\tok{cl0!0}{-50}
\tok{cl0!0}{987}
\tok{cl0!0}{-7}
\tok{cl0!6}{.}
\tok{cl0!0}{ [...]}} \\

\midrule

\small\tok{cl0!0}{\textbf{Assistant:}} & {\small
\tok{cl7!20}{:}
\tok{cl10!46}{ The}
\tok{cl3!38}{ provided}
\tok{cl6!7}{ documents}
\tok{cl9!5}{ do}
\tok{cl5!9}{ not}
\tok{cl2!7}{ contain}
\tok{cl6!3}{ information}
\tok{cl6!3}{ about}
\tok{cl10!4}{ the}
\tok{cl6!5}{ author}
\tok{cl8!2}{ of}
\tok{cl9!14}{ the}
\tok{cl3!6}{ phrase}
\tok{cl7!5}{ "}
\tok{cl9!12}{And}
\tok{cl5!9}{ did}
\tok{cl9!12}{ those}
\tok{cl9!8}{ feet}
\tok{cl9!5}{ in}
\tok{cl2!4}{ ancient}
\tok{cl6!10}{ time}
\tok{cl6!8}{."}
\tok{cl7!9}{ This}
\tok{cl3!6}{ phrase}
\tok{cl2!0}{ is}
\tok{cl8!3}{ actually}
\tok{cl6!3}{ from}
} \\
& {\small
\tok{cl4!2}{ the}
\tok{cl10!1}{ hym}
\tok{cl5!2}{n}
\tok{cl9!9}{ "}
\tok{cl6!3}{Jer}
\tok{cl6!4}{usalem}
\tok{cl6!2}{,"}
\tok{cl6!3}{ written}
\tok{cl10!3}{ by}
\tok{cl7!6}{ William}
\tok{cl9!3}{ Blake}
\tok{cl5!5}{.}
\tok{cl5!15}{ None}
\tok{cl2!6}{ of}
\tok{cl11!5}{ the}
\tok{cl9!12}{ documents}
\tok{cl2!3}{ listed}
\tok{cl4!8}{ p}
\tok{cl4!2}{ertain}
\tok{cl6!2}{ to}
\tok{cl2!9}{ this}
\tok{cl10!4}{ specific}
\tok{cl2!5}{ query}
\tok{cl6!2}{.}
}
\\[-2pt]
& {\scriptsize\quad
\docchip{cl2}{Doc.~1}{6\%}\hspace{3pt}
\docchip{cl3}{Doc.~2}{15\%}\hspace{3pt}
\docchip{cl4}{Doc.~3}{14\%}\hspace{3pt}
\docchip{cl5}{Doc.~4}{13\%}\hspace{3pt}
\docchip{cl6}{Doc.~5}{14\%}\hspace{3pt}
\docchip{cl7}{Doc.~6}{15\%}\hspace{3pt}
\docchip{cl8}{Doc.~7}{14\%}\hspace{3pt}
\docchip{cl9}{Doc.~8}{4\%}\hspace{3pt}
\docchip{cl10}{Doc.~9}{5\%}\hspace{3pt}
\docchip{cl11}{Doc.~10}{1\%}
}
\\[-2pt]
\bottomrule
\end{tabular}
}

\caption{Visual example of token-level attributions on PopQA (\expandafter{\ret{}}: Arctic Embed 2, \expandafter{\gen{}}: Qwen, WARG: -0.207).
The correct answer is "William Blake".
Attribution for the query is coded as follows: \gradlegend{}.}
\label{fig:attribution-popqa-low-true}
\end{figure*}

\begin{figure*}[p]
\centering

\definecolor{cl0}{RGB}{0,221,221}
\definecolor{cl1}{RGB}{221,0,221}
\definecolor{cl2}{RGB}{31,119,180}
\definecolor{cl3}{RGB}{255,127,14}
\definecolor{cl4}{RGB}{44,160,44}
\definecolor{cl5}{RGB}{214,39,40}
\definecolor{cl6}{RGB}{148,103,189}
\definecolor{cl7}{RGB}{140,86,75}
\definecolor{cl8}{RGB}{227,119,194}
\definecolor{cl9}{RGB}{127,127,127}
\definecolor{cl10}{RGB}{188,189,34}
\definecolor{cl11}{RGB}{23,190,207}
\definecolor{cl12}{RGB}{110,110,221}

\setlength{\tabcolsep}{2pt}
\renewcommand{\arraystretch}{1.1}

\resizetable{\linewidth}{
\begin{tabular}{rl}
\midrule

\small\tok{cl0!0}{\textbf{Query:}} & {\small
\tok{cl1!14}{What}
\tok{cl0!81}{ genre}
\tok{cl0!23}{ is}
\tok{cl12!100}{ Emme}
\tok{cl12!95}{line}
\tok{cl1!50}{?}
}
\\
\midrule

\small\tok{cl0!0}{\textbf{Doc.~1:}} & {\small
\tok{cl0!3}{ **}
\tok{cl0!0}{Al}
\tok{cl0!3}{terna}
\tok{cl0!2}{tive}
\tok{cl0!7}{ rock}
\tok{cl0!2}{:}
\tok{cl0!0}{**}
\tok{cl0!6}{ In}
\tok{cl0!3}{ the}
\tok{cl0!6}{ mid}
\tok{cl0!0}{-}
\tok{cl0!2}{1990}
\tok{cl0!2}{s}
\tok{cl0!3}{,}
\tok{cl0!0}{ Sunny}
\tok{cl0!0}{ Day}
\tok{cl0!1}{ Real}
\tok{cl0!3}{ Estate}
\tok{cl0!5}{ define}
\tok{cl0!1}{d}
\tok{cl0!6}{ the}
\tok{cl0!24}{ emo}
\tok{cl0!22}{ genre}
\tok{cl0!2}{.}
\tok{cl0!1}{ We}
\tok{cl0!2}{ezer}
\tok{cl0!0}{ '}
\tok{cl0!0}{s}
\tok{cl0!0}{ album}
\tok{cl0!0}{ *}
\tok{cl0!2}{ Pink}
\tok{cl0!0}{er}
\tok{cl0!1}{ton}
\tok{cl0!0}{*}
\tok{cl0!1}{ (1996)}
\tok{cl0!0}{ was}
\tok{cl0!0}{ also}
\tok{cl0!0}{ influ}
\tok{cl0!1}{ential}
\tok{cl0!11}{.}
\tok{cl0!0}{ [...]}} \\

\midrule

\small\tok{cl0!0}{\textbf{Doc.~2:}} & {\small
\tok{cl0!0}{ **}
\tok{cl0!16}{Rock}
\tok{cl0!23}{ music}
\tok{cl0!4}{:}
\tok{cl0!0}{**}
\tok{cl0!12}{ Er}
\tok{cl0!12}{lew}
\tok{cl0!34}{ine}
\tok{cl0!0}{,}
\tok{cl0!7}{ "}
\tok{cl0!3}{American}
\tok{cl0!7}{ Alternative}
\tok{cl0!9}{ Rock}
\tok{cl0!1}{ /}
\tok{cl0!4}{ Post}
\tok{cl0!0}{ P}
\tok{cl0!4}{unk}
\tok{cl0!0}{",}
\tok{cl0!1}{ in}
\tok{cl0!1}{ Bogdan}
\tok{cl0!2}{ov}
\tok{cl0!0}{ et}
\tok{cl0!2}{ al}
\tok{cl0!0}{.}
\tok{cl0!2}{,}
\tok{cl0!0}{ 2002}
\tok{cl0!3}{,}
\tok{cl0!4}{ pp}
\tok{cl0!1}{.}
\tok{cl0!1}{ 13}
\tok{cl0!0}{44}
\tok{cl0!0}{–}
\tok{cl0!0}{13}
\tok{cl0!0}{46}
\tok{cl0!14}{.}
\tok{cl0!0}{ [...]}} \\

\midrule

\small\tok{cl0!0}{\textbf{Doc.~3:}} & {\small
\tok{cl0!1}{ **}
\tok{cl0!31}{Em}
\tok{cl0!9}{ily}
\tok{cl0!0}{ Dick}
\tok{cl0!2}{in}
\tok{cl0!0}{son}
\tok{cl0!0}{:}
\tok{cl0!0}{**}
\tok{cl0!0}{ She}
\tok{cl0!3}{ state}
\tok{cl0!3}{d}
\tok{cl0!5}{,}
\tok{cl0!2}{ "}
\tok{cl0!2}{if}
\tok{cl0!0}{ my}
\tok{cl0!2}{ lyrics}
\tok{cl0!4}{ sound}
\tok{cl0!3}{ like}
\tok{cl0!2}{ a}
\tok{cl0!2}{ letter}
\tok{cl0!0}{ written}
\tok{cl0!1}{ by}
\tok{cl0!7}{ Emily}
\tok{cl0!2}{ Dick}
\tok{cl0!0}{in}
\tok{cl0!1}{son}
\tok{cl0!0}{'}
\tok{cl0!2}{s}
\tok{cl0!1}{ great}
\tok{cl0!0}{-}
\tok{cl0!0}{grand}
\tok{cl0!0}{mo}
\tok{cl0!0}{ther}
\tok{cl0!0}{ while}
\tok{cl0!0}{ se}
\tok{cl0!1}{wing}
\tok{cl0!0}{ a}
\tok{cl0!2}{ la}
\tok{cl0!2}{ce}
\tok{cl0!3}{ curta}
\tok{cl0!1}{in}
\tok{cl0!27}{,}
\tok{cl0!3}{ that}
\tok{cl0!0}{'}
\tok{cl0!3}{s}
\tok{cl0!0}{ [...]}} \\

\midrule

\small\tok{cl0!0}{\textbf{Doc.~4:}} & {\small
\tok{cl0!0}{ **}
\tok{cl0!0}{Al}
\tok{cl0!3}{terna}
\tok{cl0!1}{tive}
\tok{cl0!10}{ rock}
\tok{cl0!0}{:}
\tok{cl0!0}{**}
\tok{cl0!1}{ "}
\tok{cl0!5}{G}
\tok{cl0!3}{rung}
\tok{cl0!10}{e}
\tok{cl0!1}{:}
\tok{cl0!1}{ A}
\tok{cl0!3}{ Success}
\tok{cl0!2}{ Story}
\tok{cl0!3}{".}
\tok{cl0!1}{ *}
\tok{cl0!1}{The}
\tok{cl0!3}{ New}
\tok{cl0!2}{ York}
\tok{cl0!3}{ Times}
\tok{cl0!0}{*}
\tok{cl0!0}{.}
\tok{cl0!2}{ November}
\tok{cl0!1}{ 15}
\tok{cl0!1}{,}
\tok{cl0!0}{ 1992.}
\tok{cl0!2}{ -}
\tok{cl0!0}{ \^{}}
\tok{cl0!0}{ **}
\tok{cl0!0}{a}
\tok{cl0!1}{***}
\tok{cl0!0}{ **}
\tok{cl0!0}{b}
\tok{cl0!1}{***}
\tok{cl0!0}{ Az}
\tok{cl0!0}{er}
\tok{cl0!0}{rad}
\tok{cl0!1}{ 1994}
\tok{cl0!0}{,}
\tok{cl0!4}{ p}
\tok{cl0!39}{.}
\tok{cl0!1}{ 160}
\tok{cl0!39}{.}
\tok{cl0!0}{ [...]}} \\

\midrule

\small\tok{cl0!0}{\textbf{Doc.~5:}} & {\small
\tok{cl0!2}{ **}
\tok{cl0!0}{Al}
\tok{cl0!2}{terna}
\tok{cl0!0}{tive}
\tok{cl0!7}{ rock}
\tok{cl0!3}{:}
\tok{cl0!1}{**}
\tok{cl0!0}{ 1}
\tok{cl0!2}{ single}
\tok{cl0!0}{s}
\tok{cl0!1}{ and}
\tok{cl0!1}{ *}
\tok{cl0!1}{ What}
\tok{cl0!2}{ever}
\tok{cl0!0}{ People}
\tok{cl0!1}{ Say}
\tok{cl0!0}{ I}
\tok{cl0!7}{ Am}
\tok{cl0!3}{,}
\tok{cl0!0}{ That}
\tok{cl0!1}{'}
\tok{cl0!0}{s}
\tok{cl0!1}{ What}
\tok{cl0!0}{ I}
\tok{cl0!0}{'}
\tok{cl0!0}{m}
\tok{cl0!1}{ Not}
\tok{cl0!0}{*}
\tok{cl0!1}{ (2006)}
\tok{cl0!1}{,}
\tok{cl0!2}{ which}
\tok{cl0!0}{ became}
\tok{cl0!1}{ the}
\tok{cl0!1}{ fast}
\tok{cl0!0}{est}
\tok{cl0!0}{-}
\tok{cl0!0}{sel}
\tok{cl0!3}{ling}
\tok{cl0!2}{ debut}
\tok{cl0!0}{ album}
\tok{cl0!0}{ in}
\tok{cl0!0}{ [...]}} \\

\midrule

\small\tok{cl0!0}{\textbf{Doc.~6:}} & {\small
\tok{cl0!1}{ **}
\tok{cl0!15}{Rock}
\tok{cl0!19}{ music}
\tok{cl0!1}{:}
\tok{cl0!0}{**}
\tok{cl0!11}{ Er}
\tok{cl0!9}{lew}
\tok{cl0!23}{ine}
\tok{cl0!2}{,}
\tok{cl0!3}{ "}
\tok{cl0!5}{Que}
\tok{cl0!5}{en}
\tok{cl0!0}{",}
\tok{cl0!0}{ *}
\tok{cl0!0}{All}
\tok{cl0!3}{Music}
\tok{cl0!0}{*}
\tok{cl0!1}{,}
\tok{cl0!0}{ archiv}
\tok{cl0!0}{ed}
\tok{cl0!0}{ from}
\tok{cl0!0}{ the}
\tok{cl0!0}{ original}
\tok{cl0!1}{ on}
\tok{cl0!0}{ 12}
\tok{cl0!0}{ February}
\tok{cl0!4}{ 2011.}
\tok{cl0!5}{ -}
\tok{cl0!0}{ **}
\tok{cl0!0}{ \^{}}
\tok{cl0!0}{**}
\tok{cl0!1}{ J}
\tok{cl0!86}{.}
\tok{cl0!2}{ Dou}
\tok{cl0!5}{gan}
\tok{cl0!9}{,}
\tok{cl0!1}{ "}
\tok{cl0!0}{Th}
\tok{cl0!2}{in}
\tok{cl0!2}{ Li}
\tok{cl0!1}{zzy}
\tok{cl0!1}{",}
\tok{cl0!0}{ *}
\tok{cl0!0}{All}
\tok{cl0!3}{Music}
\tok{cl0!0}{*}
\tok{cl0!0}{ [...]}} \\

\midrule

\small\tok{cl0!0}{\textbf{Doc.~7:}} & {\small
\tok{cl0!1}{ **}
\tok{cl0!18}{Rock}
\tok{cl0!19}{ music}
\tok{cl0!0}{:}
\tok{cl0!0}{**}
\tok{cl0!1}{ Not}
\tok{cl0!5}{ La}
\tok{cl0!11}{me}
\tok{cl0!0}{ Record}
\tok{cl0!4}{ings}
\tok{cl0!1}{.}
\tok{cl0!1}{ p}
\tok{cl0!1}{.}
\tok{cl0!0}{ 27.}
\tok{cl0!3}{ ISBN}
\tok{cl0!0}{ 978-}
\tok{cl0!0}{0}
\tok{cl0!1}{-}
\tok{cl0!0}{97}
\tok{cl0!0}{977}
\tok{cl0!0}{14}
\tok{cl0!0}{-0}
\tok{cl0!1}{-8}
\tok{cl0!2}{.}
\tok{cl0!3}{ -}
\tok{cl0!1}{ **}
\tok{cl0!1}{ \^{}}
\tok{cl0!0}{**}
\tok{cl0!1}{ S}
\tok{cl0!2}{.}
\tok{cl0!0}{T}
\tok{cl0!2}{.}
\tok{cl0!5}{ Er}
\tok{cl0!4}{lew}
\tok{cl0!14}{ine}
\tok{cl0!11}{,}
\tok{cl0!2}{ "}
\tok{cl0!0}{New}
\tok{cl0!1}{ Wave}
\tok{cl0!0}{",}
\tok{cl0!0}{ in}
\tok{cl0!1}{ Bogdan}
\tok{cl0!0}{ov}
\tok{cl0!0}{ et}
\tok{cl0!5}{ al}
\tok{cl0!100}{.}
\tok{cl0!7}{,}
\tok{cl0!0}{ 2002}
\tok{cl0!2}{,}
\tok{cl0!0}{ pp}
\tok{cl0!2}{.}
\tok{cl0!0}{ 13}
\tok{cl0!0}{37}
\tok{cl0!0}{–1}
\tok{cl0!1}{338}
\tok{cl0!0}{.}
\tok{cl0!0}{ [...]}} \\

\midrule

\small\tok{cl0!0}{\textbf{Doc.~8:}} & {\small
\tok{cl0!0}{ **}
\tok{cl0!32}{E}
\tok{cl0!33}{mme}
\tok{cl0!53}{line}
\tok{cl0!1}{ Pan}
\tok{cl0!0}{khu}
\tok{cl0!5}{rst}
\tok{cl0!5}{:}
\tok{cl0!2}{**}
\tok{cl0!2}{ *}
\tok{cl0!3}{ Collins}
\tok{cl0!2}{ English}
\tok{cl0!3}{ Dictionary}
\tok{cl0!1}{*}
\tok{cl0!0}{.}
\tok{cl0!4}{ Har}
\tok{cl0!9}{per}
\tok{cl0!1}{Col}
\tok{cl0!7}{lin}
\tok{cl0!1}{s}
\tok{cl0!0}{.}
\tok{cl0!3}{ OC}
\tok{cl0!1}{LC}
\tok{cl0!0}{ 11}
\tok{cl0!0}{204}
\tok{cl0!0}{11}
\tok{cl0!1}{289}
\tok{cl0!0}{.}
\tok{cl0!3}{ -}
\tok{cl0!1}{ **}
\tok{cl0!0}{ \^{}}
\tok{cl0!0}{**}
\tok{cl0!0}{ Rid}
\tok{cl0!2}{dell}
\tok{cl0!2}{,}
\tok{cl0!0}{ Fern}
\tok{cl0!2}{ (6}
\tok{cl0!1}{ February}
\tok{cl0!1}{ 2018}
\tok{cl0!8}{).}
\tok{cl0!2}{ "}
\tok{cl0!0}{S}
\tok{cl0!0}{uff}
\tok{cl0!4}{rage}
\tok{cl0!7}{ttes}
\tok{cl0!2}{,}
\tok{cl0!0}{ [...]}} \\

\midrule

\small\tok{cl0!0}{\textbf{Doc.~9:}} & {\small
\tok{cl0!1}{ **}
\tok{cl0!18}{Je}
\tok{cl0!3}{ff}
\tok{cl0!19}{ Lem}
\tok{cl0!15}{ire}
\tok{cl0!7}{:}
\tok{cl0!0}{**}
\tok{cl0!2}{ He}
\tok{cl0!1}{ is}
\tok{cl0!1}{ the}
\tok{cl0!5}{ author}
\tok{cl0!1}{ of}
\tok{cl0!3}{ critical}
\tok{cl0!1}{ly}
\tok{cl0!1}{ ac}
\tok{cl0!0}{claim}
\tok{cl0!0}{ed}
\tok{cl0!0}{ title}
\tok{cl0!0}{s}
\tok{cl0!0}{ including}
\tok{cl0!1}{ the}
\tok{cl0!0}{ *}
\tok{cl0!7}{ Esse}
\tok{cl0!0}{x}
\tok{cl0!3}{ County}
\tok{cl0!2}{ Tri}
\tok{cl0!3}{log}
\tok{cl0!3}{y}
\tok{cl0!0}{*}
\tok{cl0!0}{,}
\tok{cl0!0}{ *}
\tok{cl0!1}{ Sweet}
\tok{cl0!1}{ To}
\tok{cl0!1}{oth}
\tok{cl0!0}{*}
\tok{cl0!9}{,}
\tok{cl0!1}{ and}
\tok{cl0!0}{ *}
\tok{cl0!1}{ The}
\tok{cl0!0}{ No}
\tok{cl0!0}{body}
\tok{cl0!0}{*}
\tok{cl0!3}{.}
\tok{cl0!0}{ [...]}} \\

\midrule

\small\tok{cl0!0}{\textbf{Doc.~10:}} & {\small
\tok{cl0!2}{ **}
\tok{cl0!20}{Rock}
\tok{cl0!29}{ music}
\tok{cl0!3}{:}
\tok{cl0!0}{**}
\tok{cl0!9}{ Er}
\tok{cl0!12}{lew}
\tok{cl0!29}{ine}
\tok{cl0!1}{,}
\tok{cl0!3}{ "}
\tok{cl0!0}{Post}
\tok{cl0!2}{ P}
\tok{cl0!3}{unk}
\tok{cl0!1}{",}
\tok{cl0!0}{ in}
\tok{cl0!2}{ Bogdan}
\tok{cl0!1}{ov}
\tok{cl0!0}{ et}
\tok{cl0!1}{ al}
\tok{cl0!0}{.}
\tok{cl0!1}{,}
\tok{cl0!1}{ 2002}
\tok{cl0!0}{,}
\tok{cl0!1}{ pp}
\tok{cl0!1}{.}
\tok{cl0!2}{ 13}
\tok{cl0!0}{37}
\tok{cl0!0}{–1}
\tok{cl0!0}{338}
\tok{cl0!5}{.}
\tok{cl0!7}{ -}
\tok{cl0!0}{ **}
\tok{cl0!0}{ \^{}}
\tok{cl0!0}{**}
\tok{cl0!0}{ Good}
\tok{cl0!0}{lad}
\tok{cl0!0}{ \&}
\tok{cl0!0}{ Bi}
\tok{cl0!1}{bby}
\tok{cl0!2}{ 2007,}
\tok{cl0!3}{ p}
\tok{cl0!35}{.}
\tok{cl0!1}{ 2}
\tok{cl0!0}{39}
\tok{cl0!3}{ -}
\tok{cl0!1}{ **}
\tok{cl0!0}{ \^{}}
\tok{cl0!0}{**}
\tok{cl0!0}{ C}
\tok{cl0!11}{.}
\tok{cl0!4}{ Ger}
\tok{cl0!5}{e}
\tok{cl0!1}{,}
\tok{cl0!0}{ [...]}} \\

\midrule

\small\tok{cl0!0}{\textbf{Assistant:}} & {\small
\tok{cl11!11}{ Everton}
\tok{cl11!7}{'s}
\tok{cl11!21}{ query}
\tok{cl11!8}{ does}
\tok{cl4!2}{ not}
\tok{cl9!15}{ align}
\tok{cl9!6}{ with}
\tok{cl11!6}{ any}
\tok{cl11!4}{ of}
\tok{cl6!3}{ the}
\tok{cl10!4}{ provided}
\tok{cl10!6}{ documents}
\tok{cl2!2}{.}
\tok{cl9!9}{ Emm}
\tok{cl9!26}{eline}
\tok{cl9!100}{ P}
\tok{cl9!92}{ank}
\tok{cl9!44}{hurst}
\tok{cl9!23}{ is}
\tok{cl6!2}{ a}
\tok{cl9!14}{ historical}
\tok{cl9!15}{ figure}
\tok{cl9!6}{ known}
\tok{cl10!1}{ for}
\tok{cl9!16}{ her}
\tok{cl9!7}{ role}
\tok{cl9!1}{ in}
\tok{cl9!5}{ the}
\tok{cl9!8}{ women}
\tok{cl9!15}{'s}
\tok{cl4!3}{ suff}
} \\
& {\small
\tok{cl9!7}{rage}
\tok{cl9!3}{ movement}
\tok{cl11!0}{,}
\tok{cl6!3}{ and}
\tok{cl9!20}{ none}
\tok{cl9!5}{ of}
\tok{cl9!4}{ the}
\tok{cl9!11}{ documents}
\tok{cl5!4}{ mention}
\tok{cl9!21}{ her}
\tok{cl9!16}{ in}
\tok{cl11!3}{ relation}
\tok{cl9!15}{ to}
\tok{cl9!7}{ a}
\tok{cl5!8}{ musical}
\tok{cl9!12}{ genre}
\tok{cl6!3}{.}
\tok{cl9!4}{ If}
\tok{cl9!6}{ you}
\tok{cl9!4}{ are}
\tok{cl9!10}{ looking}
\tok{cl9!6}{ for}
\tok{cl5!6}{ information}
\tok{cl9!1}{ about}
\tok{cl5!9}{ a}
\tok{cl9!5}{ specific}
\tok{cl9!30}{ genre}
\tok{cl2!4}{ or}
} \\
& {\small
\tok{cl2!16}{ artist}
\tok{cl2!3}{,}
\tok{cl6!3}{ please}
\tok{cl6!0}{ provide}
\tok{cl10!0}{ more}
\tok{cl9!4}{ context}
\tok{cl9!3}{ or}
\tok{cl9!3}{ clarify}
\tok{cl9!16}{ your}
\tok{cl9!8}{ question}
\tok{cl9!5}{.}
}
\\[-2pt]
& {\scriptsize\quad
\docchip{cl2}{Doc.~1}{-4\%}\hspace{3pt}
\docchip{cl3}{Doc.~2}{5\%}\hspace{3pt}
\docchip{cl4}{Doc.~3}{3\%}\hspace{3pt}
\docchip{cl5}{Doc.~4}{3\%}\hspace{3pt}
\docchip{cl6}{Doc.~5}{0\%}\hspace{3pt}
\docchip{cl7}{Doc.~6}{10\%}\hspace{3pt}
\docchip{cl8}{Doc.~7}{14\%}\hspace{3pt}
\docchip{cl9}{Doc.~8}{41\%}\hspace{3pt}
\docchip{cl10}{Doc.~9}{9\%}\hspace{3pt}
\docchip{cl11}{Doc.~10}{10\%}
}
\\[-2pt]
\bottomrule
\end{tabular}
}

\caption{Visual example of token-level attributions on PopQA (\expandafter{\ret{}}: Arctic Embed 2, \expandafter{\gen{}}: Qwen, WARG: -0.622).
The correct answer is "Gothic fiction" / "Gothic novel" / "Gothic literature".
Attribution for the query is coded as follows: \gradlegend{}.}
\label{fig:attribution-popqa-low-false}
\end{figure*}

\end{document}